\documentclass[letterpaper, 10 pt, conference]{ieeeconf}  

\IEEEoverridecommandlockouts                              

\usepackage[nopatch=eqnum]{microtype}
\usepackage{graphicx}
\usepackage{overpic}
\usepackage{url}
\usepackage{lipsum}
\usepackage{booktabs}
\usepackage{multirow}
\usepackage{makecell}

\usepackage{subcaption}
\usepackage{stackengine}
\usepackage{tikz}

\usepackage[hidelinks,colorlinks=true]{hyperref}
\hypersetup{
    hidelinks,
    colorlinks=true,
    allcolors = [rgb]{0.02083333333,0.2890625,0.39322916666}
}

\definecolor{customblue}{rgb}{0.02083333333,0.2890625,0.39322916666}
\definecolor{customorange}{rgb}{0.99607843137,0.29803921568,0.00784313725}

\newcommand{\bilevel}{{\sc bilevel}}
\newcommand{\adaptive}{{\sc adaptive}}
\newcommand{\direct}{{\sc direct}}
\newcommand{\thinking}{{\sc thinking}}
\newcommand{\pddl}{{\sc pddl}}
\newcommand{\poses}{{\sc poses}}
\newcommand{\integrated}{{\sc integrated}}
\newcommand{\incremental}{{\sc incremental}}
\newcommand{\focused}{{\sc focused}}
\newcommand{\abstraction}{{\sc model}}

\title{\LARGE \bf 
    A Systematic Study of Large Language Models \\for Task and Motion Planning With PDDLStream
}

\author{Jorge Mendez-Mendez\\Department of Electrical and Computer Engineering\\Stony Brook University\\\texttt{jorge.mendezmendez@stonybrook.edu}}

\begin{document}
\captionsetup[subfigure]{aboveskip=1pt, belowskip=1pt}

\maketitle
\thispagestyle{empty}
\pagestyle{empty}

\begin{abstract}

While we know that large language models (LLMs) can solve some planning problems, we do not understand the extent of these capabilities for robotics. One promising direction is to integrate the semantic knowledge of LLMs with the formal reasoning of task and motion planning (TAMP). However, designing such systems is complicated by the myriad of choices for how to integrate LLMs within TAMP. We develop 16 algorithms that use LLMs to substitute key TAMP components. Our zero-shot experiments across 13\,750 evaluations and three domains reveal that LLM-based planners exhibit lower success rates and higher planning times than engineered systems. Providing geometric details increases the number of task-planning errors compared to pure PDDL descriptions, and (faster) direct LLM variants outperform (slower) reasoning variants in most cases. 
Code and results are available at \href{https://github.com/jorge-a-mendez/llm-pddlstream}{https://github.com/jorge-a-mendez/llm-pddlstream}. 

\end{abstract}

\section{Introduction}

\begin{figure}[t]
    \centering
        \begin{subfigure}[t]{0.32\linewidth}
                \centering
            \includegraphics[height=2.12cm]{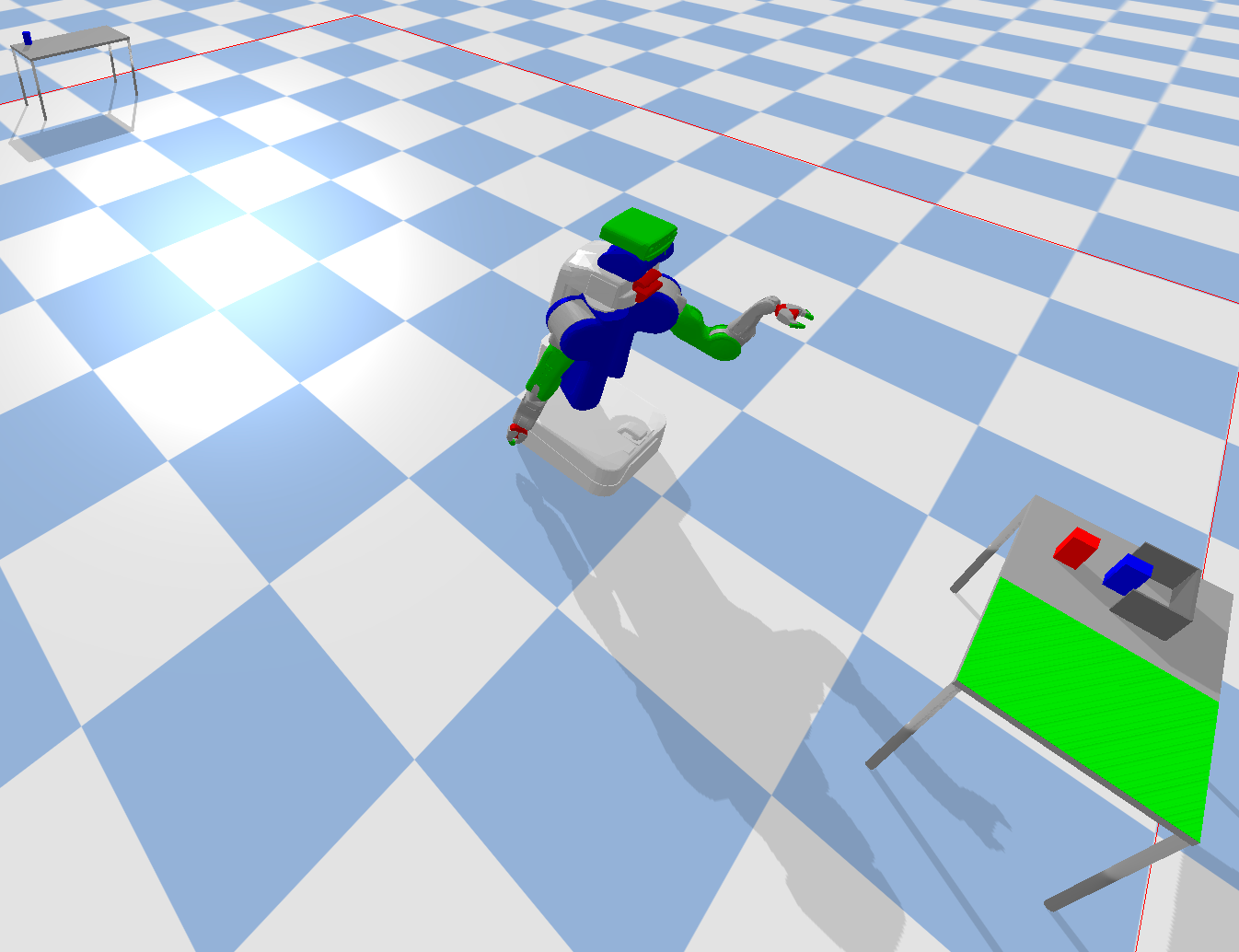}
            \label{fig:blockedDomain}
        \end{subfigure}%
        \begin{subfigure}[t]{0.235\linewidth}
                \centering
            \includegraphics[height=2.12cm]{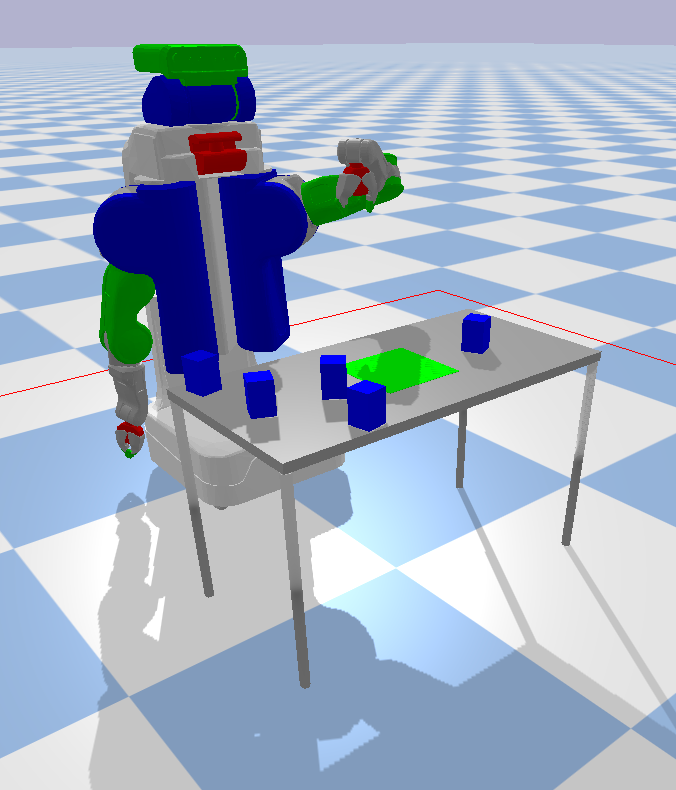}

        \end{subfigure}%
        \begin{subfigure}[t]{0.445\linewidth}
            \centering
            \includegraphics[height=2.12cm]{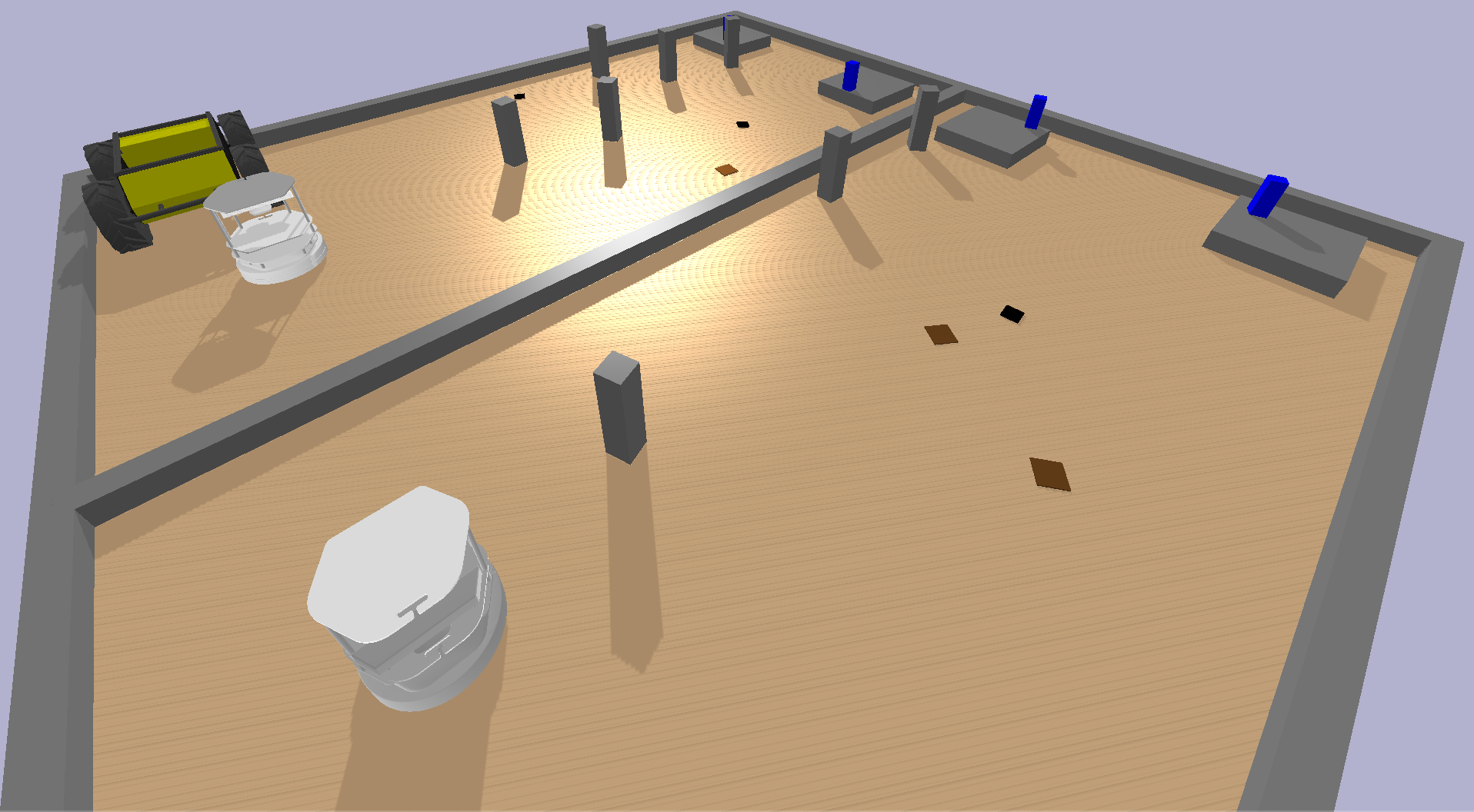}

        \end{subfigure}%
    \vspace{-1em}
    \caption{Evaluation domains~\cite{garrett2020pddlstream}. \textit{Left: Blocked}---place any blue box on green region, but the nearest box is blocked. \textit{Center: Packing}---place all boxes on green region. \textit{Right: Rovers}---use rovers (Turtlebots) to acquire one rock and one soil sample, photograph objectives, and send results to lander (Husky). Obstacles limit visibility and reachability.}
    \label{fig:domains}
\end{figure}

To be maximally helpful, autonomous robots must possess the ability to handle a breadth of long-horizon problems in unstructured environments (e.g., Fig.~\ref{fig:domains}). Hierarchical planning tackles this goal, with many recent efforts leveraging large pretrained models. The promise is that large-scale pretraining may equip models with key robotics capabilities, including building abstractions, finding plans, and generalizing to unseen scenarios. This work evaluates LLMs' ability to produce plans for problems formalized in PDDLStream~\cite{garrett2020pddlstream}.

Large models are promising as elements of robotics solutions---e.g., for perception, domain formalization, abstract planning, or end-to-end low-level planning. We can categorize these choices into making decisions and constructing abstractions---of course, the two may overlap. 
Decision-making variants face a great generalization challenge: they must \textit{efficiently} produce a \textit{correct} plan for each new problem, which may require novel action sequences (e.g., using a broomstick to collect an object under a sofa).
LLM-Modulo approaches that plan with an LLM  within a verification loop are intuitively appealing. This paper provides the first large-scale empirical answer to the question: Is the LLM-Modulo framework viable for TAMP?

To study whether LLM-modulo approaches using current models and TAMP methods generalize to unseen problems, we evaluate 16 LLM-based planners on a large collection of TAMP problems zero-shot. Beyond measuring successes, our study seeks to understand the trade-offs imposed by various design choices (e.g., reasoning or direct generation, geometric or task planning, more or less information in the prompt).

Our planners substitute TAMP components with LLM calls. The TAMP system determines when to execute task- or motion-level planning and verifies plan correctness. For each base TAMP method (\adaptive{}~\cite{garrett2020pddlstream} and \bilevel{}~\cite{kumar2023bilevel}), we create eight LLM planners (four reasoning and four direct), which generate task plans, continuous parameters, or both. Our 13\,750 experiments across three TAMP domains under time-out show that these LLM-based planners can solve many TAMP problems, but have lower success rates and higher planning times than the base planners. Our evaluation dissects the strengths and weaknesses of the LLM-based planners. We also study LLMs to generate TAMP abstractions and analyze how their generated abstractions account for key constraints.

\section{Related Work}

\subsection{Hierarchical Decision-Making for Robotics}

Decomposing problems enables solving longer-horizon tasks. This intuitive idea has been explored, among others, in hierarchical reinforcement learning~\cite{sutton1999between,dietterich2000hierarchical,dayan1993feudal,pateria2021hierarchical}, hierarchical task networks~\cite{sacerdoti1975structure,nau1999shop}, and TAMP~\cite{garrett2021integrated}. TAMP is effective at solving problems on physical robots~\cite{dantam2016incremental,jiang2019task,garrett2020pddlstream,garrett2020online,shen2024differentiable}, searching over \textit{what} actions to execute and \textit{how} to execute them. 
Two major obstacles limit the broad applicability of TAMP: 1)~specifying domains for TAMP requires substantial engineering effort, and 2)~TAMP is typically computationally slow. Recent approaches learn elements of TAMP systems to address these shortcomings~\cite{silver2021learning,silver2022learning,kumar2023learning,kumar2024practice,hedegaard2025beyond}. 

\subsection{Large Models for Planning} 

Other efforts have used large models as decision makers. SayCan uses an LLM to propose actions and ranks them by combined likelihood of usefulness and success~\cite{ahn2022can}. Various extensions have been proposed~\cite{chen2023open,driess2023palme,black2024pi05}, including TAMP-inspired variants~\cite{curtis2024trust,shcherba2025meta}. However, these works do not compare engineered and LLM-based planners. On the other hand, the planning community has compared LLM- and PDDL-based planners, finding that verifying the outputs of an LLM planner by a domain model in a loop (LLM-Modulo) is an effective strategy~\cite{kambhampati2024llms}. Here, we construct LLM-Modulo TAMP approaches and compare them to engineered systems. 

Other works have used large models to build abstractions for robot planning. Liu et al.\ build PDDL problem definitions given a PDDL domain~\cite{liu2023llm}, while Athalye et al.\ discover predicate-level state abstractions for robot planning~\cite{athalye2024pixels}. Our evaluation uses an LLM to generate PDDLStream abstractions and assesses the correctness of the resulting plans.

\subsection{Large Models in Robotics}

Beyond hierarchical decision-making, large vision models (e.g.,~\cite{radford2021learning,kirillov2023segment,ravi2025sam,zhang2023dino,oquab2024dinov}) are central to many modern robotics solutions. Other recent work has focused on end-to-end continuous control policies~\cite{chi2023diffusion,black2024pi0}. While these approaches exhibit an impressive ability to generalize to variations in visual and language cues, they do not by themselves generalize to novel combinations of steps. A recent effort executed a comprehensive evaluation of large pretrained policies finetuned to novel tasks~\cite{barreiros2025careful}. While similar in spirit, their work and ours address distinct questions. 

\section{Background on Task and Motion Planning}

TAMP tackles long-horizon problems with geometric constraints that are difficult to specify in a discrete representation. Many approaches operate at a ``task'' level with abstract states and actions, and a ``motion'' level with continuous states and actions. Critically, TAMP does not assume that the task-level abstraction captures all constraints, and so an integrated approach is needed where the two levels interact.

\subsection{PDDLStream and the Adaptive Algorithm}

PDDLStream augments PDDL with streams that consume inputs $x$ and produce outputs $y$ that satisfy a certificate $\mathtt{cert}(x, y)$. For example, $\mathtt{sample\mbox{-}placement}(\mathtt{surf}, \mathtt{obj})\!\rightarrow$ $\mathtt{pose}$ certifies $\mathtt{isStable}(\mathtt{pose}, \mathtt{surf}, \mathtt{obj})$. 
Special test streams produce no outputs but certify inputs, such as $\mathtt{test\mbox{-}cfree}(\mathtt{obj1}, \mathtt{pose1}, \mathtt{obj2}, \mathtt{pose2}) \rightarrow [\ ]$ to certify $\mathtt{cFree(obj1, pose1, obj2, pose2)}$. A stream fails if it cannot satisfy its certificates. Our LLM planners operate on PDDLStream problem descriptions. 

A PDDLStream solution is a sequence of actions and continuous values that satisfies all consecutive action preconditions and the goal. The \incremental{} approach first generates stream outputs (poses, trajectories) and uses them as PDDL objects, iteratively increasing the maximum depth $l$ of allowed stream evaluations. \incremental{} is probabilistically complete but may spend too much time evaluating streams (e.g., $\mathtt{plan\mbox{-}motion}$) that do not contribute to the goal-achieving plan. \focused{} performs PDDL planning on ``optimistic'' stream outputs up to level $l$ and evaluates only streams that support the candidate plan. \focused{} often outperforms \incremental{} in cases where PDDL planning is faster than stream evaluation. \adaptive{}, the base for half of our LLM-based planners, balances time between stream evaluation and PDDL planning, bridging between the two methods and maintaining probabilistic completeness. 

\subsection{The Bilevel Planning Algorithm}

\bilevel{} planning greedily assumes that PDDL plans may be refined into motion plans~\cite{kumar2023bilevel}. \bilevel{} finds a PDDL plan and uses backtracking search to find the continuous values, calling PDDL again upon failure. Rejecting a PDDL plan via backtracking may take exponential time in the plan length, so \bilevel{} is fastest when 1)~most PDDL plans can be refined and 2)~backtracking fails early in the plan. We use \bilevel{} as the base for the second half of our LLM-based planners. 

Our version of \bilevel{} uses PDDLStream descriptions instead of NSRTs~\cite{silver2022learning}, which does not affect execution. We perform backtracking search in the order of streams used by \adaptive{}, which enforces input-output dependency ordering instead of temporal ordering and defers more expensive streams. This maintains theoretical guarantees but may reduce planning time. Following Silver et al.~\cite{silver2022learning}, backtracking failures trigger the addition of a precondition to the failed action that requires moving objects that cause the failure.

\section{Problem Setting}

We consider an LLM-Modulo setting~\cite{kambhampati2024llms}, where the LLM produces candidate solutions and an engineered system verifies their validity, reprompting the LLM upon failure. Our verifier is a probabilistically complete TAMP system based on PDDLStream (e.g., \adaptive{} or \bilevel{}). This requires access to a formal definition of the problem in PDDLStream. 

Alternatively, the LLM could directly produce plans or dictate the TAMP algorithm for alternating between task planning and stream evaluation. However, we consider a setting that provides additional guidance to the LLM to increase its chances of success. The TAMP algorithm guides the search, and the LLM replaces individual components of the system: the task planner or stream evaluator. 

We are interested in the generalization ability of LLMs to completely new problems. While we have no guarantees that the LLM did not encounter similar problems in its training data, we prevent further leakage by providing zero examples of solved problems to the LLM---i.e., we evaluate its zero-shot planning ability. We consider the LLM planner successful if, within the allotted time, it produces a sequence of actions and stream outputs that achieves the PDDLStream goal.

We also consider an alternate setting where the LLM produces a PDDLStream formalization. In this setting, we provide one in-context example and request a formalization of a novel domain. A TAMP system plans in the LLM-designed abstraction and succeeds if it produces a plan that is valid according to the ground-truth PDDLStream problem. 

\section{Large Language Model Planners}
\label{sec:LLMPlanners}

For each planner, we choose one option for each dimension discussed below, resulting in 16 LLM-based planners (Fig.~\ref{fig:LLMPlanners}). 

\subsection{Base Algorithm}

Each LLM-based planner uses either \adaptive{} or \bilevel{} as the base algorithm. The two methods differ in their expectations of the PDDL planner. \bilevel{} assumes that most PDDL plans are valid, and so it expends significant computation searching for stream evaluations that support a given PDDL plan. \adaptive{} instead uses PDDL planning to determine which streams to evaluate but does not assume that any optimistic plan may succeed. \adaptive{} incrementally constructs a set of stream outputs, which PDDL planning uses to produce plans that are guaranteed to be valid. 

With engineered PDDL planners (e.g., Fast Downward), \adaptive{} is often faster. Reasons include that PDDL planning is very fast, making it cheap to generate many candidate plans, and that PDDL planners are not aware of scene geometry, so they cannot assess geometric constraints (e.g., if an object fits in a container, or if an obstacle must be moved before grasping an object). While LLMs are slower at PDDL planning, they could in principle leverage information about scene geometry to produce plans that satisfy geometric constraints. However, we do not observe this geometry-aware PDDL planning behavior in our experiments.

\begin{figure}
    \centering
    \begin{tikzpicture}[every node/.style={minimum size=1cm,font=\scriptsize, text=white}]
    \clip[use as bounding box] (-1.5, -1) rectangle (7.5, 3.11);
    \begin{scope}[every node/.append style={yslant=-0.62},yslant=-0.62]
        \fill[customblue!100,opacity=0.6] (-1, 0) rectangle (1.6, 1.25);
        \fill[customblue!100,opacity=0.8] (-1, 0) rectangle (-0.6, 1.25);
        \fill[customorange!100,opacity=0.6] (-1, 1.25) rectangle (1.6, 2.5);
        \fill[customorange!100,opacity=0.8] (-1, 1.25) rectangle (-0.6, 2.5);
        \node[rotate=90] at (-0.8,1.875) {\adaptive{}};
        \node at (-0.15,1.875) {\direct{}};
        \node at (0.95,1.875) {\thinking{}};
        \node[rotate=90] at (-0.8,0.625) {\bilevel{}};
        \node at (-0.15,0.625) {\direct{}};
        \node at (0.95,0.625) {\thinking{}};

        \draw (-1, 0) -- (-1, 2.5);
        \draw (-0.6, 0) -- (-0.6, 2.5);
        \draw (0.35, 0) -- (0.35, 2.5);
        \draw (1.6, 0) -- (1.6, 2.5);
        \draw (-1, 0) -- (1.6, 0);
        \draw (-1, 1.25) -- (1.6, 1.25);
        \draw (-1, 2.5) -- (1.6, 2.5);
    \end{scope}
    \begin{scope}[every node/.append style={yslant=0.38},yslant=0.38]
        \fill[customblue!100,opacity=0.6] (1.6, -1.6) rectangle (5.8, -0.35);
        \fill[customorange!100,opacity=0.6] (1.6, -0.35) rectangle (5.8, 0.9);
        \node at (2.,0.25) {\pddl{}};
        \node at (2.8,0.25) {\poses{}};
        \node[text width=1.2cm, align=center] at (3.7,0.25) {\pddl{}+\\\poses{}};
        \node at (5.,0.25) {\integrated{}};
        \node at (2.,-1.) {\pddl{}};
        \node at (2.8,-1.) {\poses{}};
        \node[text width=1.2cm, align=center] at (3.7,-1.) {\pddl{}+\\\poses{}};
        \node at (5.,-1.) {\integrated{}};

        \draw (2.4, -1.6) -- (2.4, 0.9);
        \draw (3.25, -1.6) -- (3.25, 0.9);
        \draw (4.2, -1.6) -- (4.2, 0.9);
        \draw (5.8, -1.6) -- (5.8, 0.9);
        \draw (1.6, -1.6) -- (5.8, -1.6);
        \draw (1.6, -0.35) -- (5.8, -0.35);
        \draw (1.6, 0.9) -- (5.8, 0.9);
    \end{scope}
    \begin{scope}[every node/.append style={
        yslant=0.38,xslant=-1},yslant=0.38,xslant=-1
      ]
        \fill[customorange!100,opacity=0.6] (2.5, 0.9) rectangle (6.7, 3.5);
        \fill[customorange!100,opacity=0.8] (2.5, 3.1) rectangle (6.7, 3.5);

        \draw (2.5, 3.5) -- (6.7, 3.5);
        \draw (2.5, 3.1) -- (6.7, 3.1);
        \draw (2.5, 2.15) -- (6.7, 2.15);

        \draw (3.3, 3.5) -- (3.3, 0.9);
        \draw (4.15, 3.5) -- (4.15, 0.9);
        \draw (5.1, 3.5) -- (5.1, 0.9);
        \draw (6.7, 3.5) -- (6.7, 0.9);
    \end{scope}
    \end{tikzpicture}    
    \caption{Summary of the 16 LLM-based planners.}
    \label{fig:LLMPlanners}
\end{figure}

\subsection{Component to Substitute}

The key technical advancement required to construct the LLM-based planners is designing the interaction between the TAMP system and the LLM. This section discusses the roles of the LLM in our planners and our prompting strategies. 

\subsubsection{\pddl{} Planning} One way to use an LLM within TAMP is to query the LLM for PDDL plans. While the LLM could in principle produce plans that account for geometric constraints, our early tests suggested that including geometric details instead harms performance: the LLM does not consider the geometric constraints and it makes more formatting and PDDL errors. In consequence, our prompts omit geometric details. 

\paragraph{System prompt} We instruct the LLM to act as a PDDL planner and provide the domain name and the sets of predicates, axioms, and actions. We then explain what each problem's prompt will include, clarify the closed-world setting (only facts in the initial state are true), and emphasize that preconditions must be satisfied. We request that each response include a list of actions and provide the format. 

\paragraph{Initial prompt} We provide the LLM with the list of objects, the initial PDDL state, and the PDDL goal.  

\paragraph{Subsequent prompts} We provide details about any previous plan failures, including PDDL (e.g., missing preconditions) and stream (e.g., no collision-free grasps) failures. 

\subsubsection{\poses{} Stream Evaluation} Producing continuous values that satisfy geometric constraints is often more difficult than finding the PDDL sequence of actions. For example, consider the Packing problem in Fig.~\ref{fig:domains}, where multiple boxes must be placed on a small surface. \adaptive{} and \bilevel{} use geometric samplers to produce poses individually for each object. This often requires producing many candidate samples for each object until a combination is found that avoids collisions. Our \poses{} LLM evaluates sampling streams for object placements (for the Packing and Blocked domains) and robot base configurations (for the Rovers domain). The TAMP system prompts the LLM to produce poses that are consistent with the \textit{fixed} scene geometry. Note that, when producing individual samples, it is not possible to account for movable objects, since their poses may be different by the time the action is executed. 

\paragraph{System prompt} We instruct the LLM to reason geometrically to provide stable object placements or base configurations with line of sight to a target. We explain that multiple requests may be made, which possibly indicates that a previous attempt resulted in collision or occlusion. The line-of-sight sampler is given the geometry of fixed obstacles, while the placement sampler is told to avoid collisions with objects it placed in the past (there are no fixed obstacles). We state that we will provide past examples of failure and success, clarifying that a success requires only satisfying local constraints (e.g., no collisions with fixed obstacles), but that such samples may still violate global constraints involving movable objects. We explain the content of each future prompt and specify the format for the responses.

\paragraph{Initial prompt} For placements, we list possible surfaces and the body dimensions (as 2D axis-aligned bounding boxes). For line-of-sight, we include the target point and scene bounds. 

\paragraph{Subsequent prompts} We list past successful and failed samples, with explanations for failures (e.g., collision). 

\subsubsection{\integrated{} PDDL Planning and Stream Evaluation} To reason about both PDDL and geometry, we could use a separate LLM instance for each purpose (\pddl{}+\poses{}). However, this misses the opportunity to produce action sequences that take into account the geometry of the scene, or samples that account for global constraints imposed by the action sequence. Our \integrated{} planner jointly produces an action sequence and samples poses that support the plan to satisfy both global geometric and logical constraints. 

\paragraph{System prompt} Beyond the \pddl{} and \poses{} prompts, we instruct the LLM to act as an \integrated{} TAMP system. We explain how to compute action costs and request to minimize sum cost---this is not possible for \pddl{}-only variants because action costs are sensitive to geometry.

\paragraph{Initial prompt} We add a description of every geometric object in the PDDLStream problem, including physical objects and previously sampled objects (e.g., poses, trajectories). Geometric descriptions enable the integrated planner to reason jointly about logical and geometric constraints.

\paragraph{Subsequent prompts} Identical to \pddl{} and \poses{}. 

\subsection{Thinking Budget}

Models trained to ``think'' exhibit improved reasoning capabilities, in part because they can expend additional computation to tackle more computationally complex problems. At first glance, the reasoning capabilities required to generalize to unseen TAMP problems would suggest that a high thinking budget would yield a stronger TAMP system.

However, our goal is to plan both \textit{effectively} and \textit{efficiently}, using the LLM as only part of a broader planning system. In consequence, identifying the optimal trade-off between plan correctness and computation time requires empirical evidence. For example, a non-thinking model might produce many invalid plans and one valid plan faster than a thinking model produces one valid plan. We construct both \thinking{} and \direct{} variants of our LLM-based planners.

\subsection{\abstraction{} Design} One distinct mechanism to leverage LLMs in TAMP is for abstraction design. We study this setting by having the LLM produce a PDDLStream domain and problem definition and then directly using an engineered TAMP system for planning.

\paragraph{System prompt} We instruct the LLM to act as a PDDLStream designer, describe the PDDLStream syntax and semantics, and provide one in-context example. We request a domain file, a stream file, a mapping from streams to (given) Python functions, the initial state atoms, and the goal atoms. 

\paragraph{Initial prompt} We describe the problem, the object geometries, and the Python functions available for the streams.

\paragraph{Subsequent prompts} We list past failures in parsing the abstraction or planning with the abstraction.

\begin{figure*}[p]
    \centering
    \begin{subfigure}[b]{0.36\textwidth}
        \raisebox{0.2cm}{\rotatebox{90}{\tiny Gemini 2.5 Flash}}~
        \includegraphics[trim={1.cm 2.75cm 1cm 1.5cm},clip,height=1.85cm]{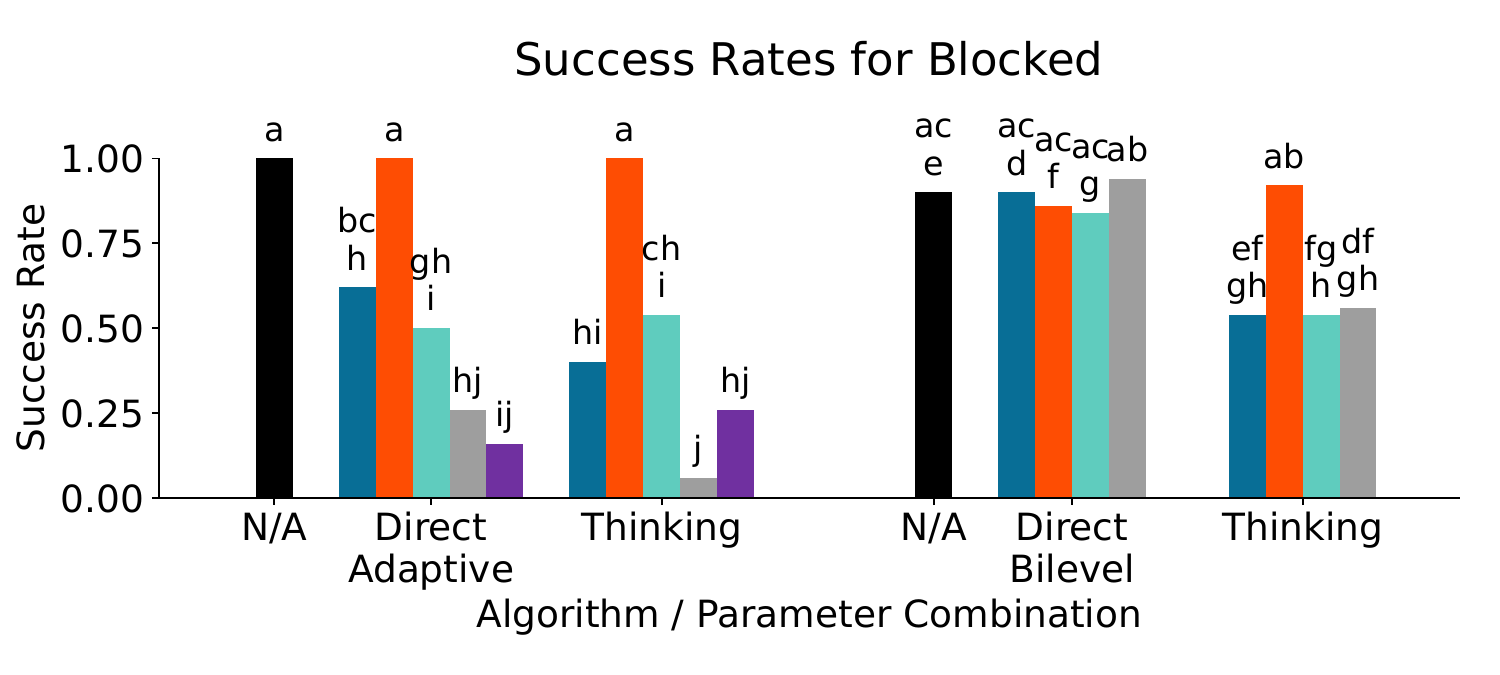}
        \raisebox{0.3cm}{\rotatebox{90}{\tiny Gemini 3 Flash}}~
        \includegraphics[trim={1.cm 2.75cm 1cm 1.5cm},clip,height=1.85cm]{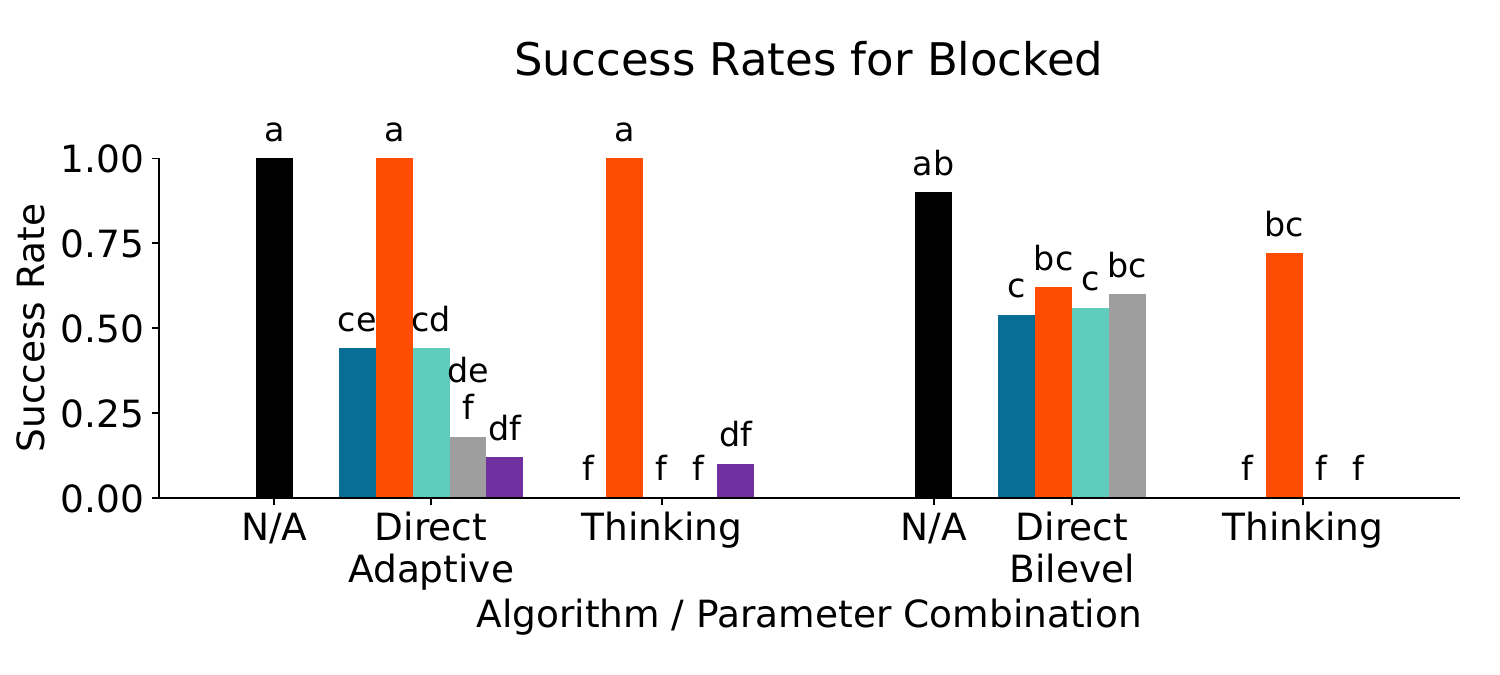}
        \raisebox{0.8cm}{\rotatebox{90}{\tiny GPT-5 mini}}~~
        \includegraphics[trim={1.cm 1.3cm 1cm 1.5cm},clip,height=2.25cm]{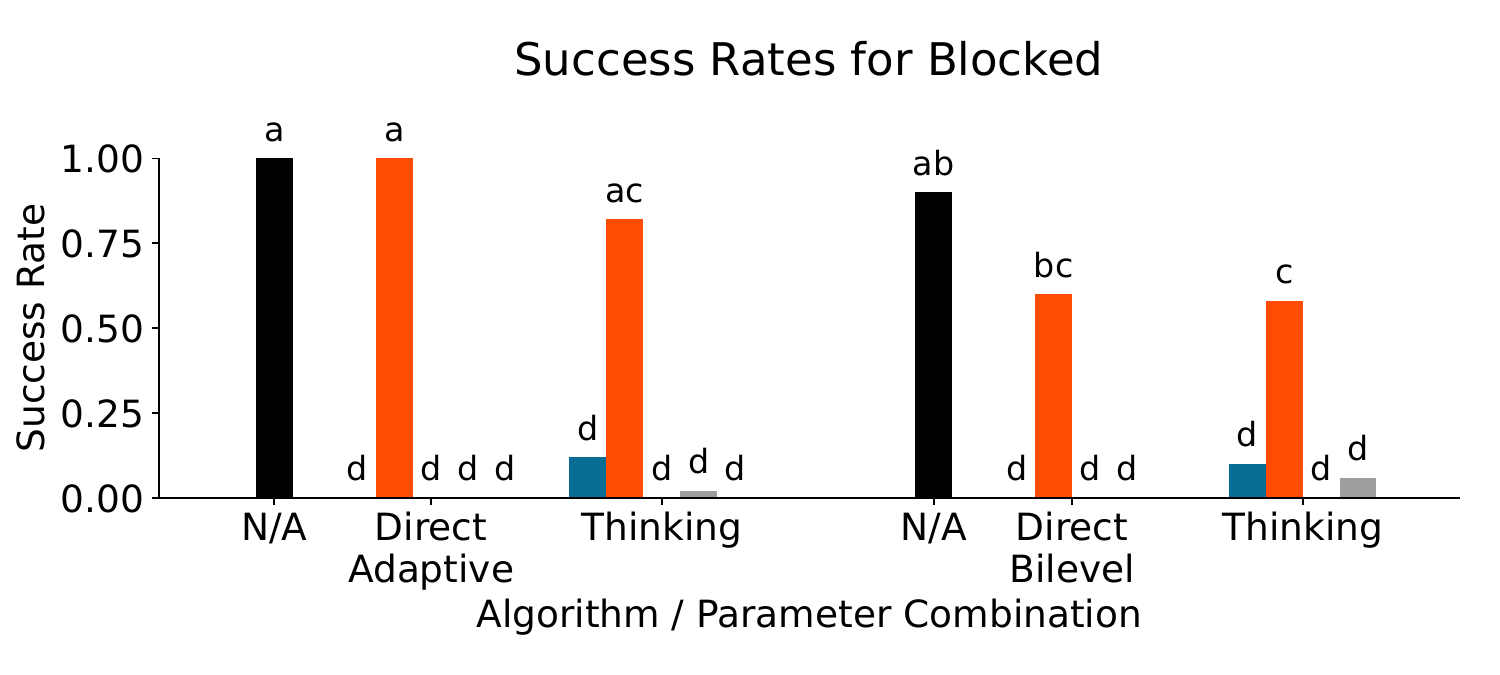}
        \caption{Blocked}
    \end{subfigure}\hfill
    \begin{subfigure}[b]{0.319\textwidth}
        \includegraphics[trim={2.5cm 2.75cm 1cm 1.5cm},clip,height=1.85cm]{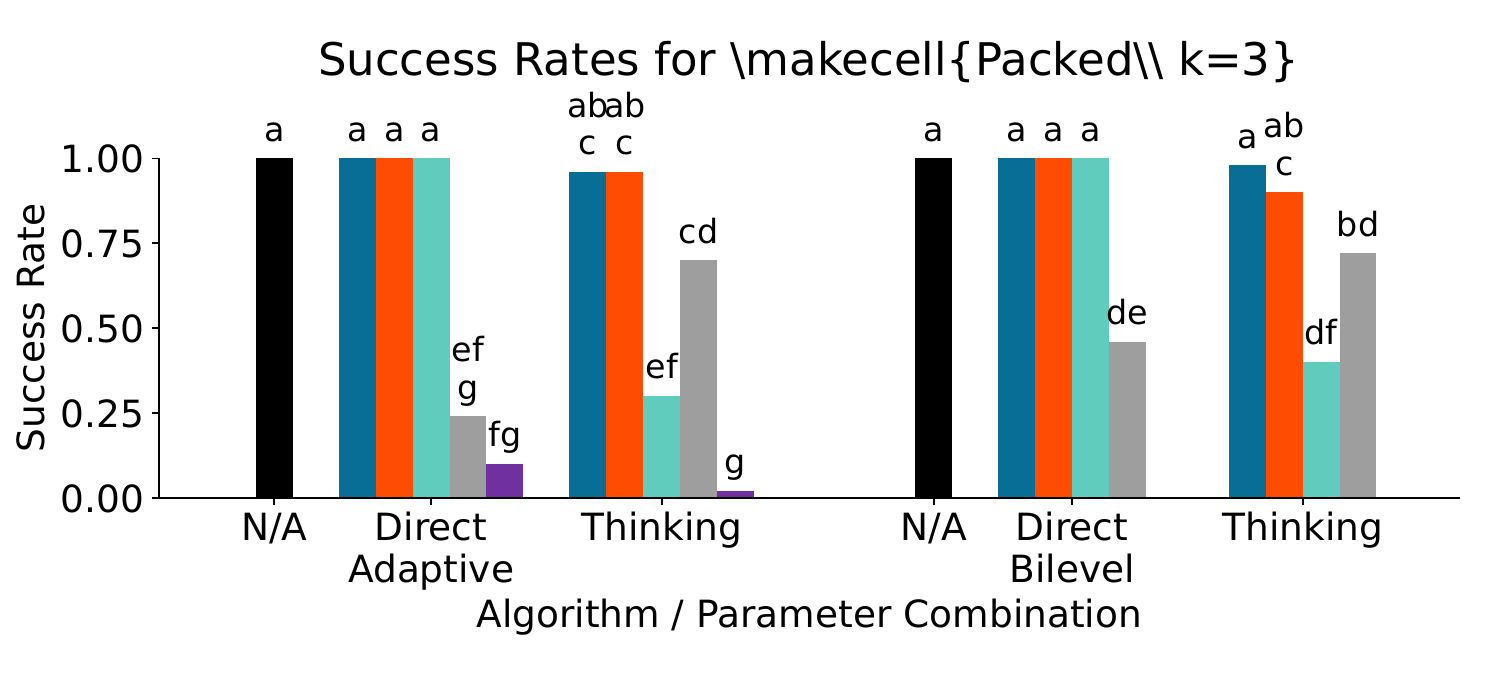}
        \includegraphics[trim={2.5cm 2.75cm 1cm 1.5cm},clip,height=1.85cm]{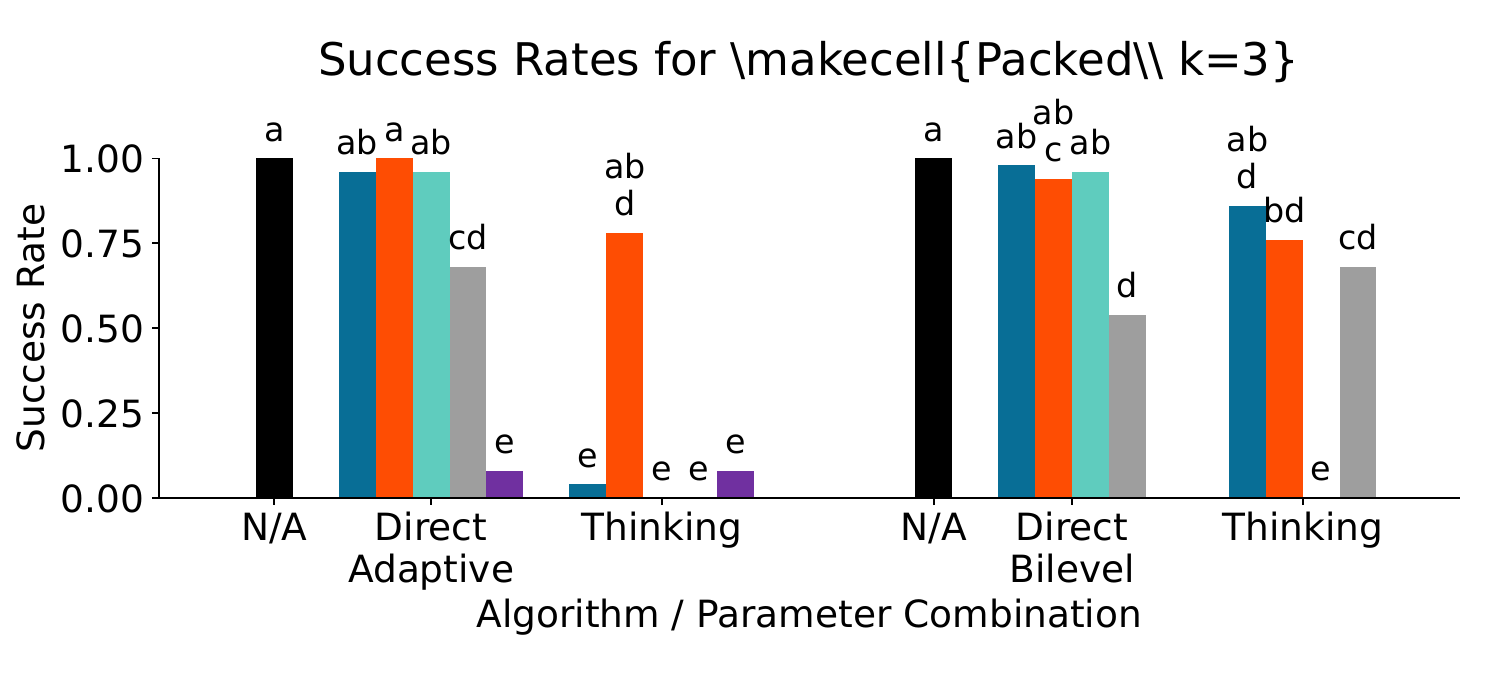}
        \includegraphics[trim={2.5cm 1.3cm 1cm 1.5cm},clip,height=2.25cm]{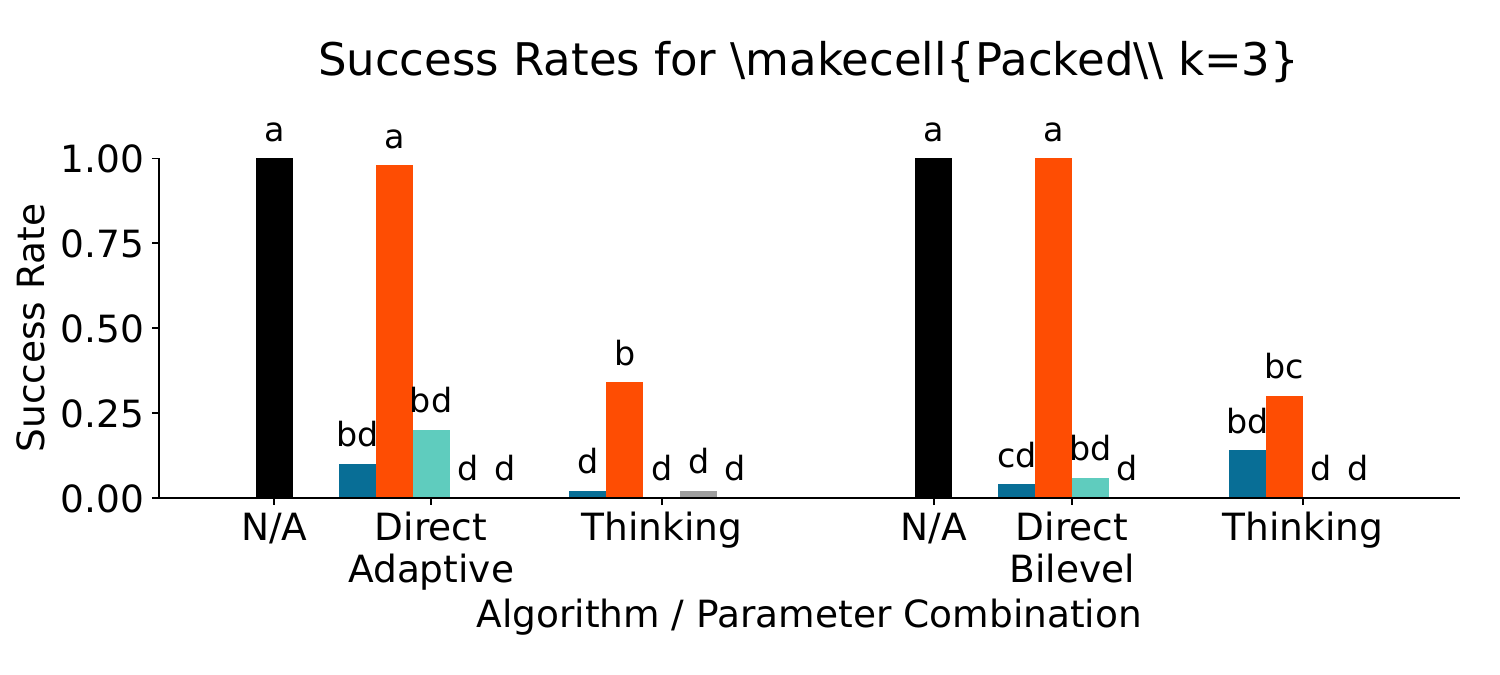}
        \caption{Packing, $k=3$}
    \end{subfigure}\hfill
    \begin{subfigure}[b]{0.319\textwidth}
        \includegraphics[trim={2.5cm 2.75cm 1cm 1.5cm},clip,height=1.85cm]{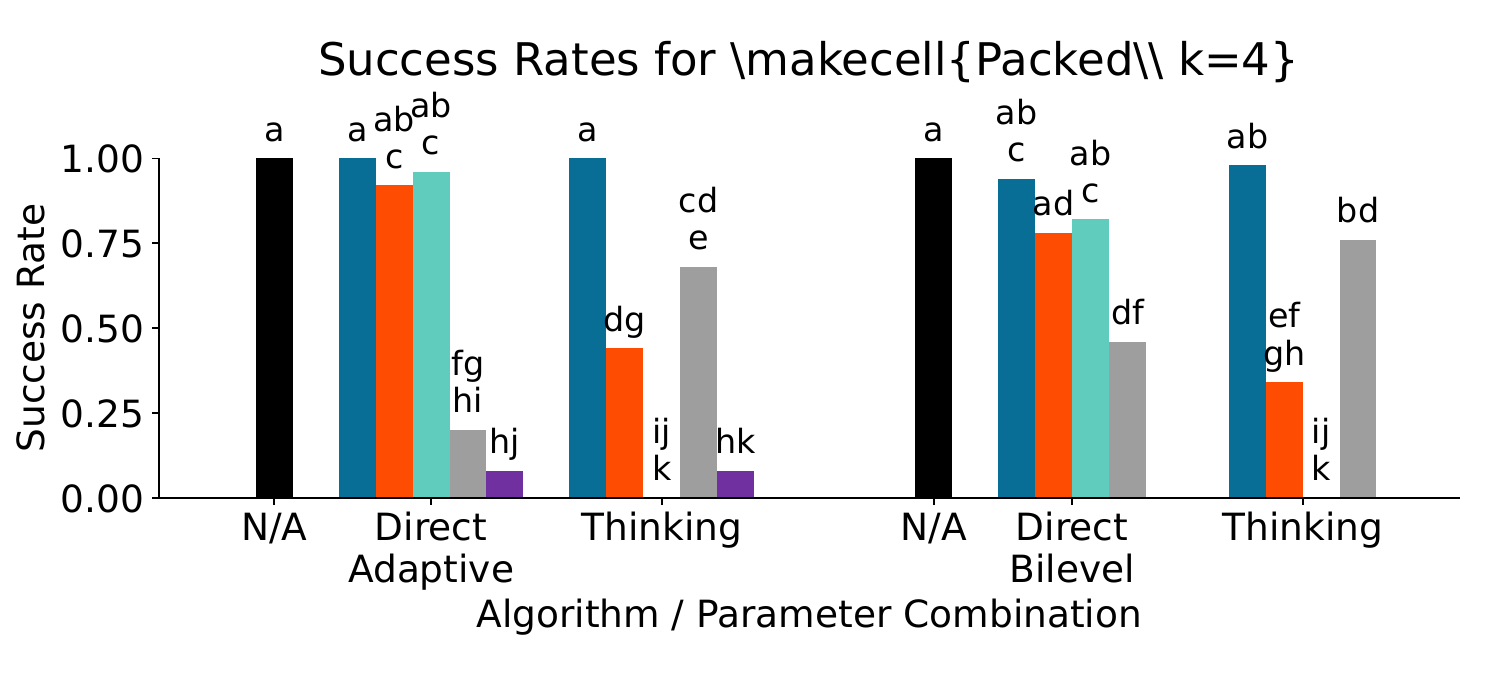}
        \includegraphics[trim={2.5cm 2.75cm 1cm 1.5cm},clip,height=1.85cm]{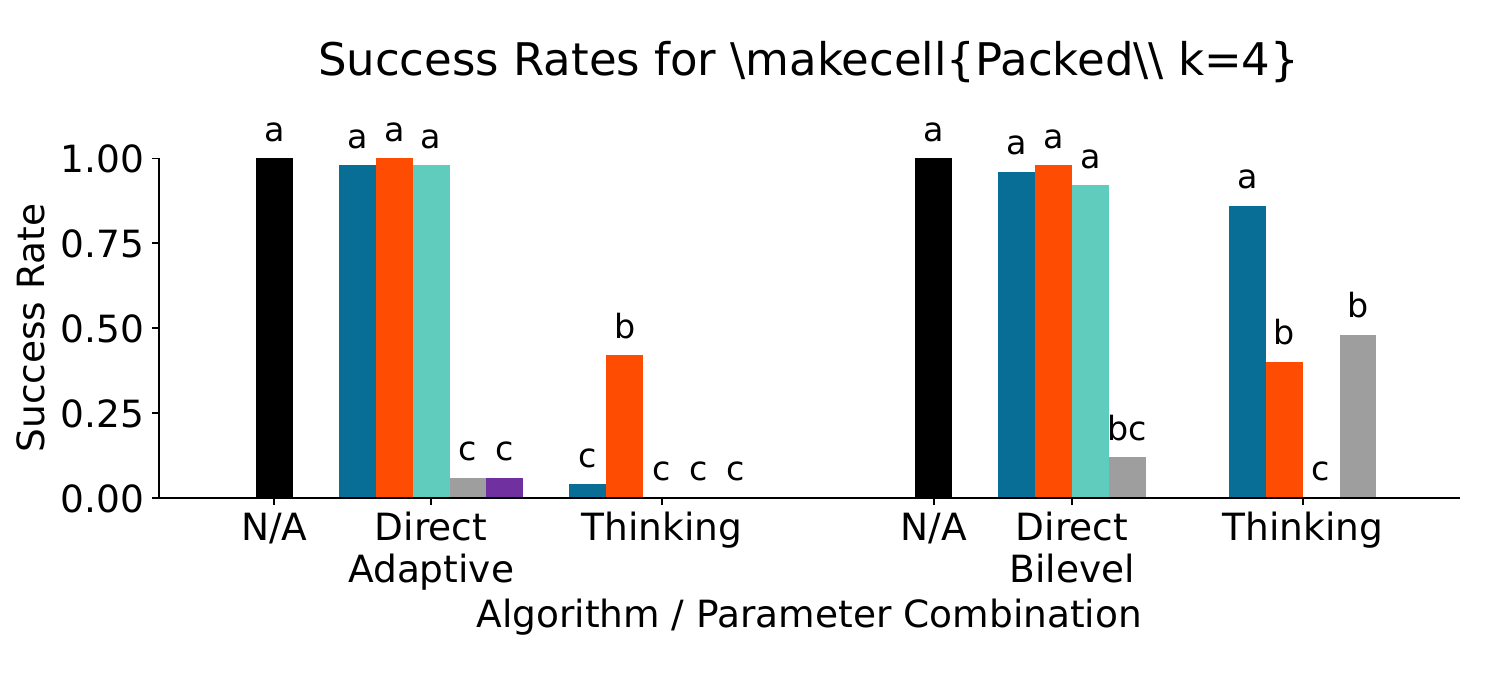}
        \includegraphics[trim={2.5cm 1.3cm 1cm 1.5cm},clip,height=2.25cm]{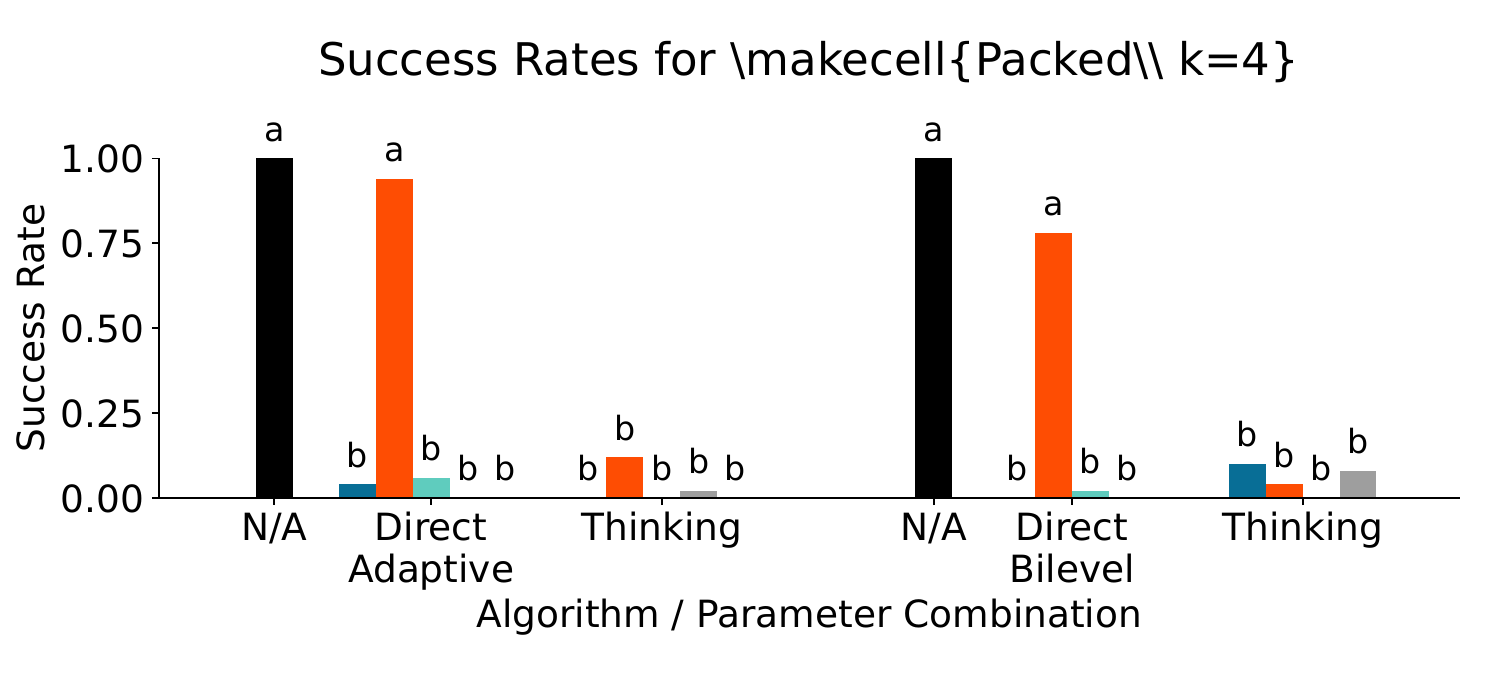}
        \caption{Packing, $k=4$}
    \end{subfigure}
    \par\medskip
    \mbox{%
        \begin{subfigure}[b]{0.36\textwidth}
            \raisebox{0.2cm}{\rotatebox{90}{\tiny Gemini 2.5 Flash}}~
            \includegraphics[trim={1.cm 2.75cm 1cm 1.5cm},clip,height=1.85cm]{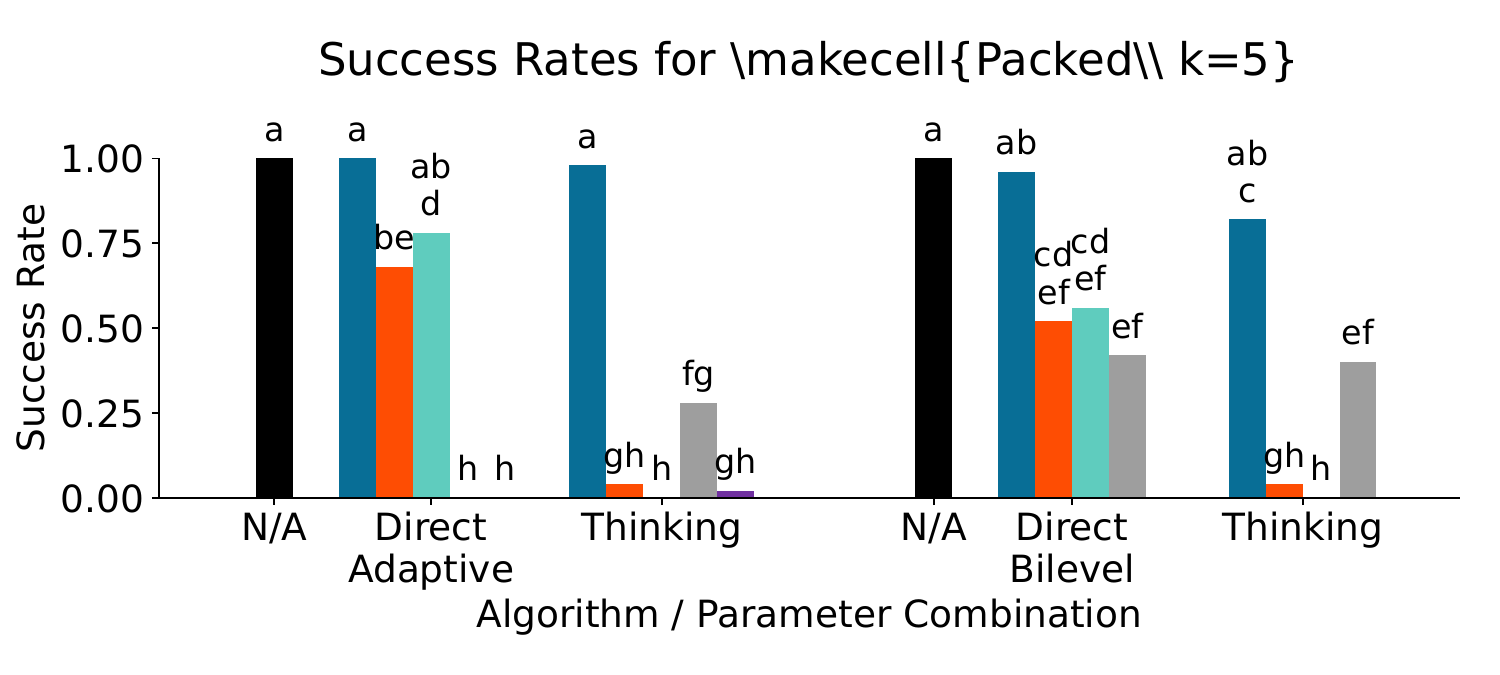}
            \raisebox{0.3cm}{\rotatebox{90}{\tiny Gemini 3 Flash}}~
            \includegraphics[trim={1.cm 2.75cm 1cm 1.5cm},clip,height=1.85cm]{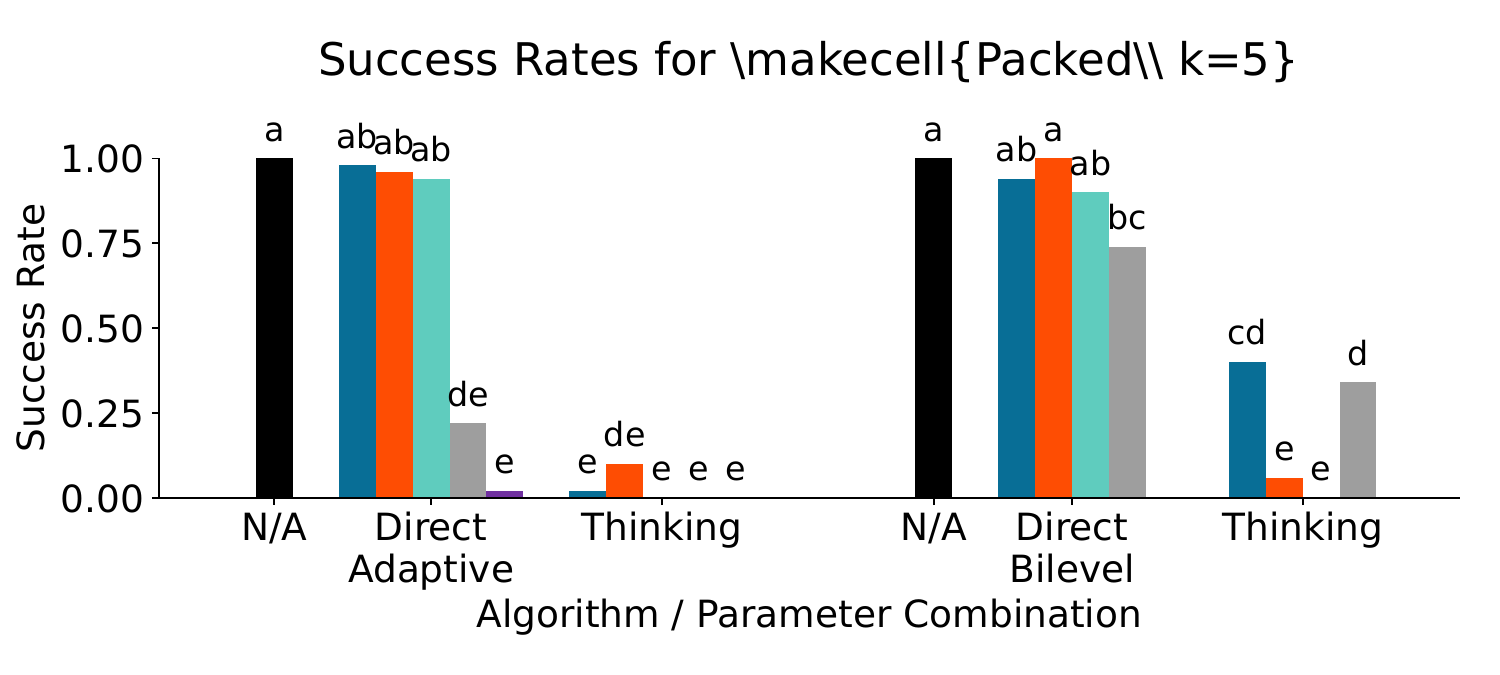}
            \raisebox{0.8cm}{\rotatebox{90}{\tiny GPT-5 mini}}~~
            \includegraphics[trim={1.cm 1.3cm 1cm 1.5cm},clip,height=2.25cm]{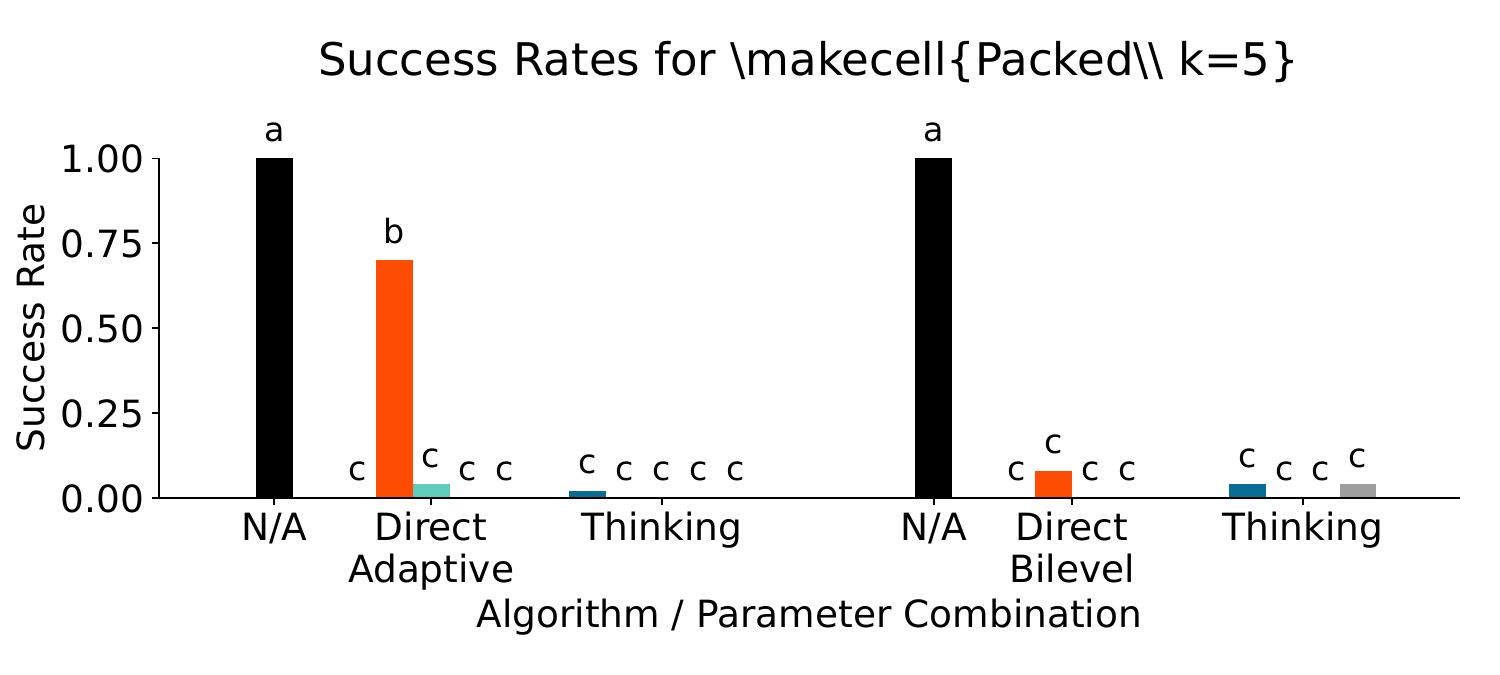}
            \caption{Packing, $k=5$}
        \end{subfigure}\hfill
        \begin{subfigure}[b]{0.182\textwidth}
            \includegraphics[trim={2.5cm 2.75cm 12cm 1.5cm},clip,height=1.85cm]{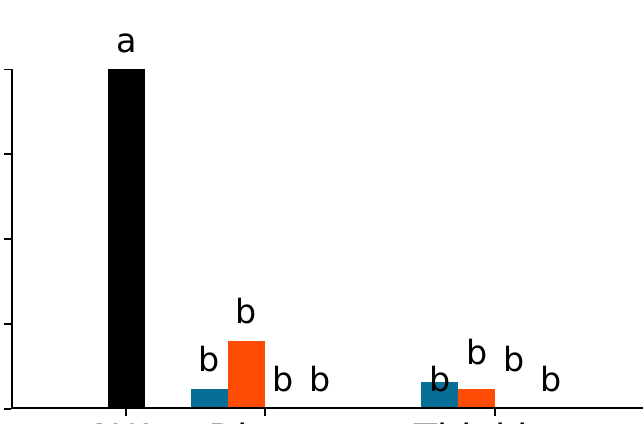}
            \includegraphics[trim={2.5cm 2.75cm 12cm 1.5cm},clip,height=1.85cm]{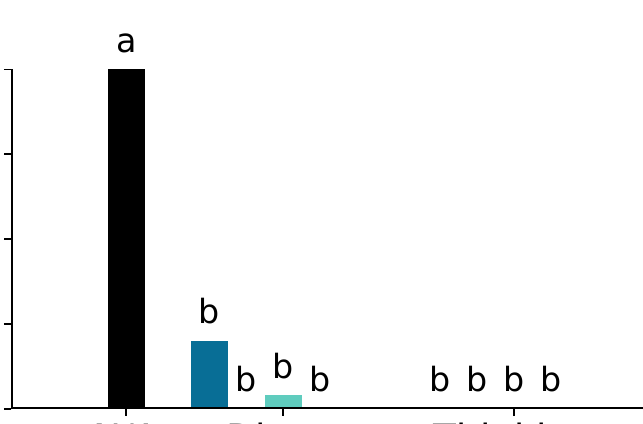}
            \includegraphics[trim={2.5cm 1.3cm 12cm 1.5cm},clip,height=2.25cm]{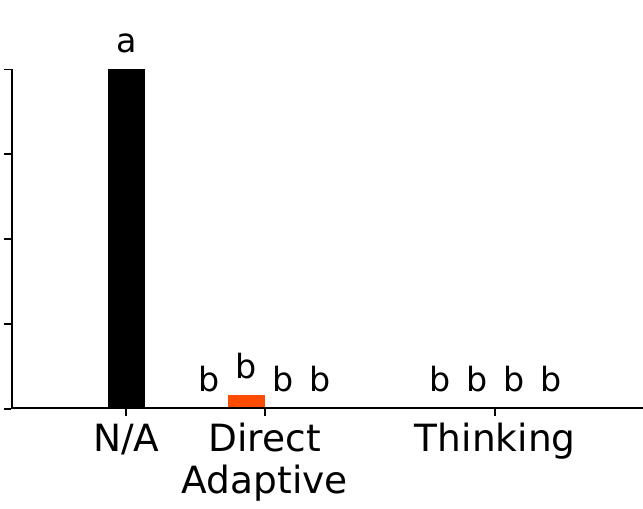}
            \caption{Rovers, $k=2$}
        \end{subfigure}\hfill
        \begin{subfigure}[b]{0.182\textwidth}
            \includegraphics[trim={2.5cm 2.75cm 12cm 1.5cm},clip,height=1.85cm]{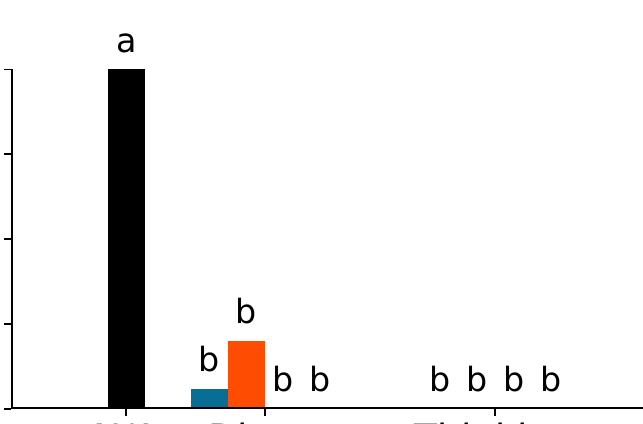}
            \includegraphics[trim={2.5cm 2.75cm 12cm 1.5cm},clip,height=1.85cm]{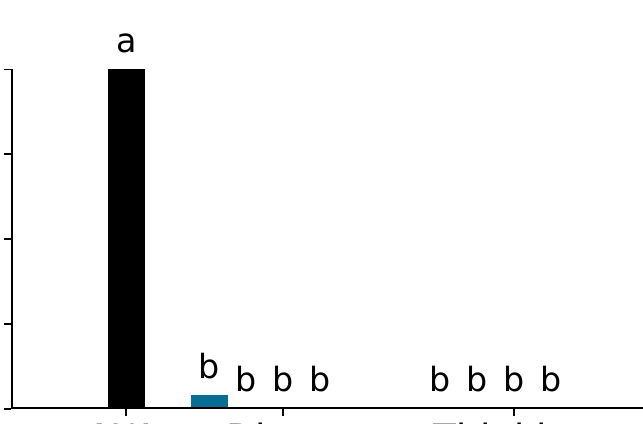}
            \includegraphics[trim={2.5cm 1.3cm 12cm 1.5cm},clip,height=2.25cm]{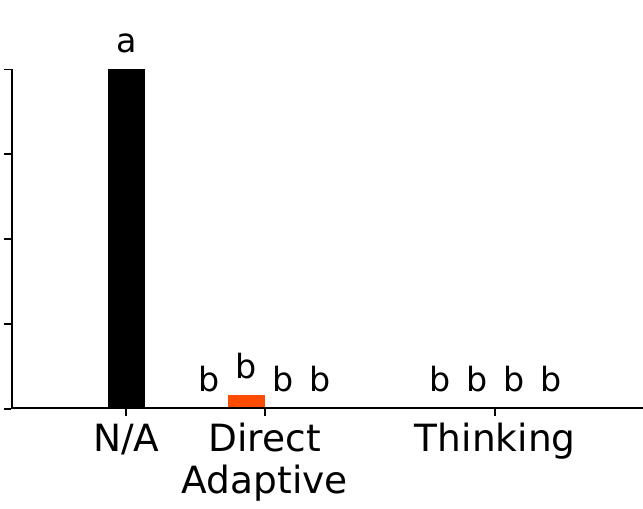}
            \caption{Rovers, $k=3$}
        \end{subfigure}\hfill
        \begin{subfigure}[b]{0.182\textwidth}
            \includegraphics[trim={2.5cm 2.75cm 12cm 1.5cm},clip,height=1.85cm]{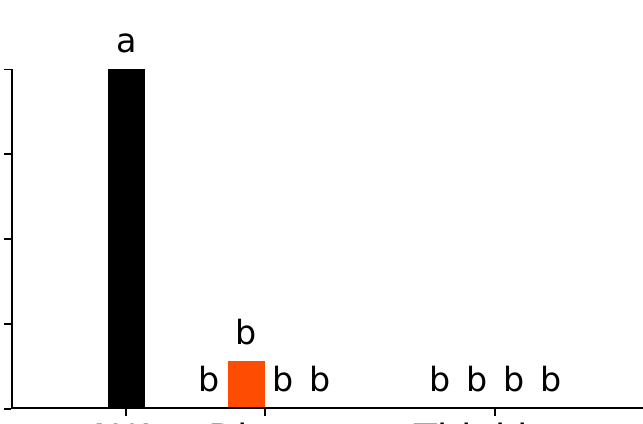}
            \includegraphics[trim={2.5cm 2.75cm 12cm 1.5cm},clip,height=1.85cm]{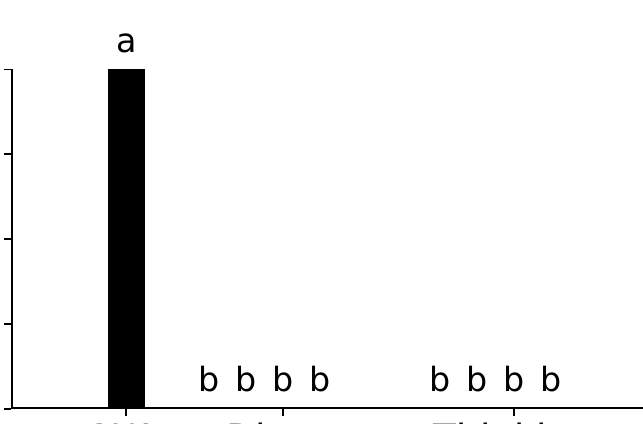}
            \includegraphics[trim={2.5cm 1.3cm 12cm 1.5cm},clip,height=2.25cm]{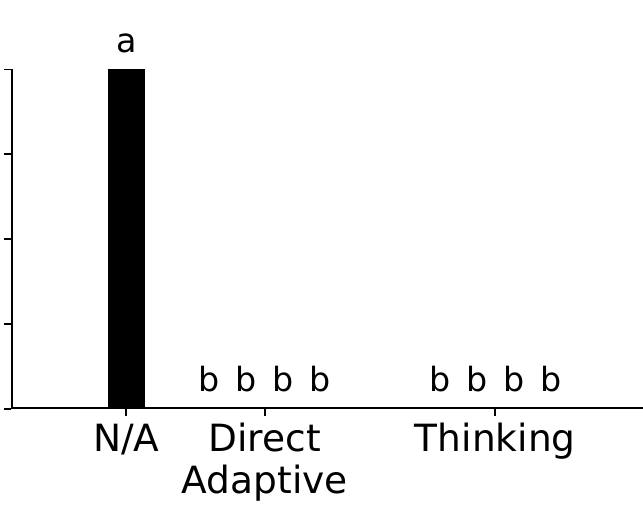}
            \caption{Rovers, $k=4$}
        \end{subfigure}
        \begin{subfigure}[b]{0.09\textwidth}
            \includegraphics[trim={18.cm 2.8cm 1.cm 0.cm},clip,height=2.25cm]{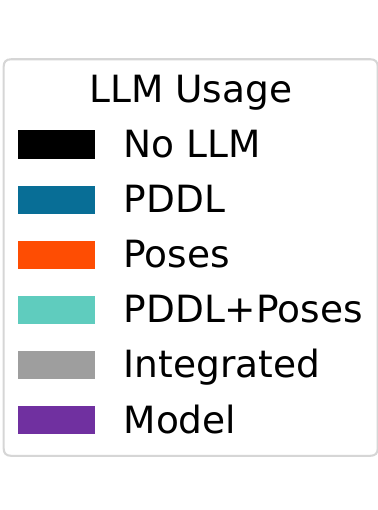}
            \caption*{\ }
        \end{subfigure}%
    }
    \vspace{-0.3em}
    \caption{Success rate across 50 problems per algorithm and domain (compact letter display atop bars). When considering all domains, non-LLM methods achieve higher success rates than the LLM-based variants. \integrated{} and \abstraction{} approaches achieve the lowest success rates, and all LLM-based methods solve a small portion of the problems in Rovers domains.}
    \label{fig:successRate}

\vspace{1.0em}
    \begin{subfigure}[b]{\linewidth}
        \centering
        \includegraphics[trim={0.cm 1.5cm 0cm 1.5cm},clip,height=0.5cm]{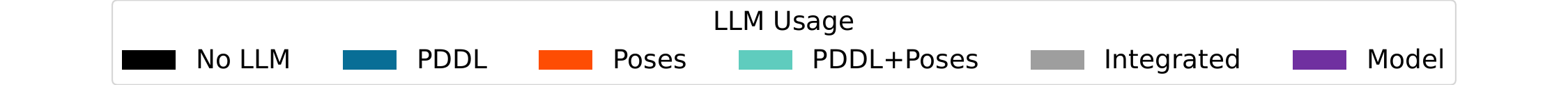}
    \end{subfigure}\\
    \vspace{-0.8em}
    \begin{subfigure}[b]{0.37\textwidth}
        \raisebox{0.5cm}{\rotatebox{90}{\tiny Blocked}}~
        \includegraphics[trim={1.3cm 2.75cm 1cm 1.5cm},clip,height=1.85cm]{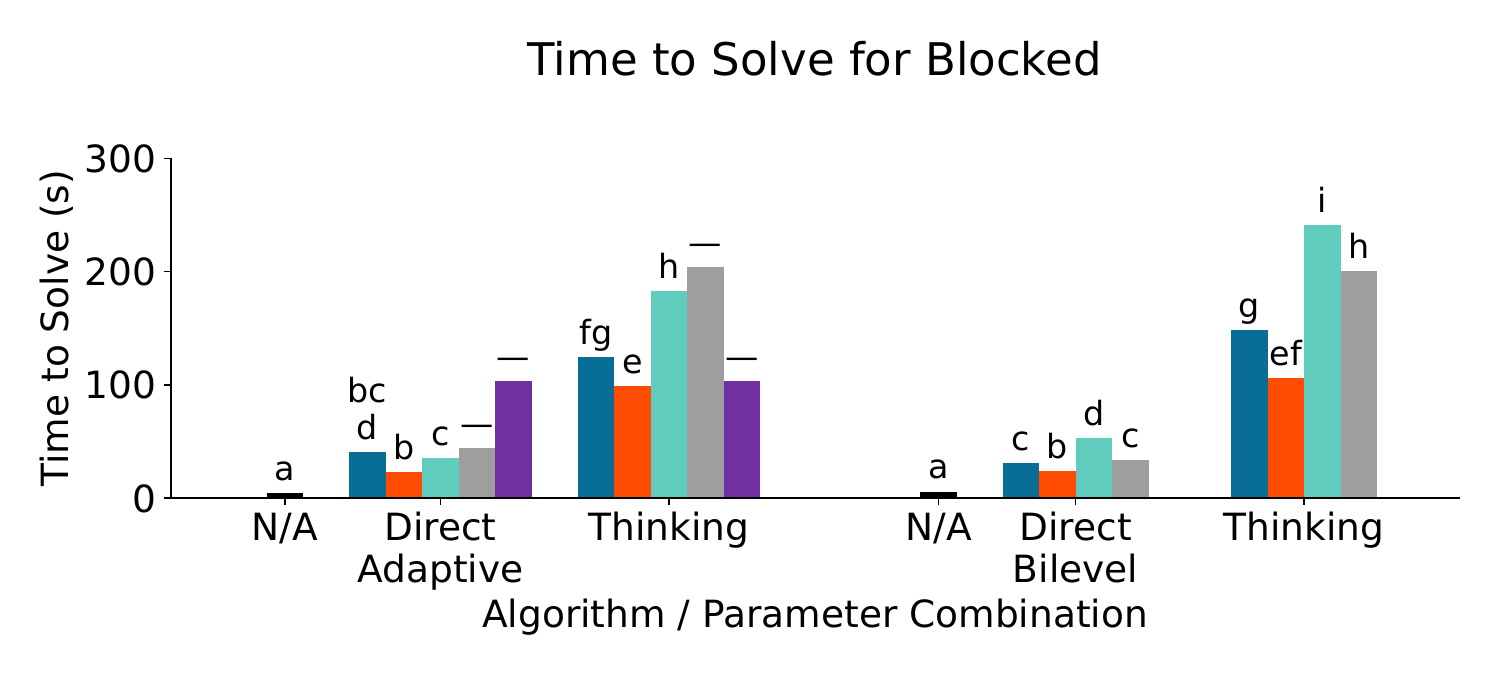}
        \raisebox{0.25cm}{\rotatebox{90}{\tiny Packing $k=3$}}~
        \includegraphics[trim={1.3cm 2.75cm 1cm 1.5cm},clip,height=1.85cm]{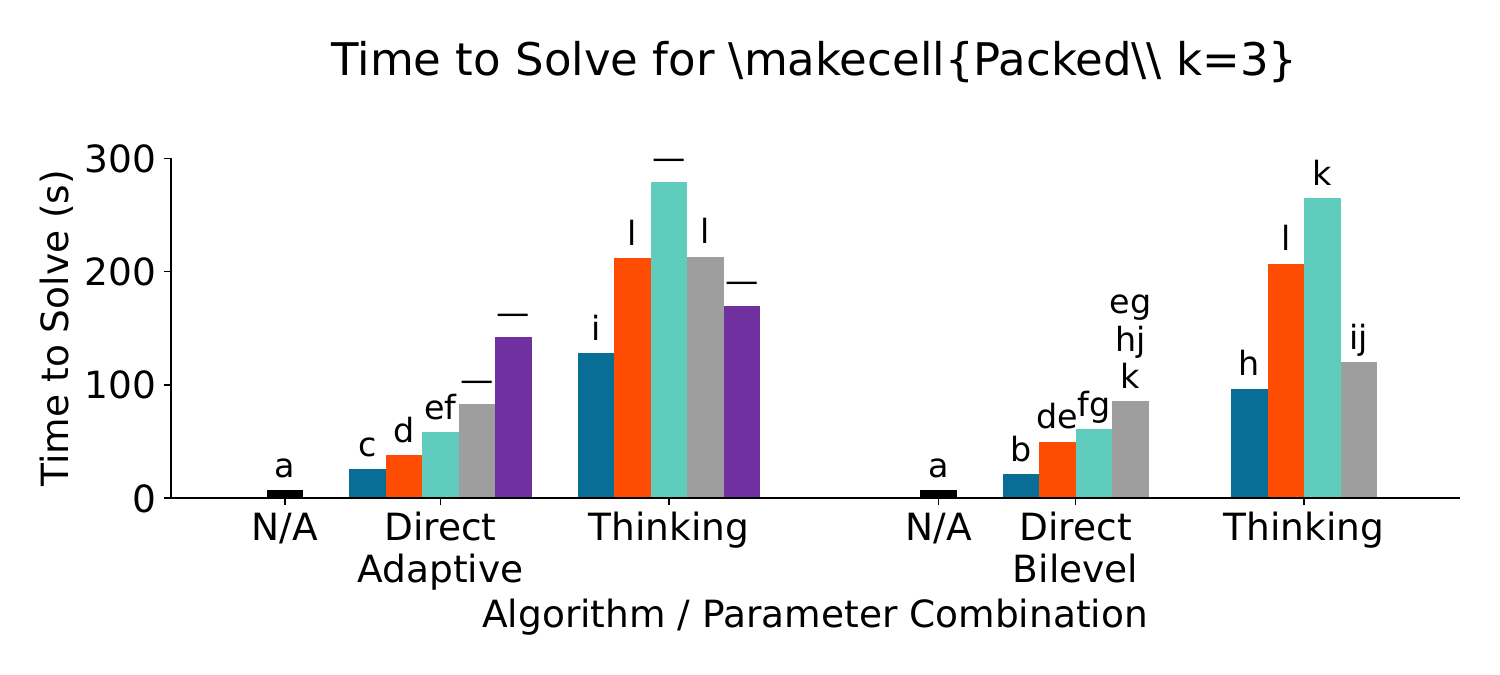}
        \raisebox{0.25cm}{\rotatebox{90}{\tiny Packing $k=4$}}~
        \includegraphics[trim={1.3cm 2.75cm 1cm 1.5cm},clip,height=1.85cm]{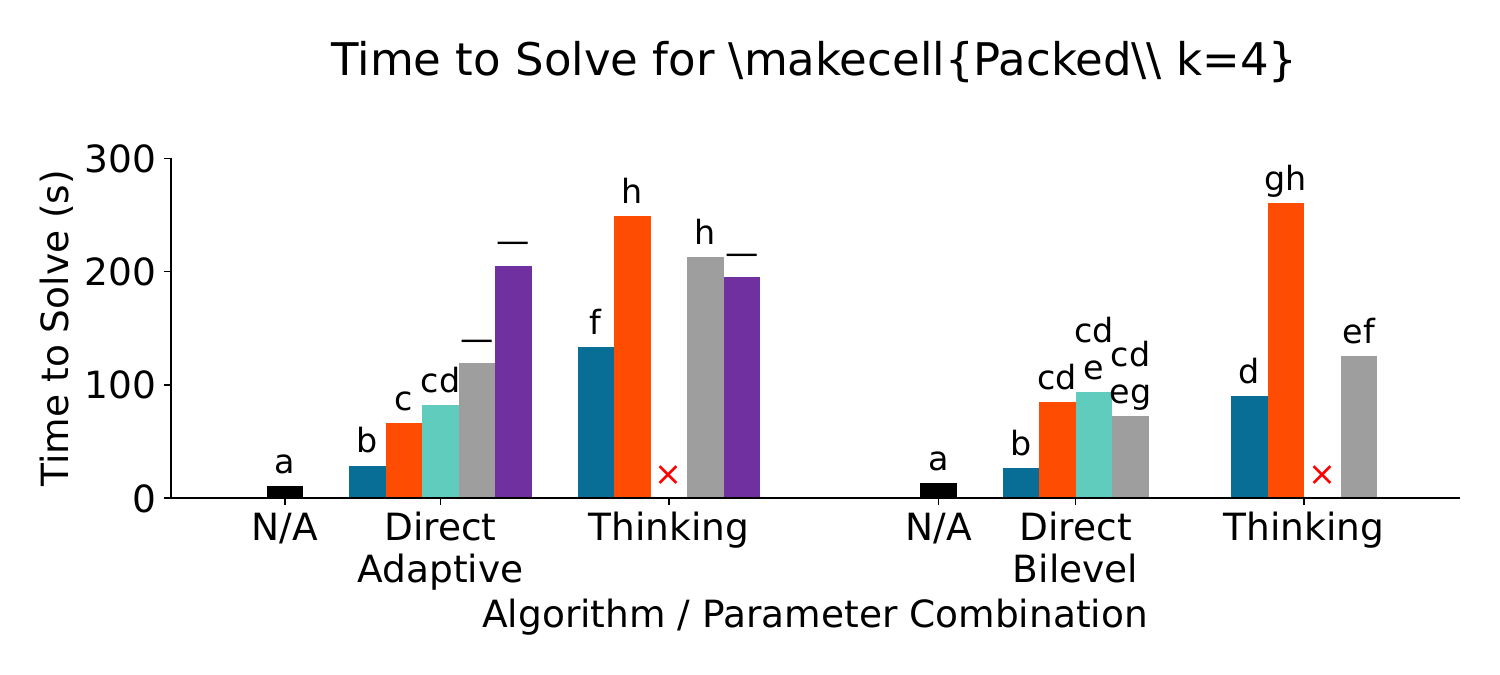}
        \raisebox{0.6cm}{\rotatebox{90}{\tiny Packing $k=5$}}~~~~
        \includegraphics[trim={1.3cm 1.3cm 1cm 1.5cm},clip,height=2.22cm]{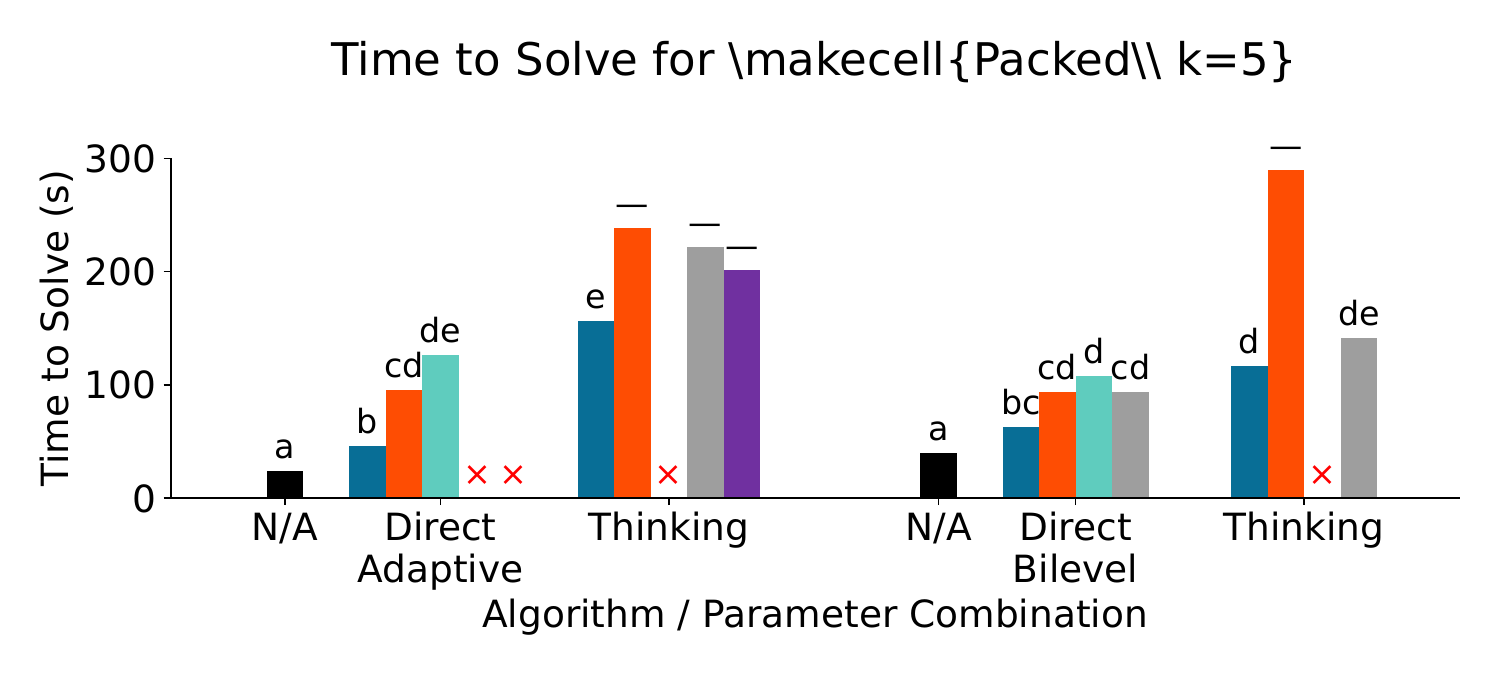}
        \caption{Gemini 2.5 Flash}
    \end{subfigure}\hfill
    \begin{subfigure}[b]{0.312\textwidth}
        \includegraphics[trim={2.8cm 2.75cm 1cm 1.5cm},clip,height=1.85cm]{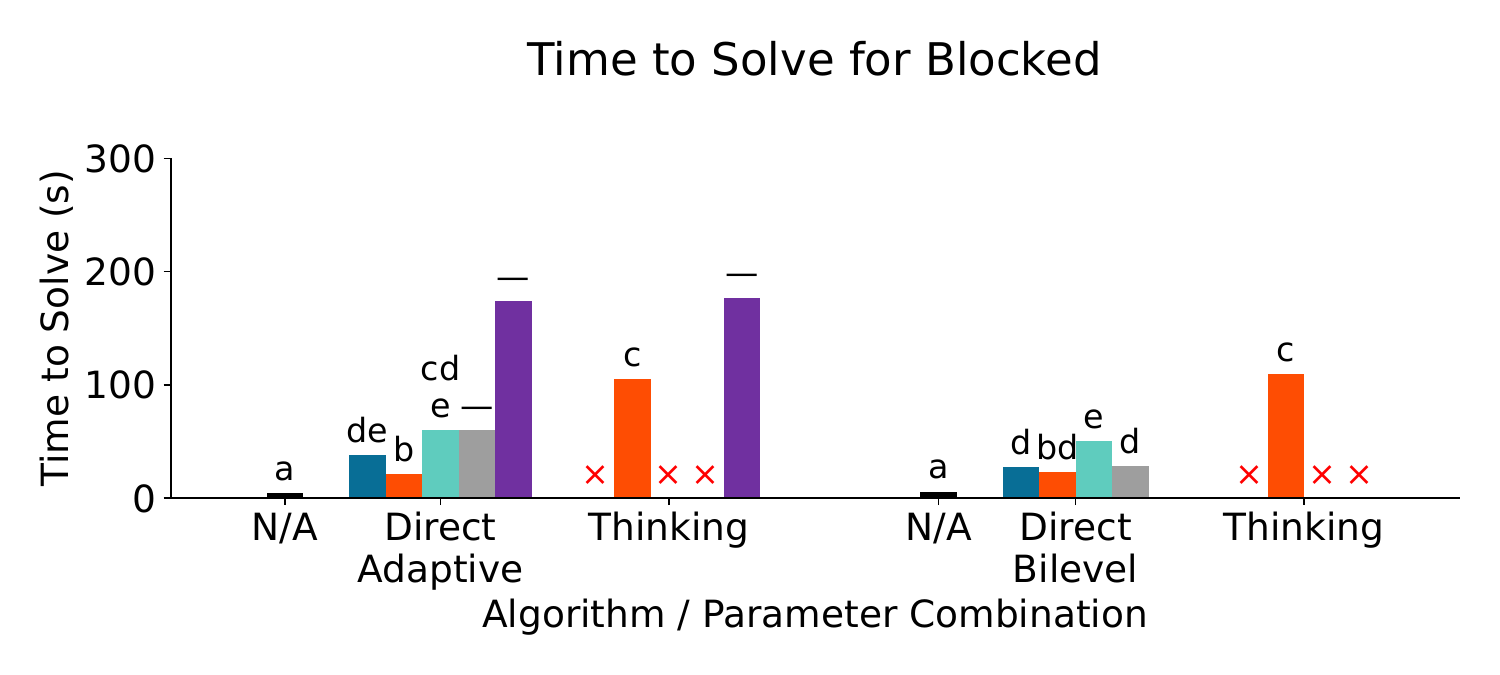}
        \includegraphics[trim={2.8cm 2.75cm 1cm 1.5cm},clip,height=1.85cm]{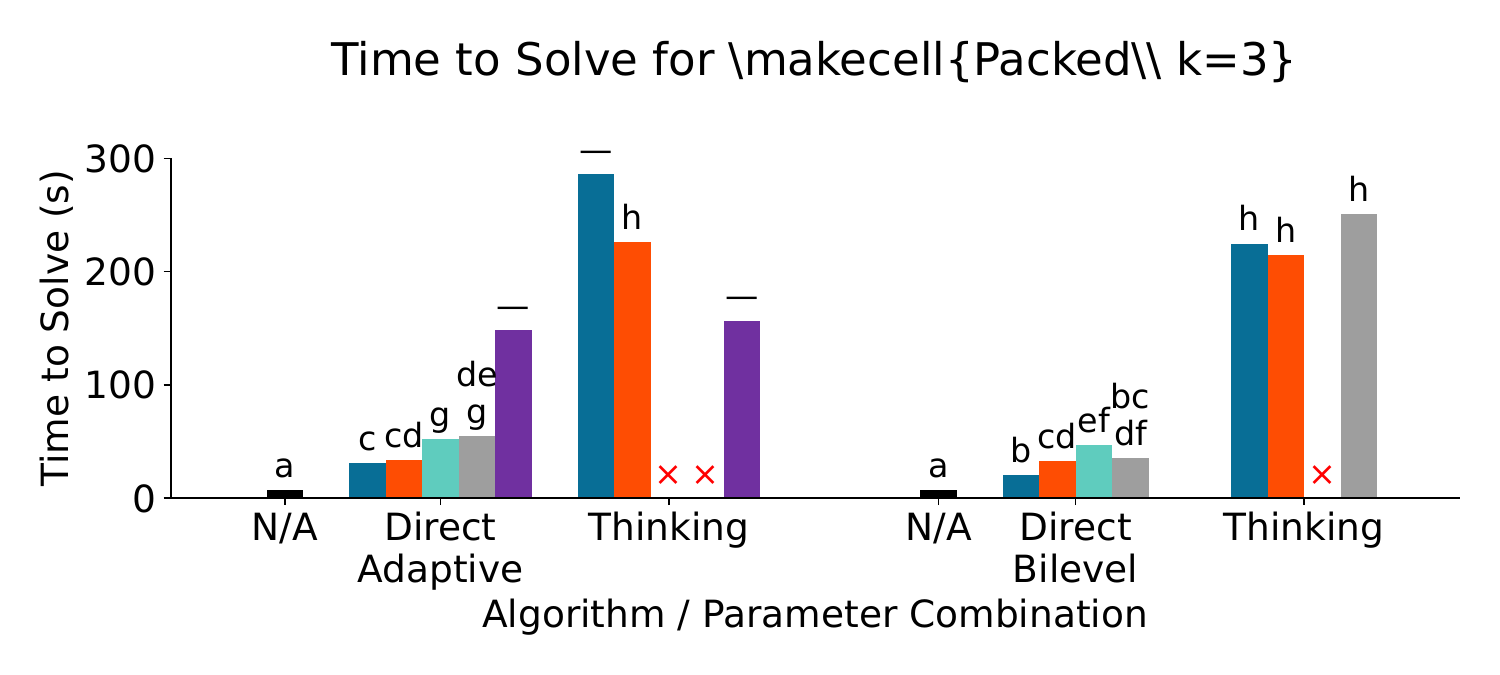}
        \includegraphics[trim={2.8cm 2.75cm 1cm 1.5cm},clip,height=1.85cm]{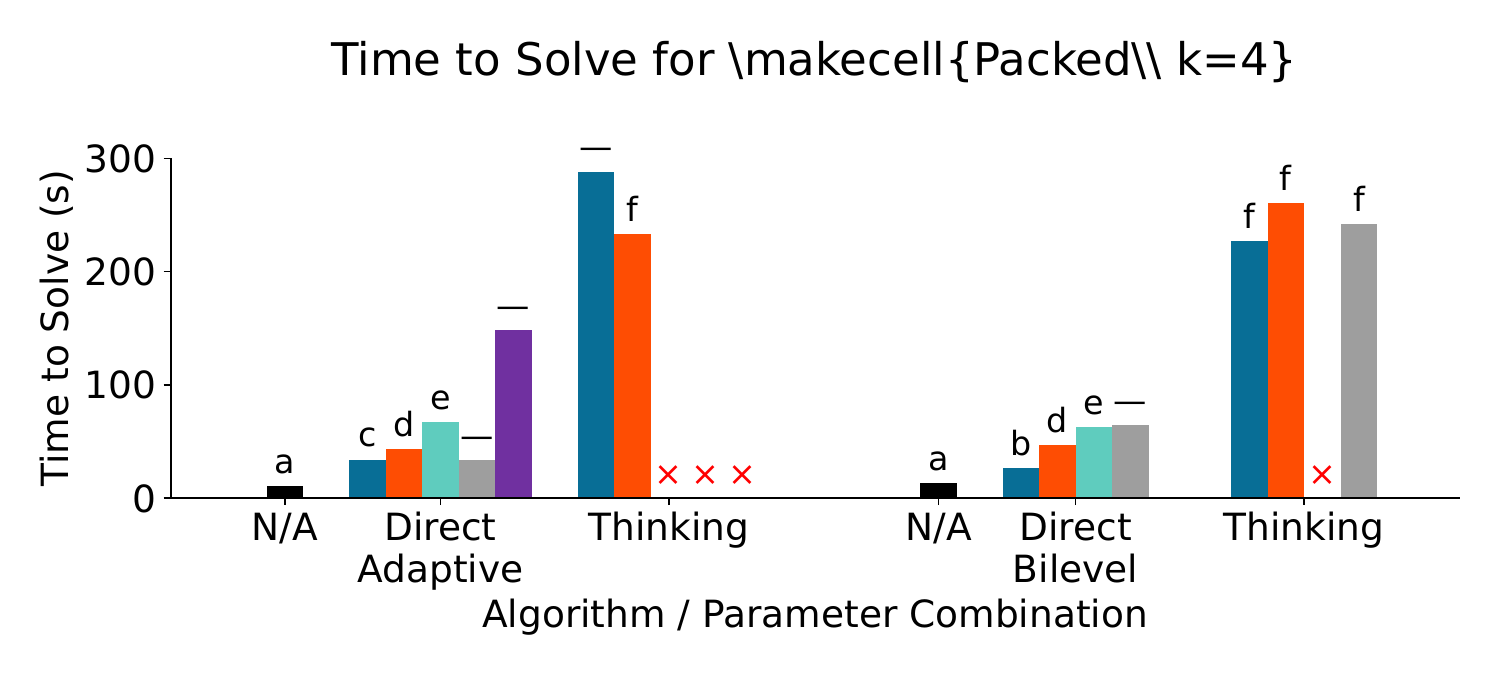}
        \includegraphics[trim={2.8cm 1.3cm 1cm 1.5cm},clip,height=2.22cm]{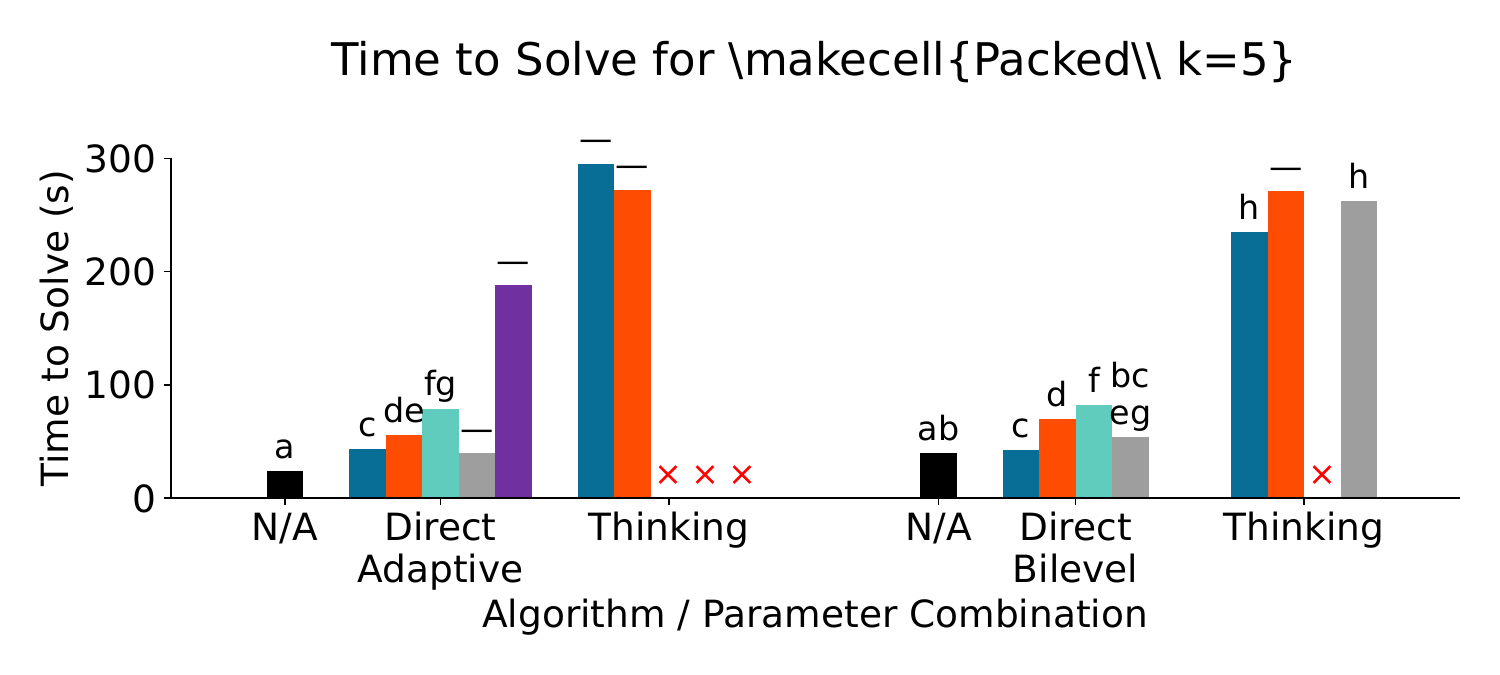}
        \caption{Gemini 3 Flash}
    \end{subfigure}\hfill
    \begin{subfigure}[b]{0.312\textwidth}
        \includegraphics[trim={2.8cm 2.75cm 1cm 1.5cm},clip,height=1.85cm]{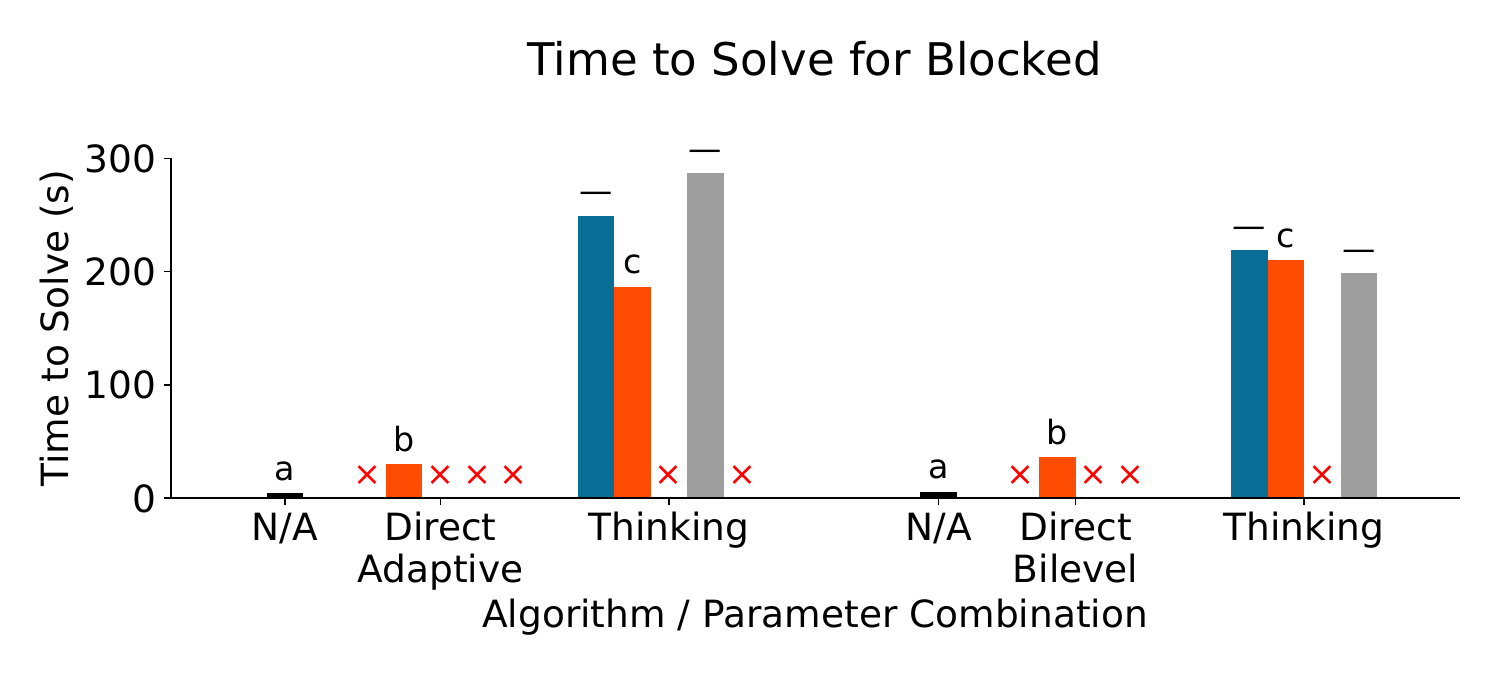}
        \includegraphics[trim={2.8cm 2.75cm 1cm 1.5cm},clip,height=1.85cm]{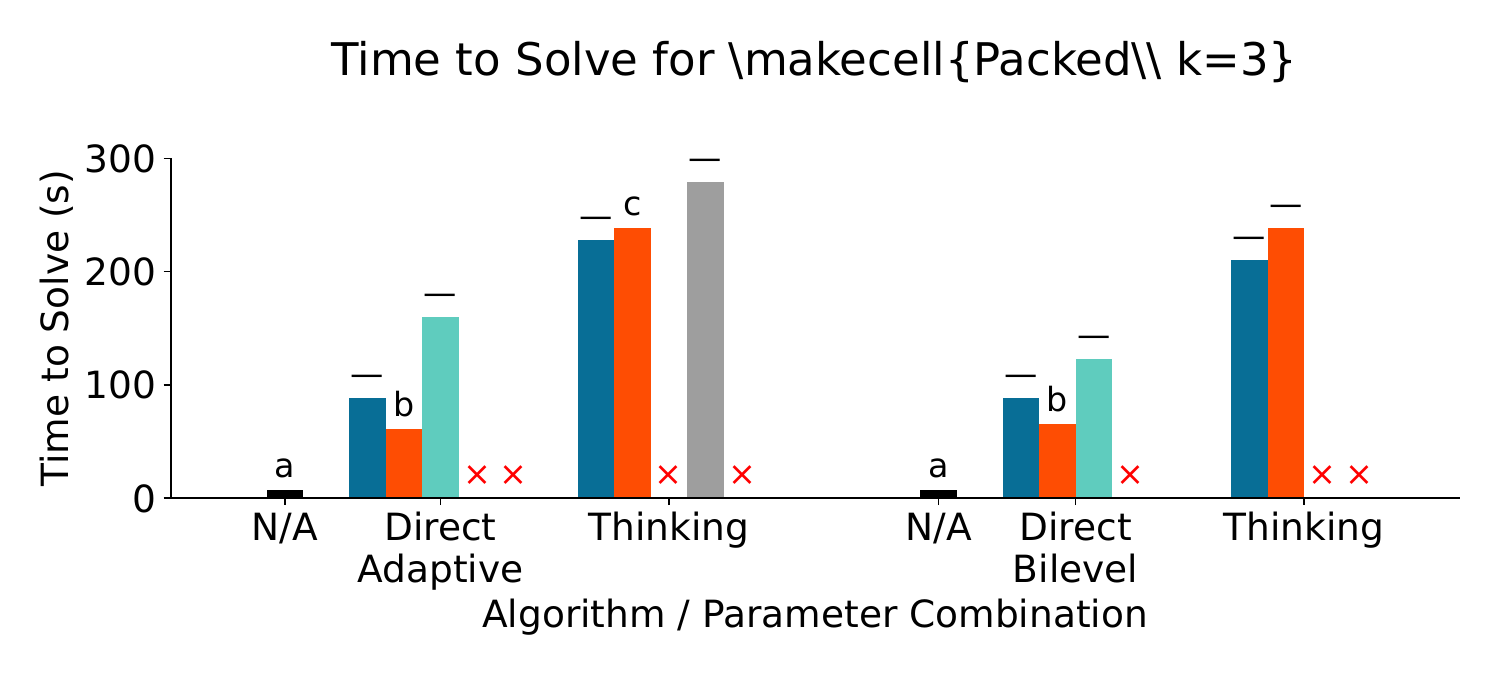}
        \includegraphics[trim={2.8cm 2.75cm 1cm 1.5cm},clip,height=1.85cm]{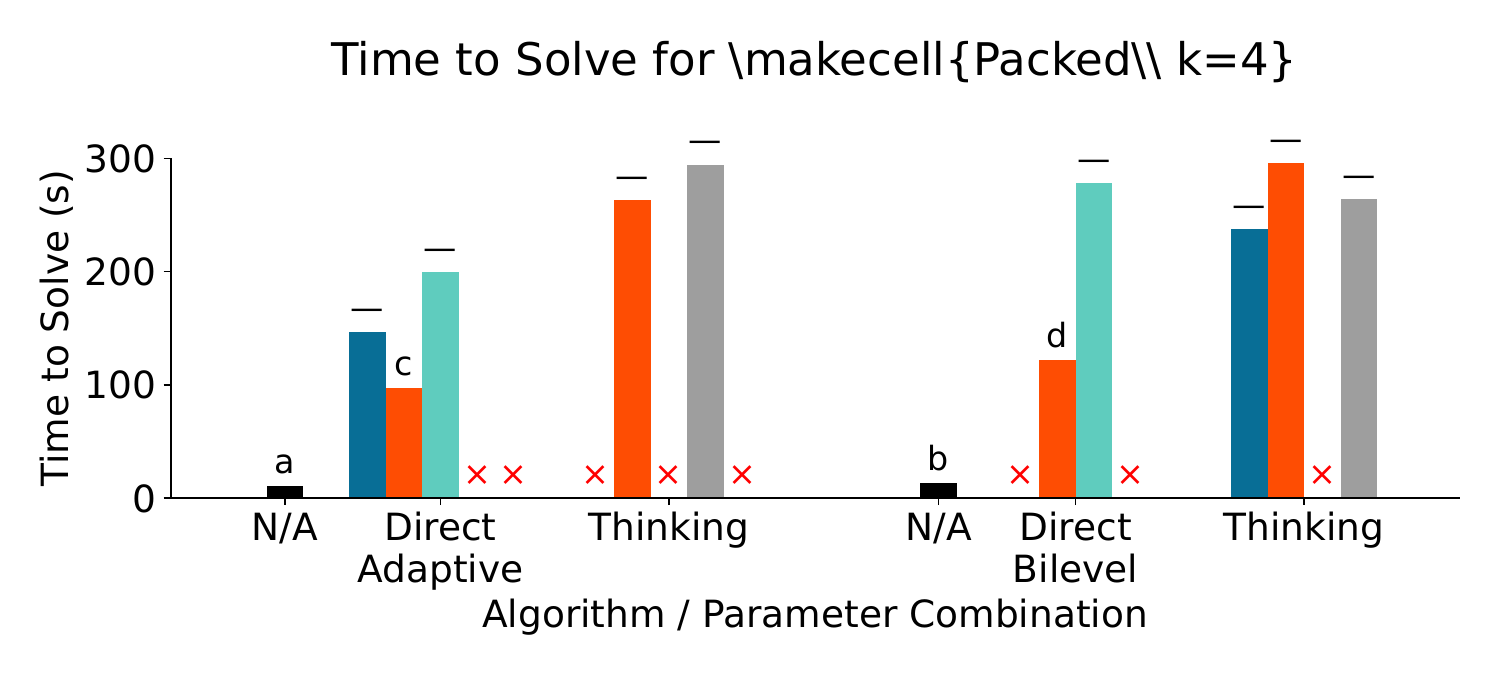}
        \includegraphics[trim={2.8cm 1.3cm 1cm 1.5cm},clip,height=2.22cm]{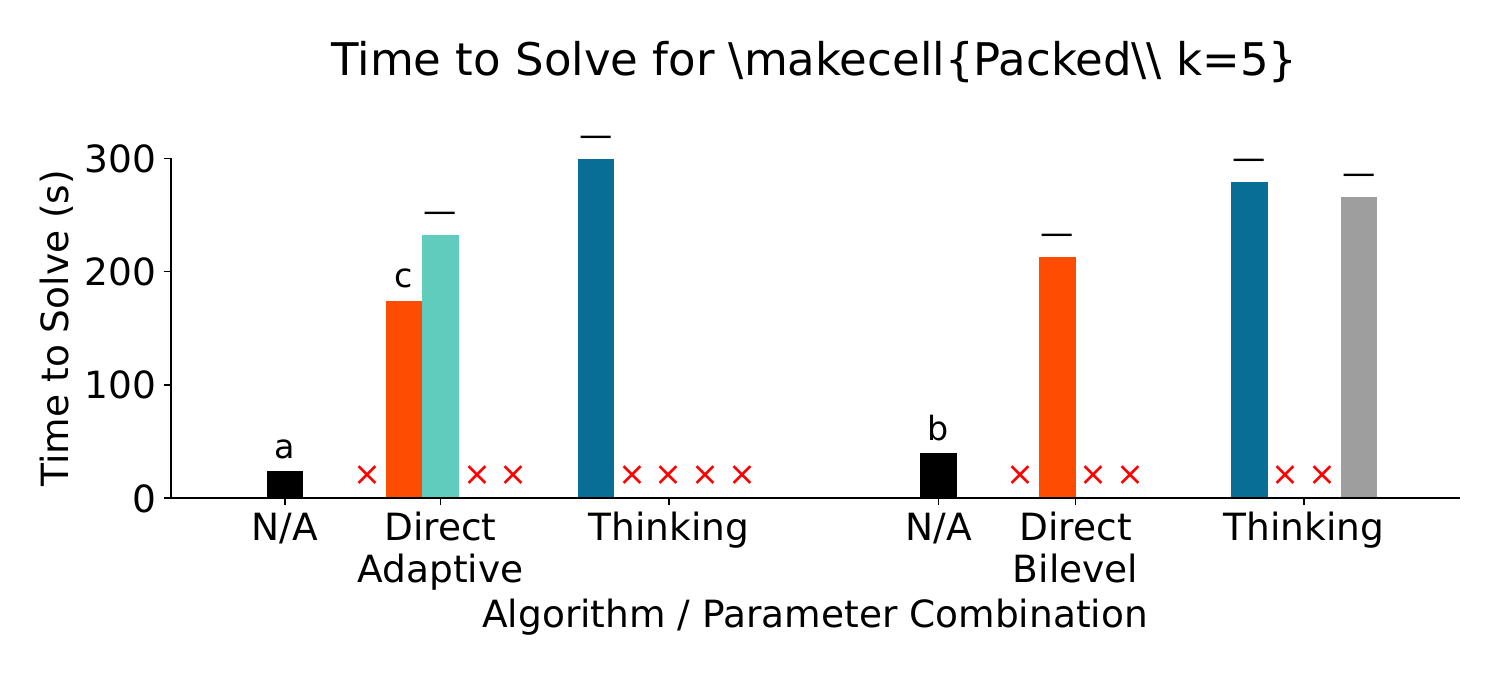}
        \caption{GPT-5 mini}
    \end{subfigure}\hfill
    \vspace{-0.3em}
    \caption{Average success runtime. Non-LLM methods are much faster in all domains. \direct{} variants are faster than their \thinking{} counterparts pairwise. We omit Rovers domains since only the \adaptive{} baseline achieves a success rate $\geq 30\%$.}
    \label{fig:time}
\end{figure*}

\begin{figure*}[t!]
    \centering
    \begin{subfigure}[b]{0.35\textwidth}
        \raisebox{0.05cm}{\rotatebox{90}{\tiny Gemini 2.5 Flash}}~
        \includegraphics[trim={1.4cm 4.25cm 1cm 2.4cm},clip,height=1.3cm]{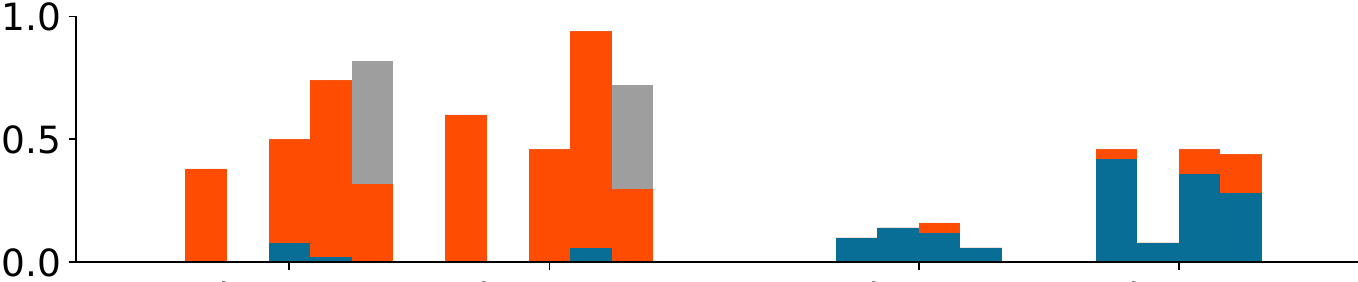}
        \raisebox{0.05cm}{\rotatebox{90}{\tiny Gemini 3 Flash}}~
        \includegraphics[trim={1.4cm 4.25cm 1cm 2.4cm},clip,height=1.3cm]{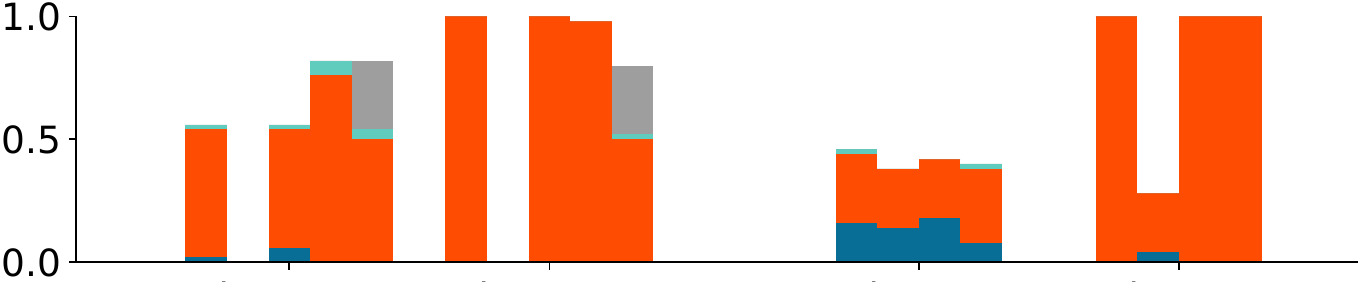}
        \raisebox{0.8cm}{\rotatebox{90}{\tiny GPT-5 mini}}~
        \includegraphics[trim={1.4cm 2.2cm 1cm 2.4cm},clip,height=1.85cm]{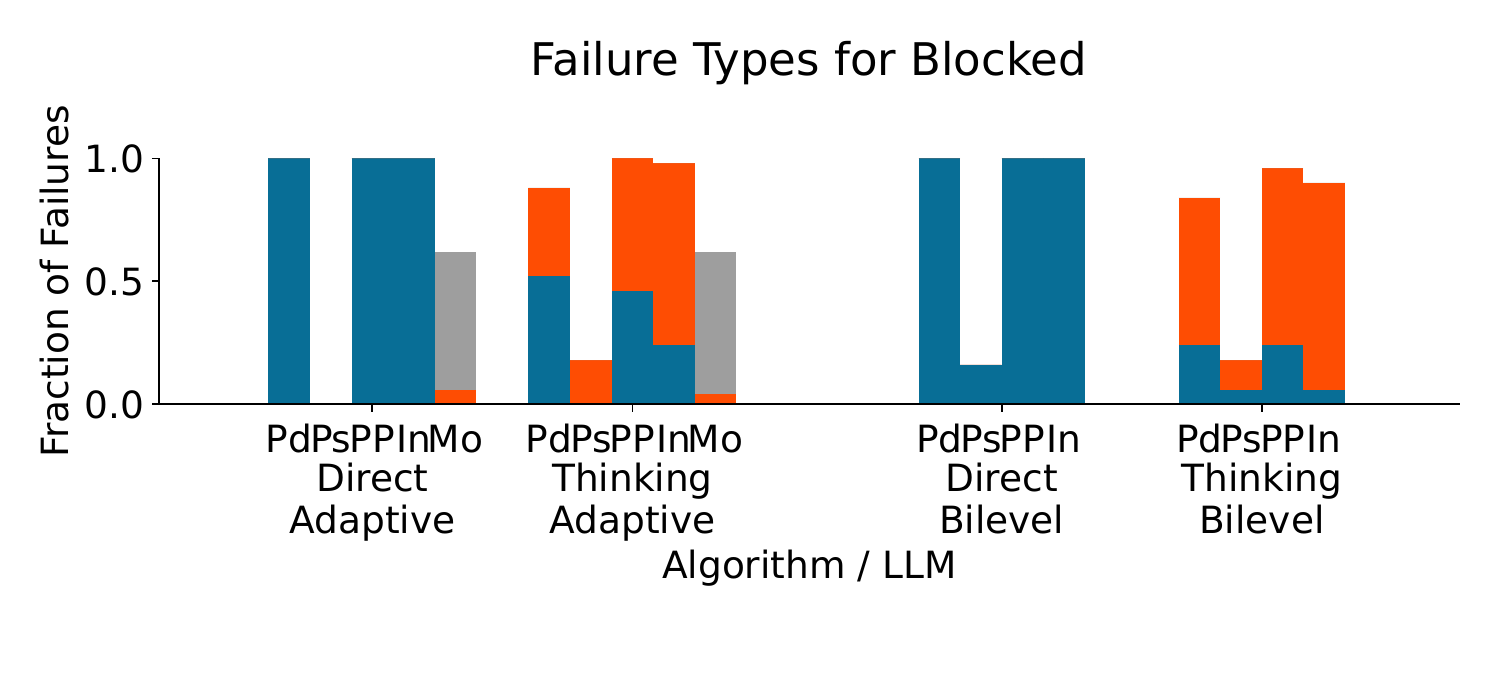}
        \caption{Blocked}
    \end{subfigure}\hfill
    \begin{subfigure}[b]{0.319\textwidth}
        \includegraphics[trim={2.5cm 4.25cm 1cm 2.4cm},clip,height=1.3cm]{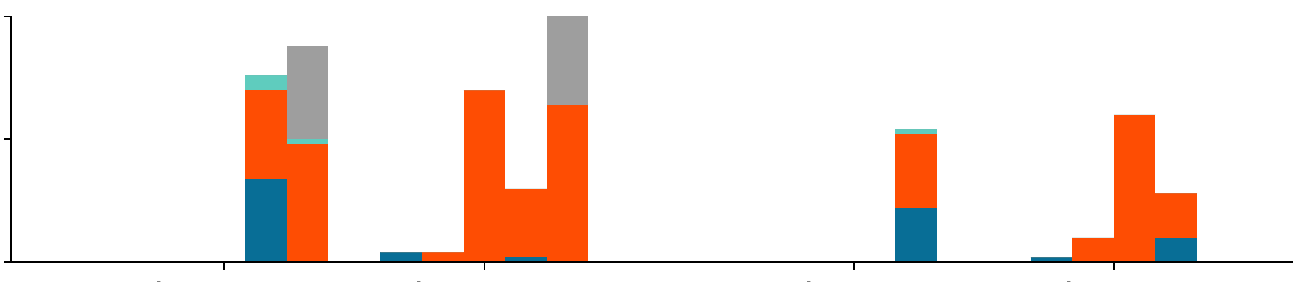}
        \includegraphics[trim={2.5cm 4.25cm 1cm 2.4cm},clip,height=1.3cm]{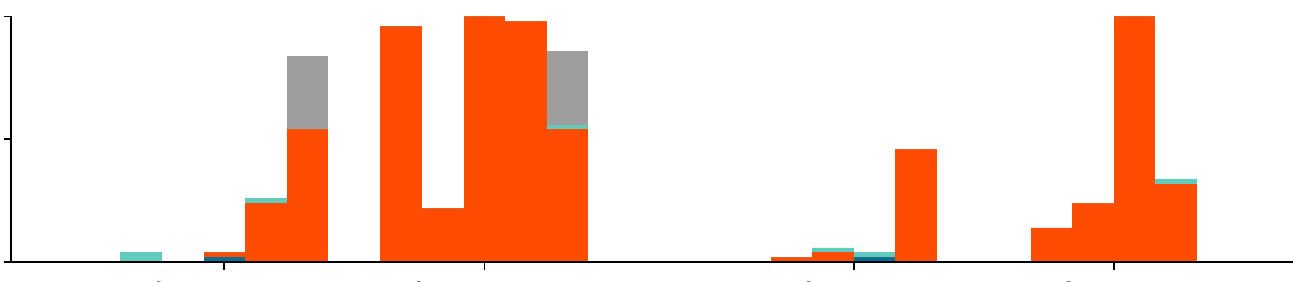}
        \includegraphics[trim={2.5cm 2.2cm 1cm 2.4cm},clip,height=1.85cm]{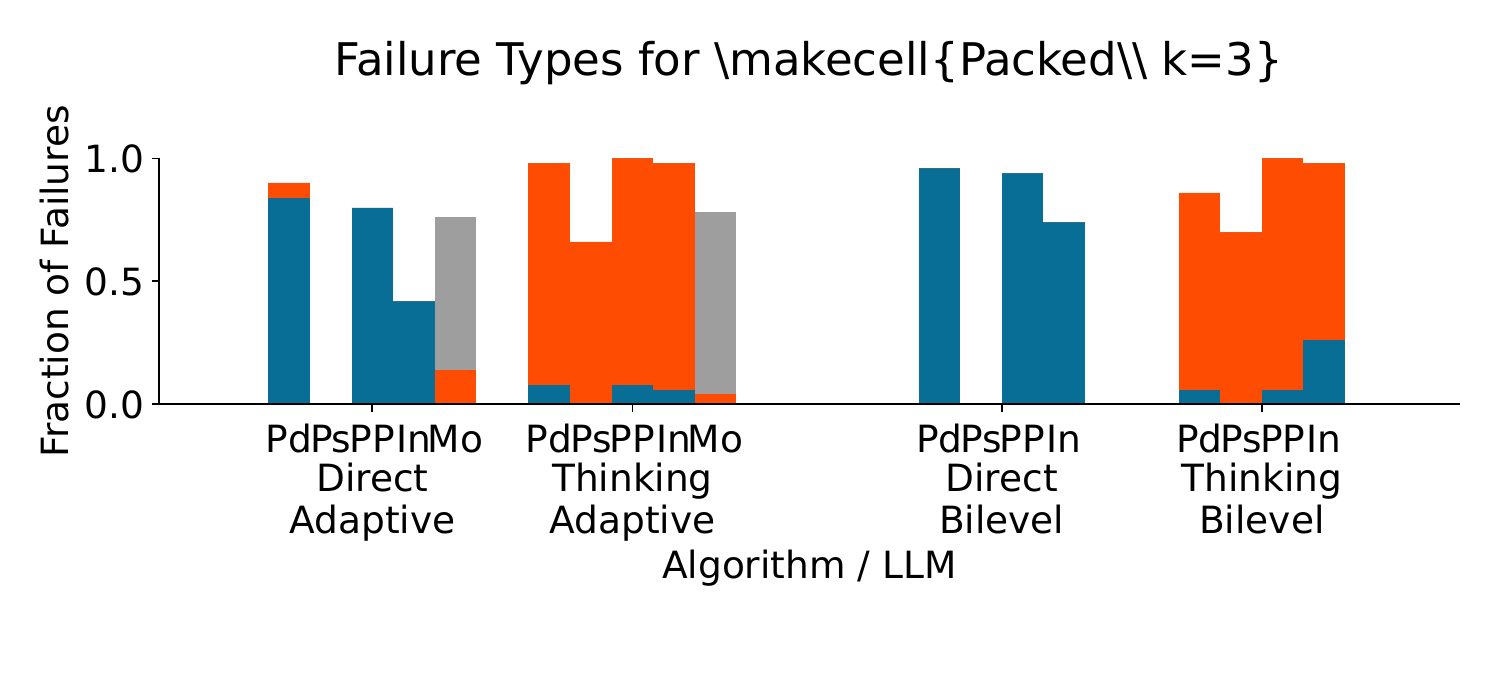}
        \caption{Packing, $k=3$}
    \end{subfigure}\hfill
    \begin{subfigure}[b]{0.319\textwidth}
        \includegraphics[trim={2.5cm 4.25cm 2cm 2.4cm},clip,height=1.3cm]{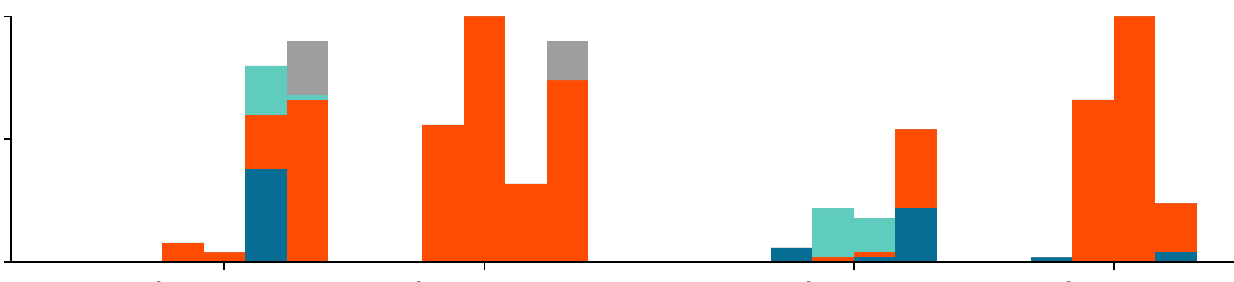}
        \includegraphics[trim={2.5cm 4.25cm 2cm 2.4cm},clip,height=1.3cm]{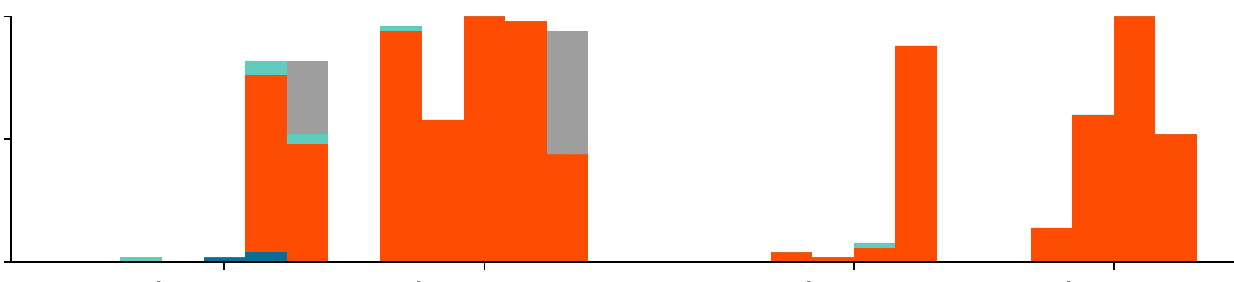}
        \includegraphics[trim={2.5cm 2.2cm 2cm 2.4cm},clip,height=1.85cm]{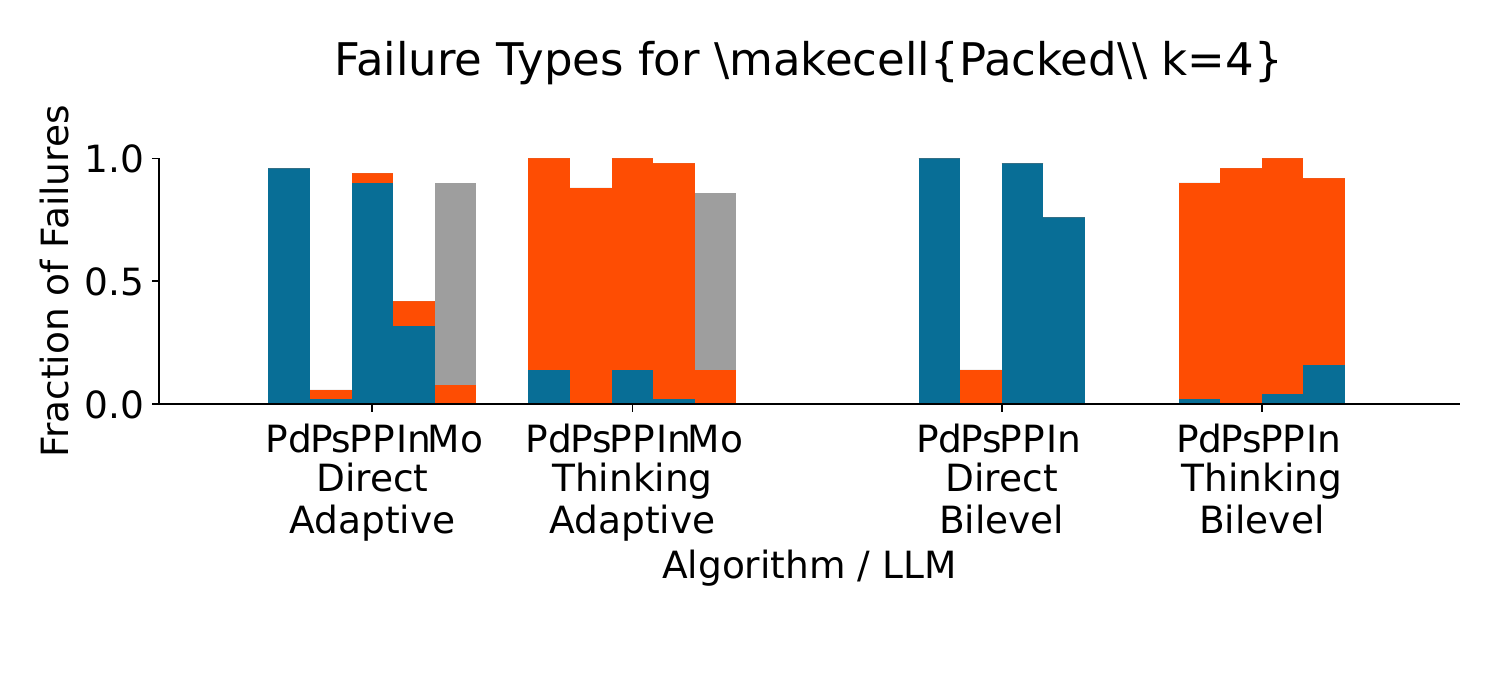}
        \caption{Packing, $k=4$}
    \end{subfigure}
    \par\medskip
    \begin{subfigure}[b]{0.36\textwidth}
        \raisebox{0.05cm}{\rotatebox{90}{\tiny Gemini 2.5 Flash}}~
        \includegraphics[trim={1.4cm 4.25cm 1cm 2.4cm},clip,height=1.3cm]{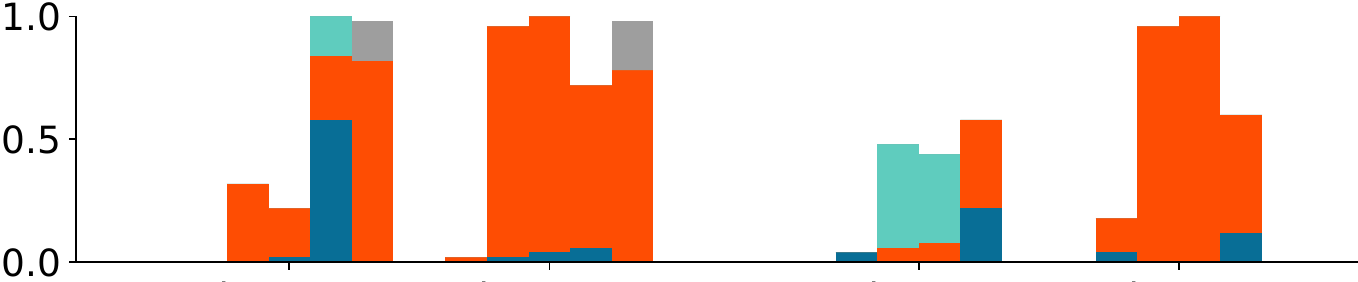}
        \raisebox{0.05cm}{\rotatebox{90}{\tiny Gemini 3 Flash}}~
        \includegraphics[trim={1.4cm 4.25cm 1cm 2.4cm},clip,height=1.3cm]{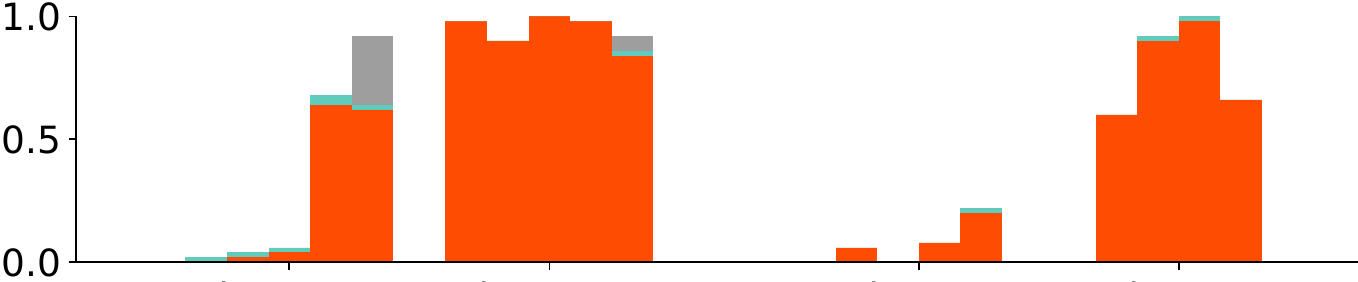}
        \raisebox{0.8cm}{\rotatebox{90}{\tiny GPT-5 mini}}~
        \includegraphics[trim={1.4cm 2.2cm 1cm 2.4cm},clip,height=1.85cm]{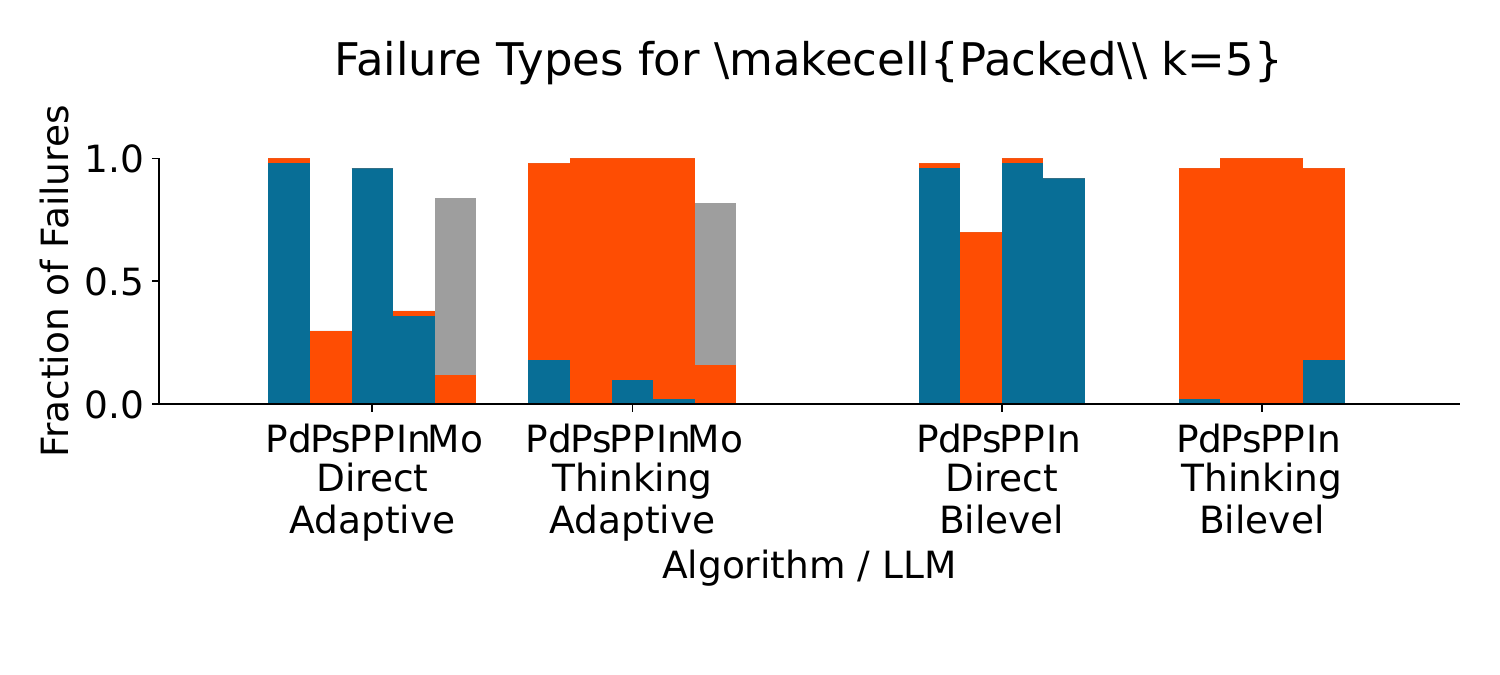}
        \caption{Packing, $k=5$}
    \end{subfigure}\hfill
    \begin{subfigure}[b]{0.177\textwidth}
        \includegraphics[trim={2.5cm 4.25cm 11cm 2.4cm},clip,height=1.3cm]{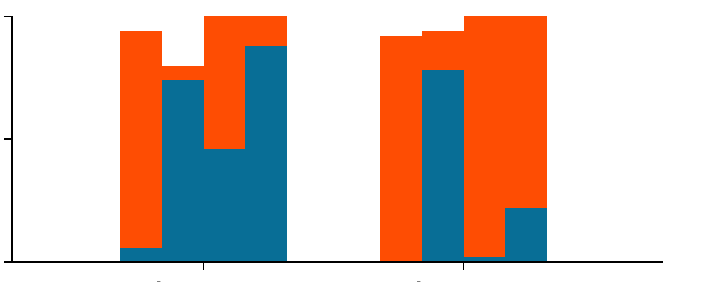}
        \includegraphics[trim={2.5cm 4.25cm 11cm 2.4cm},clip,height=1.3cm]{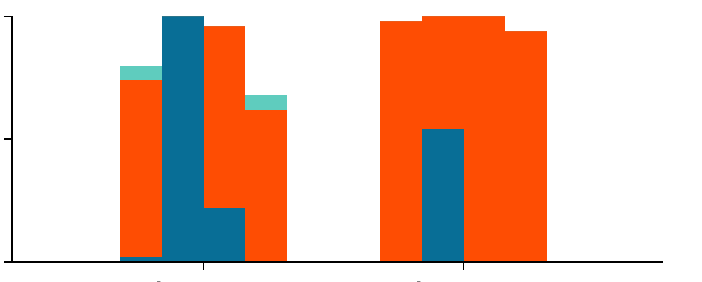}
        \includegraphics[trim={2.5cm 2.2cm 11cm 2.4cm},clip,height=1.85cm]{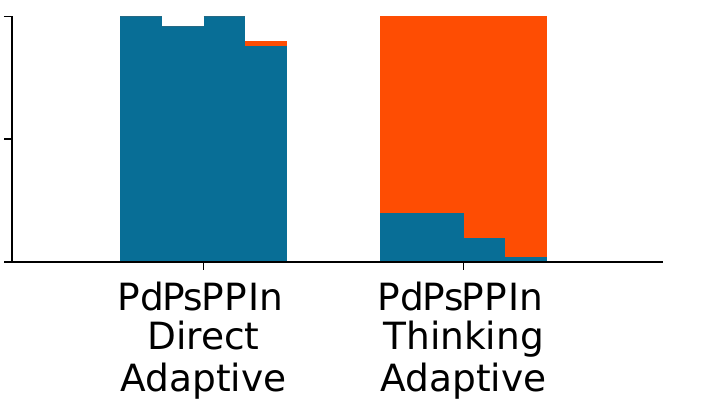}
        \caption{Rovers, $k=2$}
    \end{subfigure}\hfill
    \begin{subfigure}[b]{0.177\textwidth}
        \includegraphics[trim={2.5cm 4.25cm 11cm 2.4cm},clip,height=1.3cm]{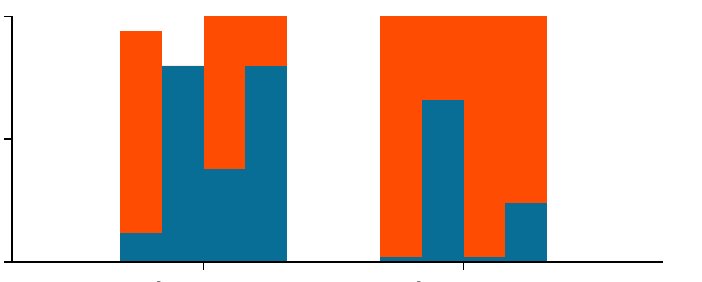}
        \includegraphics[trim={2.5cm 4.25cm 11cm 2.4cm},clip,height=1.3cm]{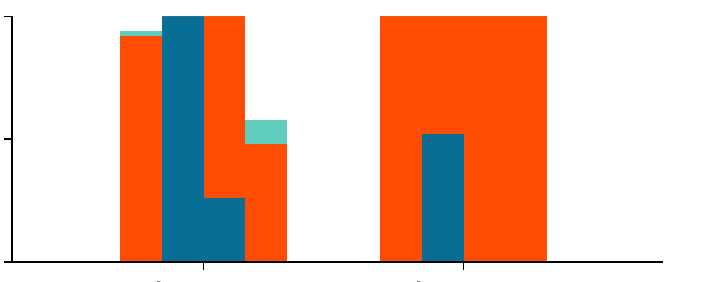}
        \includegraphics[trim={2.5cm 2.2cm 11cm 2.4cm},clip,height=1.85cm]{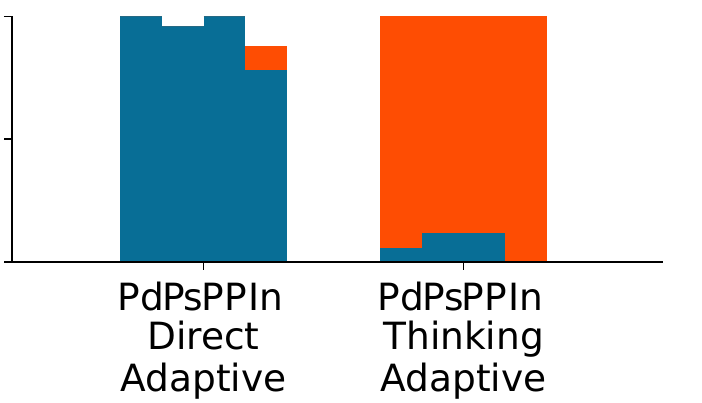}
        \caption{Rovers, $k=3$}
    \end{subfigure}\hfill
    \begin{subfigure}[b]{0.177\textwidth}
        \includegraphics[trim={2.5cm 4.25cm 11cm 2.4cm},clip,height=1.3cm]{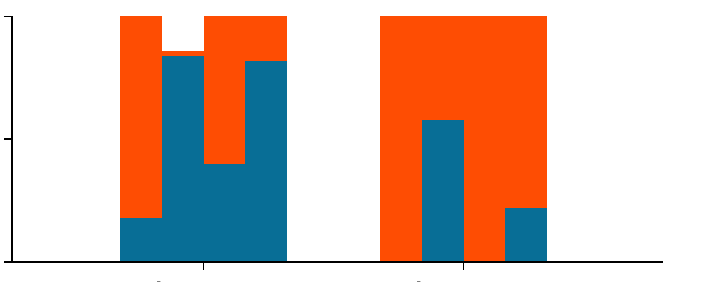}
        \includegraphics[trim={2.5cm 4.25cm 11cm 2.4cm},clip,height=1.3cm]{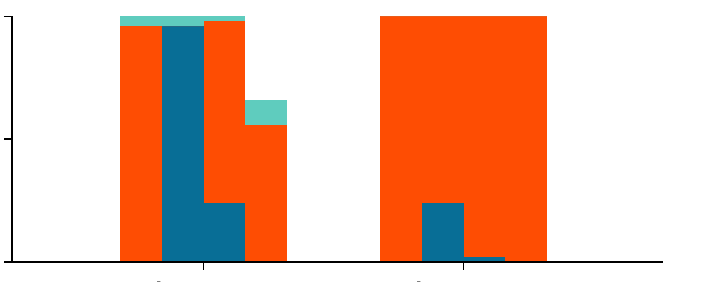}
        \includegraphics[trim={2.5cm 2.2cm 11cm 2.4cm},clip,height=1.85cm]{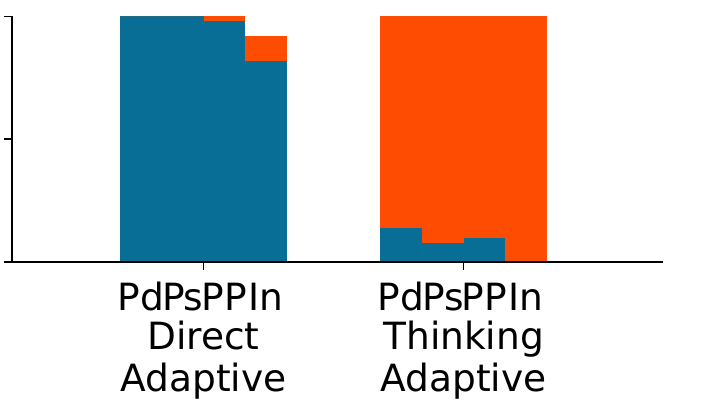}
        \caption{Rovers, $k=4$}
    \end{subfigure}\hfill
    \begin{subfigure}[b]{0.09\textwidth}
        \includegraphics[trim={18.7cm 2.6cm 0.8cm 1.cm},clip,height=1.85cm]{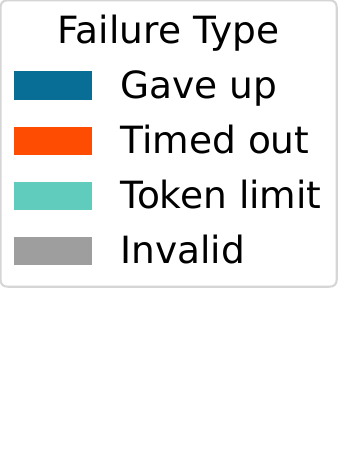}
        \caption*{\ }
    \end{subfigure}
    \caption{Fraction of 50 problems per domain that failed for each reason per algorithm. Most failures occur due to time-outs, but LLMs also often give up and assert that a (solvable) problem is not solvable, or produce an invalid abstraction. Some algorithms often face token limits. (LLM usage legend: Pd=\pddl{}, Ps=\poses{}, PP=\pddl{}+\poses{}, In=\integrated{}, Mo=\abstraction{}.)}
    \label{fig:failureReasons}
\end{figure*}

\begin{figure*}[t!]
    \centering
    \mbox{%
        \begin{subfigure}[b]{0.105\textwidth}

            \begin{overpic}[trim={1.4cm 4.4cm 15.1cm 2.3cm},clip,height=4.01cm]{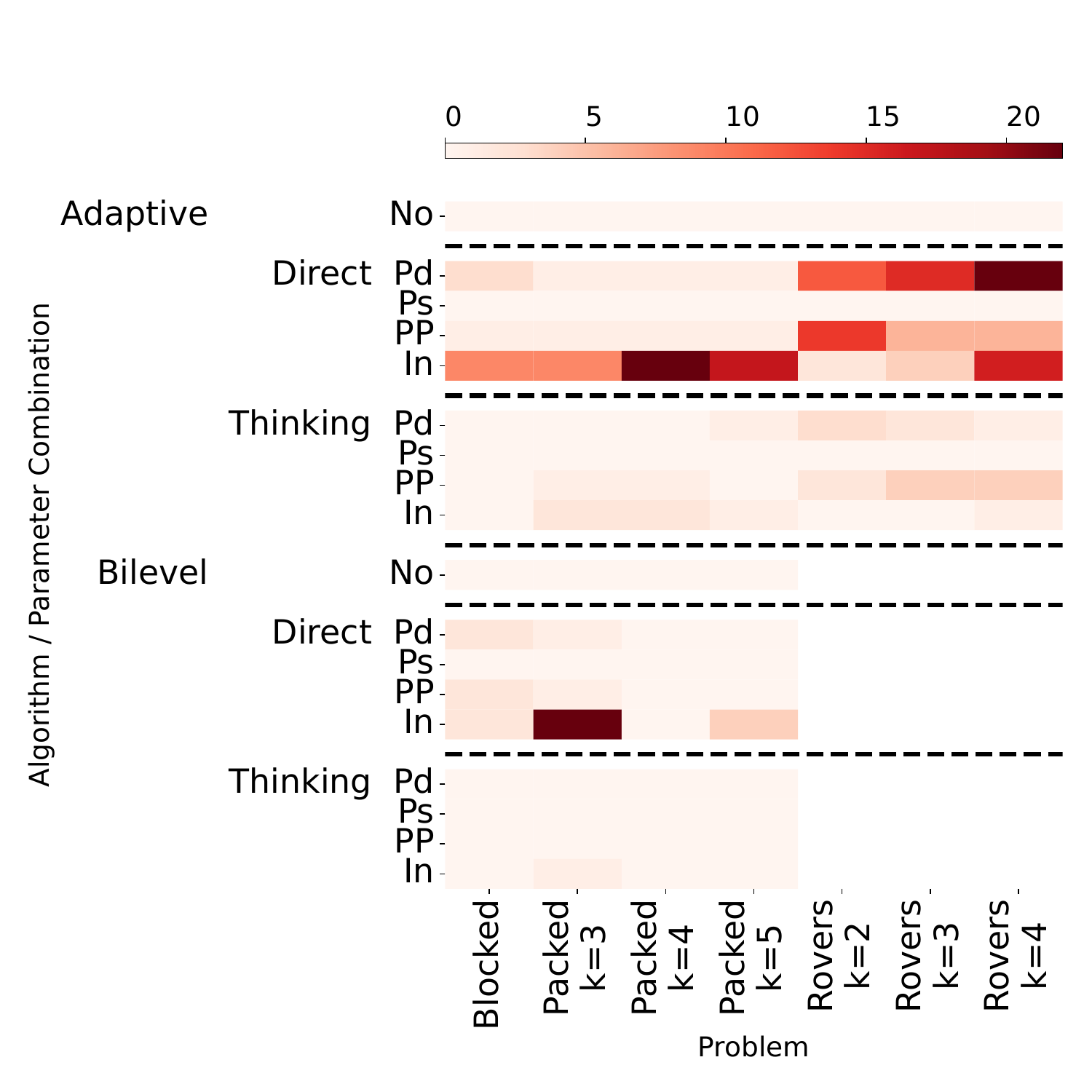}
                \put(0, 35){\rotatebox{90}{\tiny Gemini 2.5 Flash}}
            \end{overpic}
            \hrule
            \medskip
            \begin{overpic}[trim={1.4cm 4.4cm 15.1cm 4.7cm},clip,height=3.48cm]{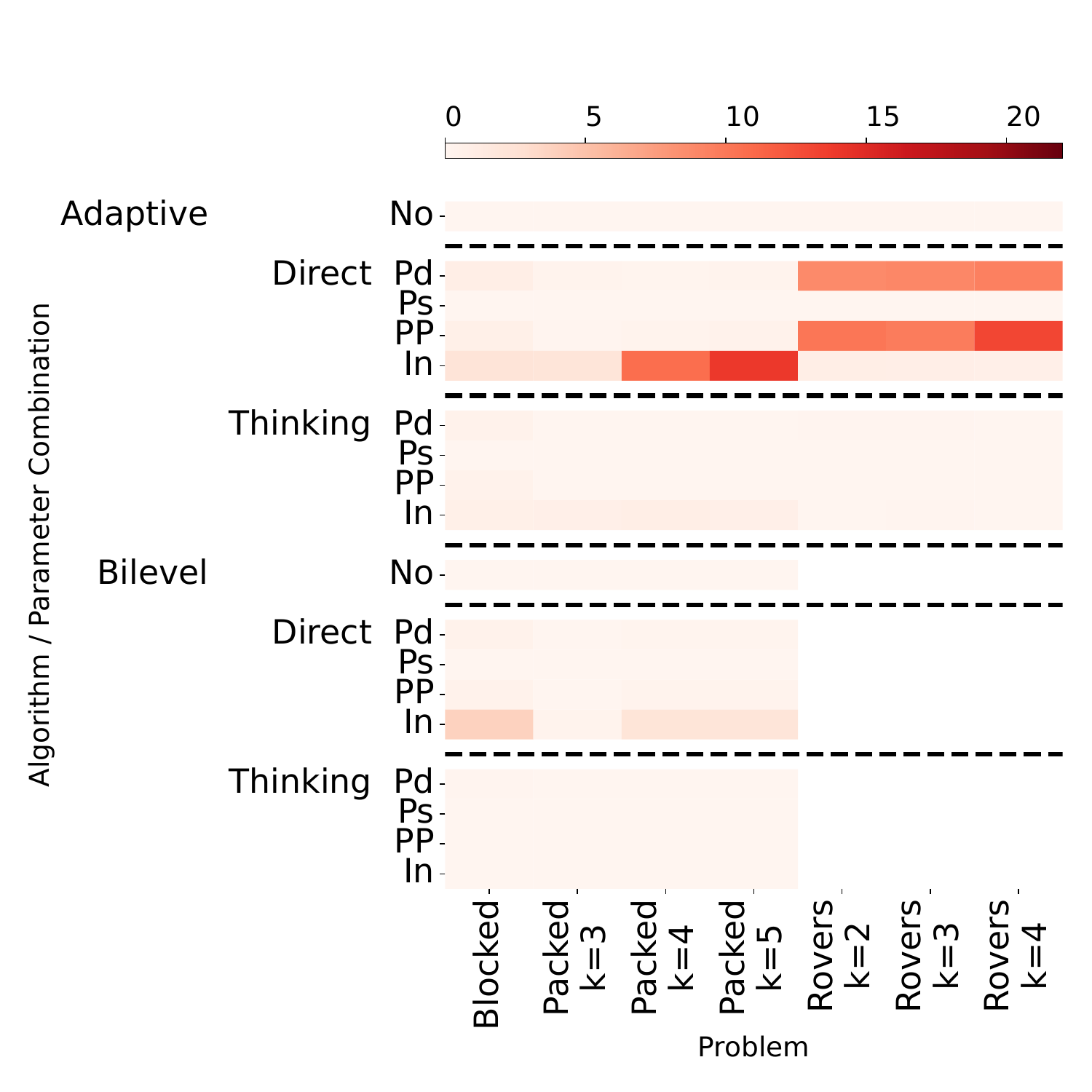}
                \put(0, 40){\rotatebox{90}{\tiny Gemini 3 Flash}}
            \end{overpic}
            \hrule
            \medskip
            \begin{overpic}[trim={1.4cm 1.5cm 15.1cm 4.7cm},clip,height=4.10cm]{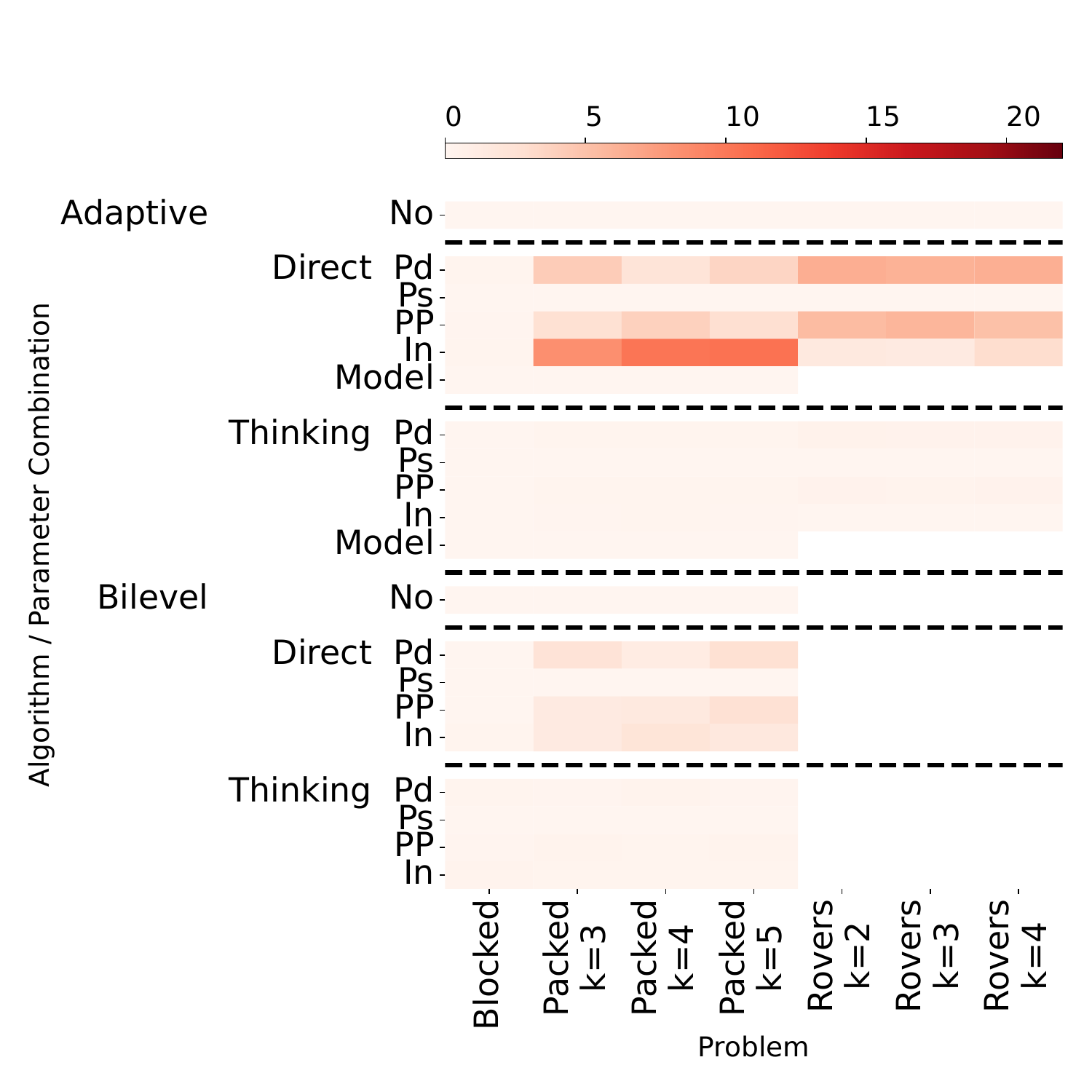}
                \put(0, 50){\rotatebox{90}{\tiny GPT-5 mini}}
            \end{overpic}
            \caption*{\ }
            \label{fig:dummy1}
        \end{subfigure}%
        \begin{subfigure}[b]{0.179\textwidth}
            \includegraphics[trim={10.2cm 4.4cm 0.4cm 2.3cm},clip,height=4.01cm]{Figures/heatmap_pddl_failures_gemini25.pdf}
            \hrule
            \medskip
            \includegraphics[trim={10.2cm 4.4cm 0.4cm 4.7cm},clip,height=3.48cm]{Figures/heatmap_pddl_failures_gemini3.pdf}
            \hrule
            \medskip
            \includegraphics[trim={10.2cm 1.5cm 0.4cm 4.7cm},clip,height=4.10cm]{Figures/heatmap_pddl_failures_gpt5.pdf}
            \caption{PDDL failures}
            \label{fig:pddlFailuresHeatmap}
        \end{subfigure}%
        \begin{subfigure}[b]{0.179\textwidth}
            \includegraphics[trim={10.2cm 4.4cm 0.4cm 2.3cm},clip,height=4.01cm]{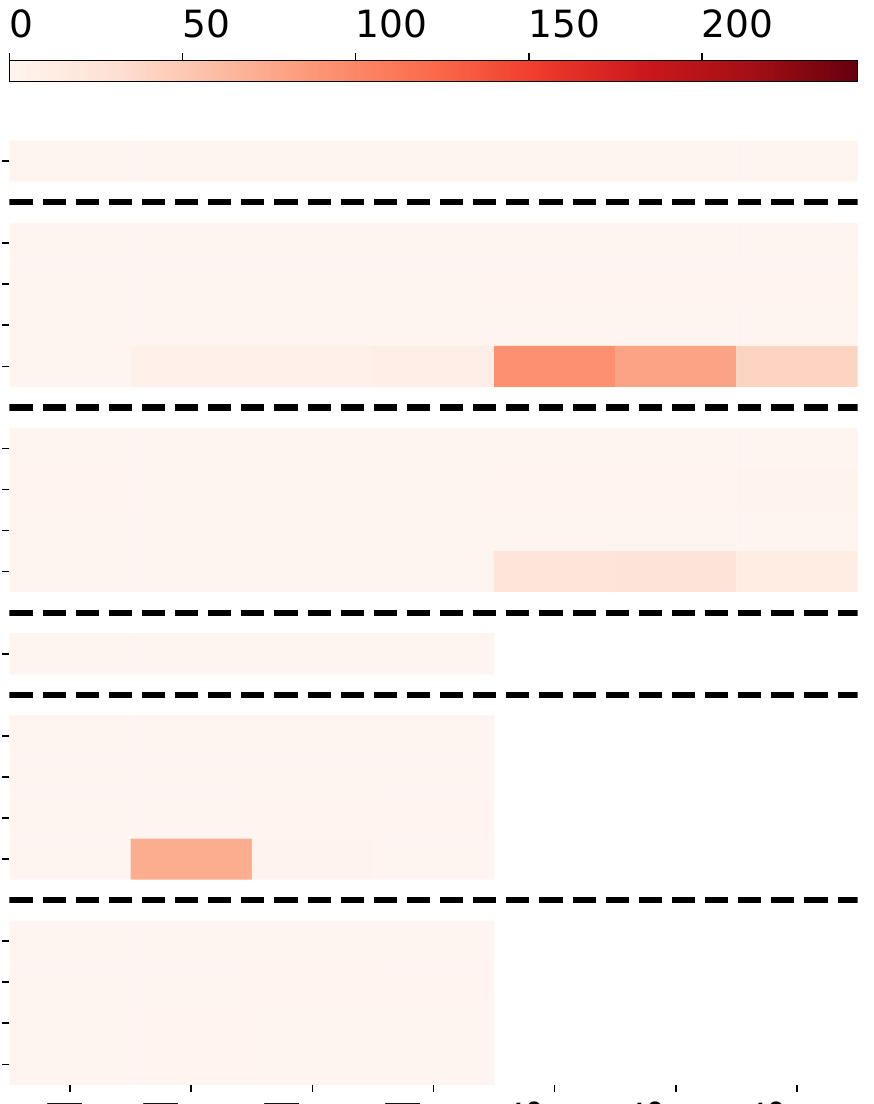}
            \hrule
            \medskip
            \includegraphics[trim={10.2cm 4.4cm 0.4cm 4.7cm},clip,height=3.48cm]{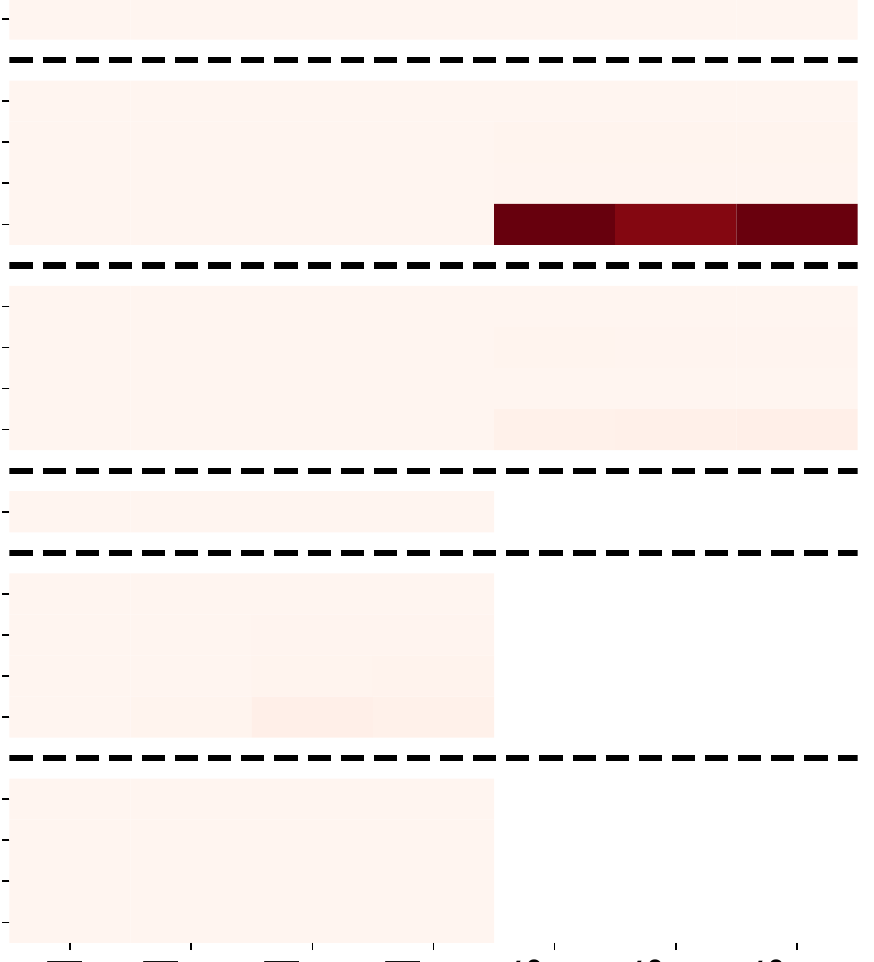}
            \hrule
            \medskip
            \includegraphics[trim={10.2cm 1.5cm 0.4cm 4.7cm},clip,height=4.10cm]{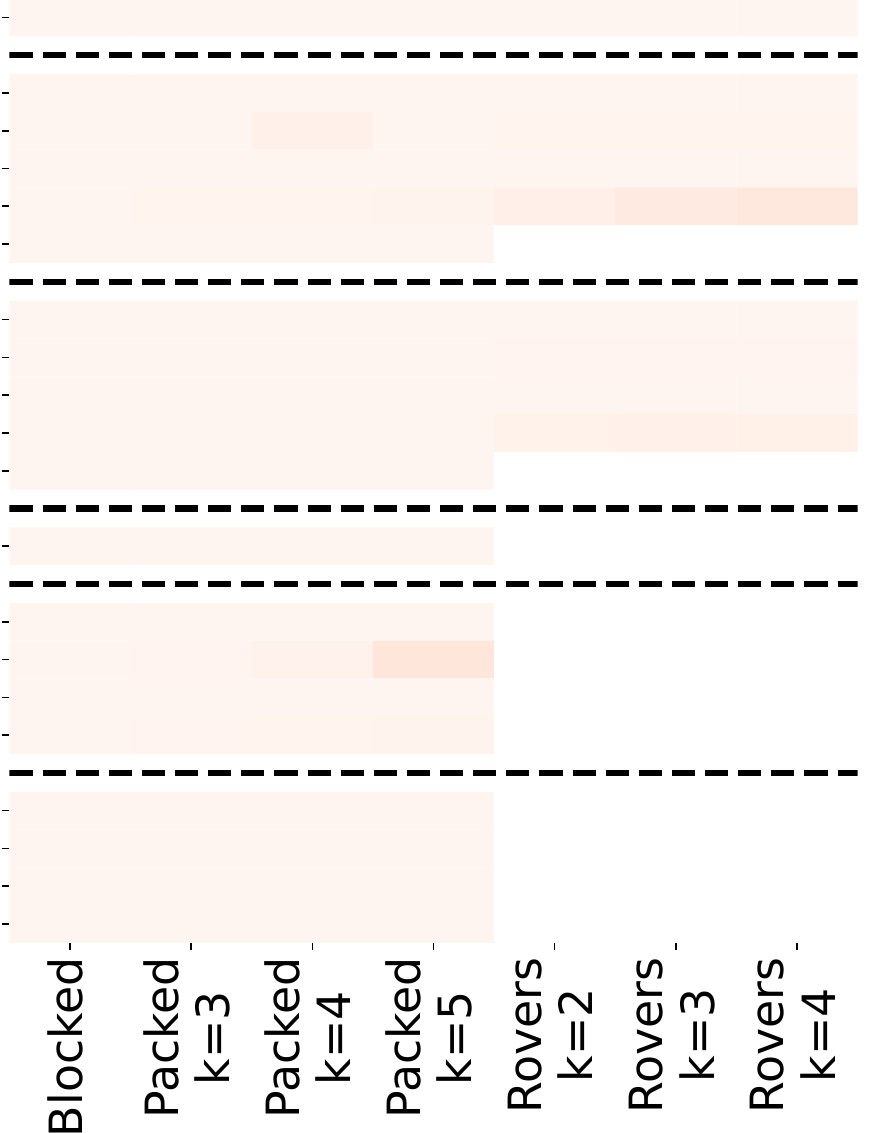}
            \caption{Sampling failures}
            \label{fig:samplingFailuresHeatmap}
        \end{subfigure}%
        \begin{subfigure}[b]{0.179\textwidth}
            \includegraphics[trim={10.2cm 4.4cm 0.4cm 2.3cm},clip,height=4.01cm]{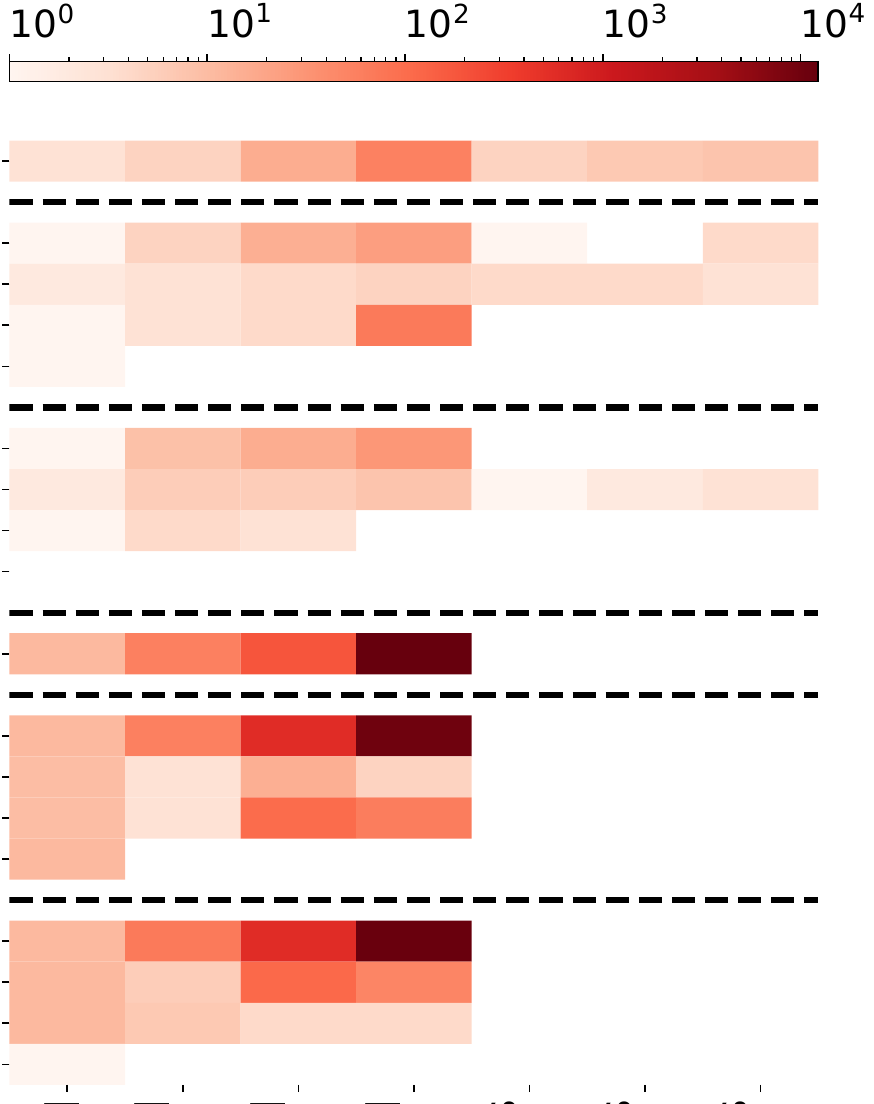}
            \hrule
            \medskip
            \includegraphics[trim={10.2cm 4.4cm 0.4cm 4.7cm},clip,height=3.48cm]{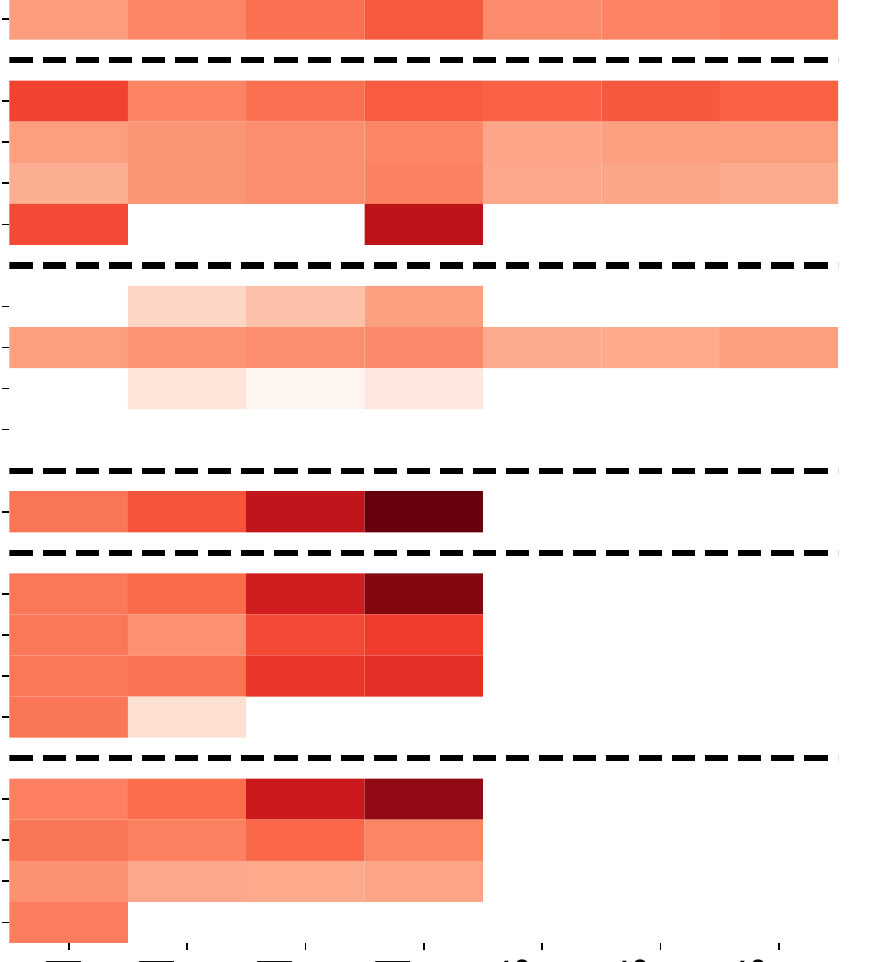}
            \hrule
            \medskip
            \includegraphics[trim={10.2cm 1.5cm 0.4cm 4.7cm},clip,height=4.10cm]{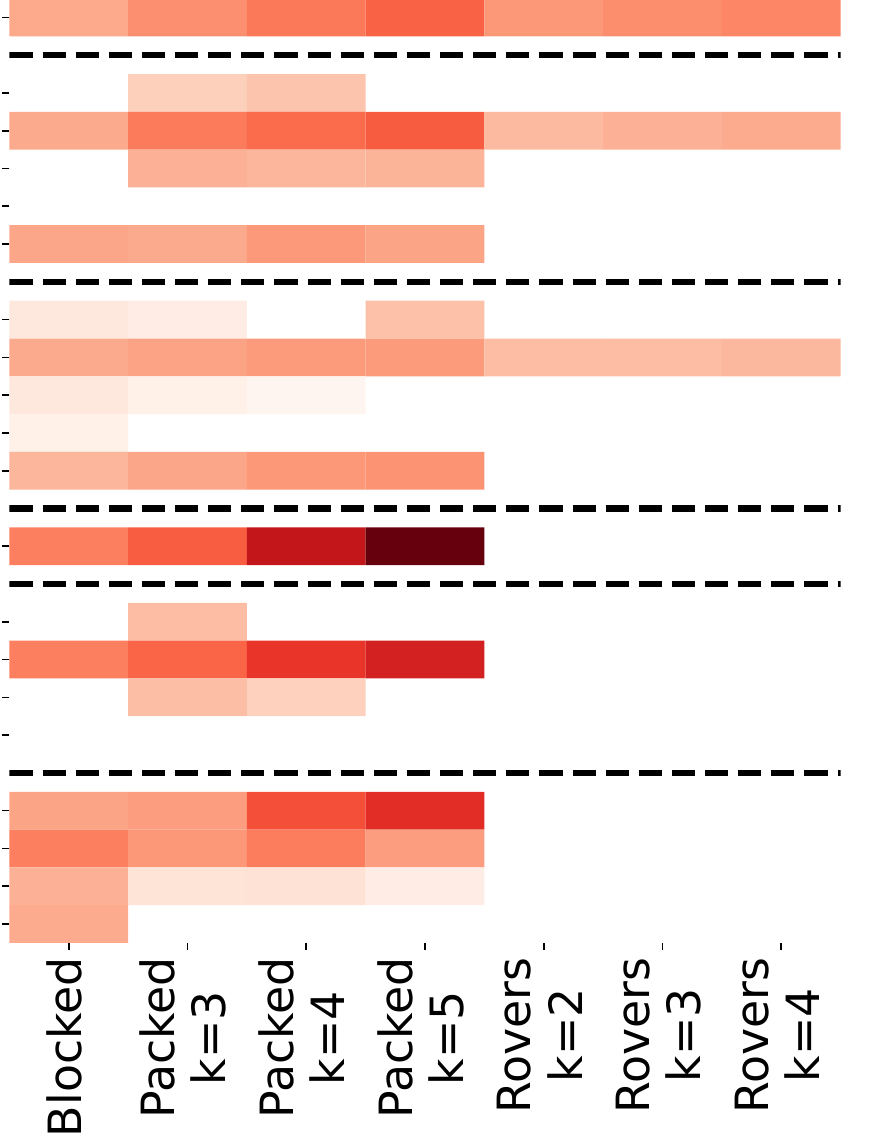}
            \caption{Samples used (log)}
            \label{fig:samplesHeatmap}
        \end{subfigure}%
        \begin{subfigure}[b]{0.179\textwidth}
            \includegraphics[trim={10.2cm 4.4cm 0.4cm 2.3cm},clip,height=4.01cm]{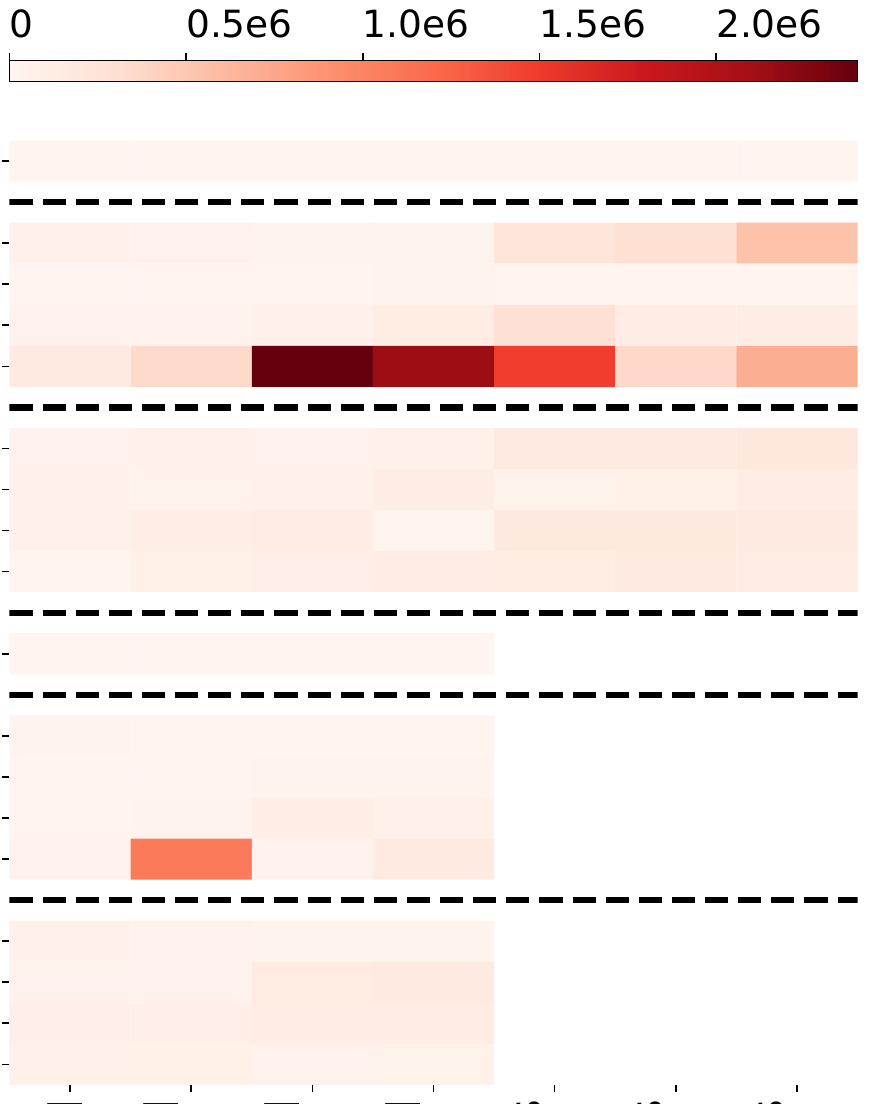}
            \hrule
            \medskip
            \includegraphics[trim={10.2cm 4.4cm 0.4cm 4.7cm},clip,height=3.48cm]{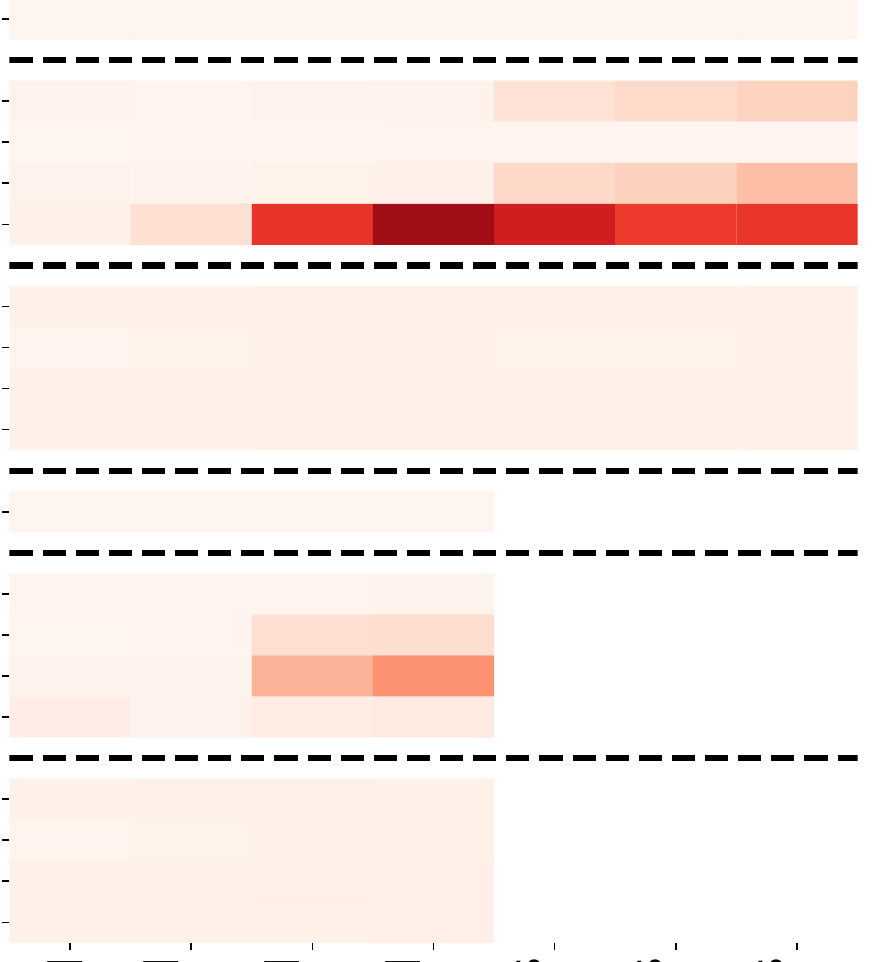}
            \hrule
            \medskip
            \includegraphics[trim={10.2cm 1.5cm 0.4cm 4.7cm},clip,height=4.10cm]{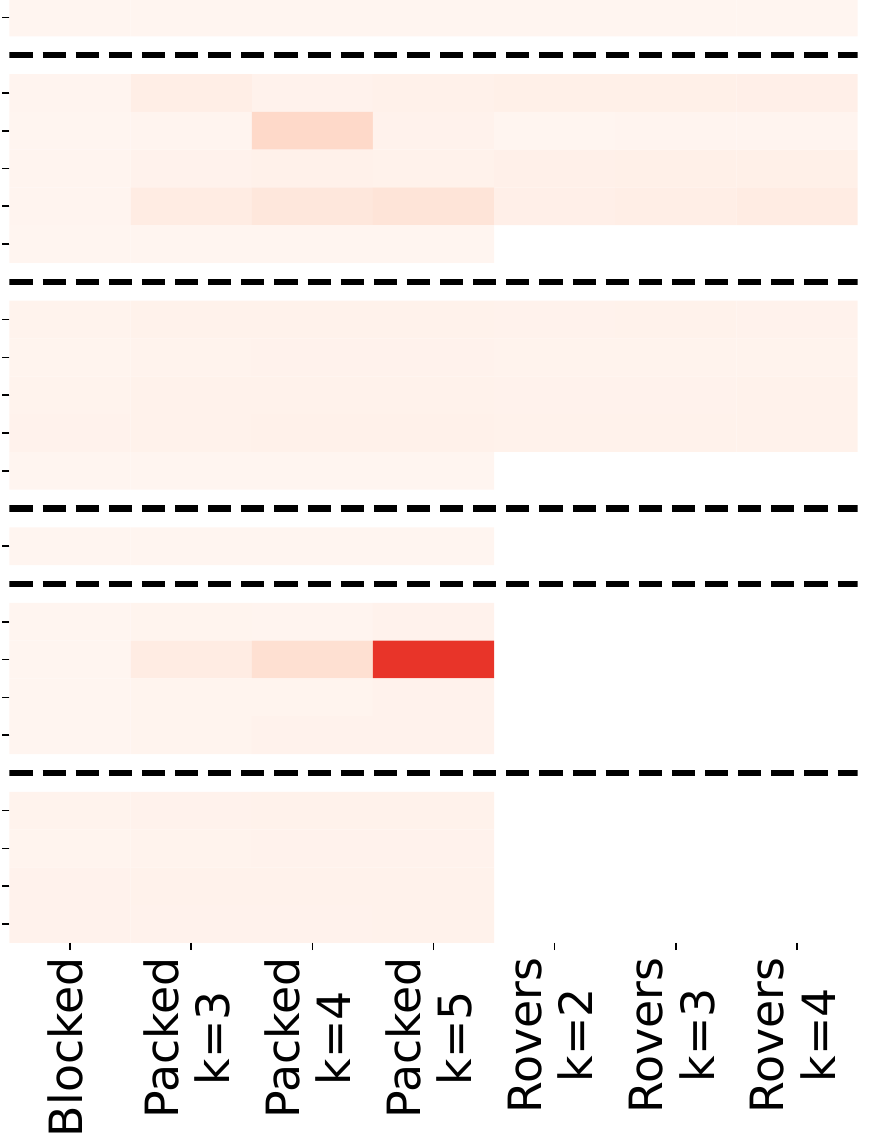}
            \caption{Tokens used}
            \label{fig:tokensHeatmap}
        \end{subfigure}%
        \begin{subfigure}[b]{0.179\textwidth}
            \includegraphics[trim={10.2cm 4.4cm 0.4cm 2.3cm},clip,height=4.01cm]{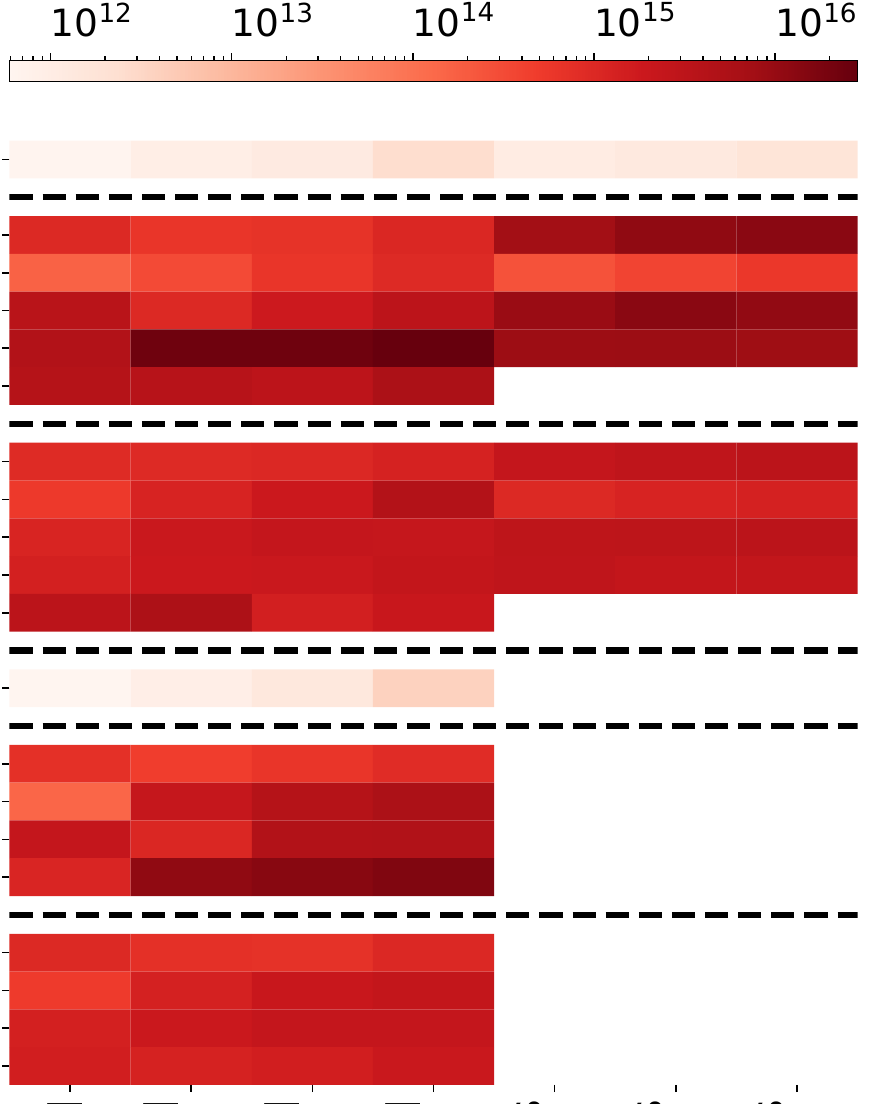}
            \hrule
            \medskip
            \includegraphics[trim={10.2cm 4.4cm 0.4cm 4.7cm},clip,height=3.48cm]{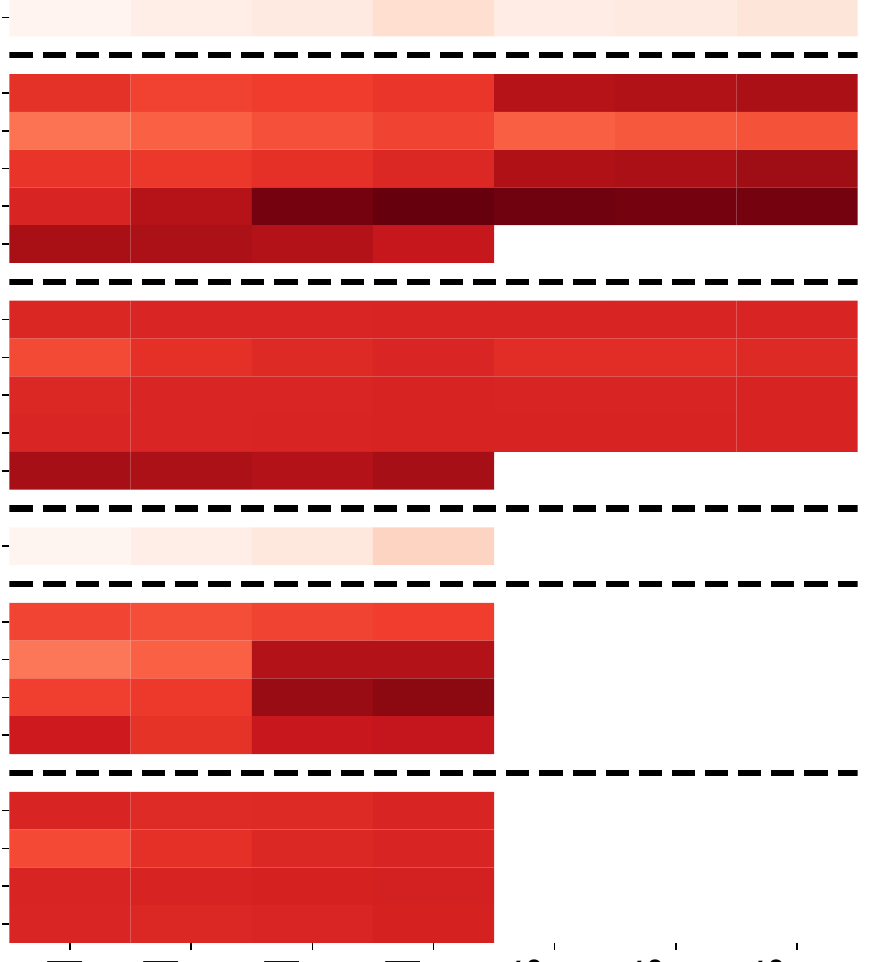}
            \hrule
            \medskip
            \includegraphics[trim={10.2cm 1.5cm 0.4cm 4.7cm},clip,height=4.10cm]{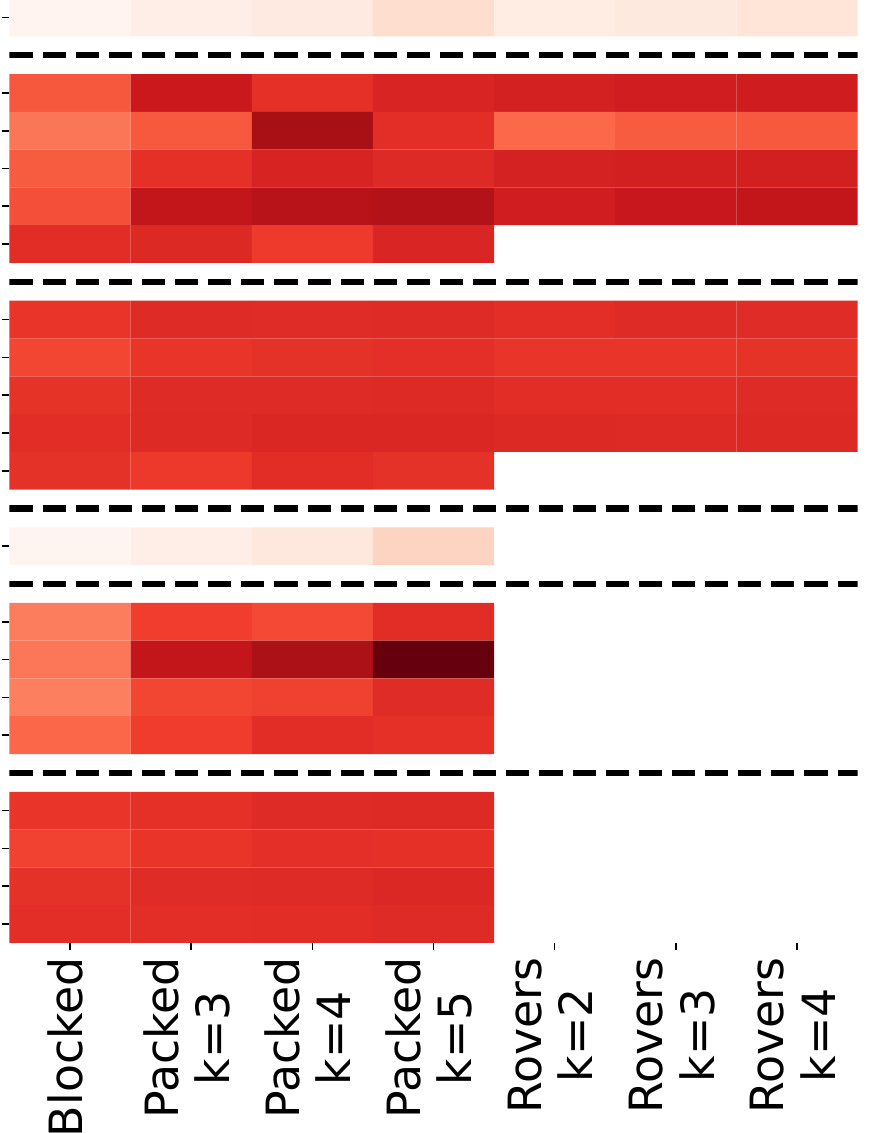}
            \caption{$\sim$FLOPs used (log)}
            \label{fig:flopsHeatmap}
        \end{subfigure}%
    }
    \caption{Heatmaps of metrics related to algorithm operation, averaged across 50 problems per domain. (d) does not include tasks that fail due to token limits. (c) and (e) use log scale due to large differences between algorithms. \direct{} methods make the most PDDL and sampling mistakes, and use the most tokens. Bilevel methods use the most samples in Packing domain with $k=5$. LLM methods use orders of magnitude more FLOPs, approximated as $2\cdot\mathrm{tokens}\cdot\mathrm{parameters}$ (at an underestimate of $10^{10}$ active parameters) for LLMs and as proportional to energy use for engineered methods (an overestimate). (LLM usage legend: No=No LLM, Pd=\pddl{}, Ps=\poses{}, PP=\pddl{}+\poses{}, In=\integrated{}.)}
    \label{fig:heatmaps}

\end{figure*}

\section{Experimental Design}

\subsection{Domains}

We used three domains introduced by Garrett et al.~\cite{garrett2020pddlstream} to evaluate the planners (Fig.~\ref{fig:domains}). The \textit{Blocked} domain requires placing one of two blue boxes on a target region, but a red box blocks the nearest blue box. The planner must move the red box or pick the farther blue box. The goal in the \textit{Packing} domain is to place $k$ boxes on a small square region. As $k$ becomes larger, the solution set shrinks. In the \textit{Rovers} domain, two rovers must collaboratively collect one rock sample and one soil sample, photograph $k$ objectives, and communicate the results to the lander via line of sight. The problem horizon grows with $k$, allowing us to study the ability of our planners to handle longer-horizon tasks. We considered only \adaptive{} methods in Rovers domains, since the long horizon and geometric constraints make even the base \bilevel{} method ineffective. We evaluated \abstraction{} design in Packing and Blocked domains with \adaptive{} as the planner.

\subsection{Large Language Models}

We used three LLMs: Gemini 2.5 Flash, Gemini 3 Flash, and GPT-5 mini. Gemini 2.5 and 3 Flash both have a token limit of $10^6$ input tokens per minute, while GPT-5 mini has a per-prompt input token limit of $272\,000$ tokens. 

\subsection{Experimental Protocol}

We used $k\in\{3,4,5\}$ for Packing and $k\in\{2,3,4\}$ for Rovers, resulting in seven effective domains. Problem instances varied only in the initial continuous values for the objects in the domain. We ran each planner for a maximum of $T=300$ seconds, and considered it successful if it produced a valid plan before timing out or reaching the LLM's resource limits---i.e., we counted time-outs and token limits as failures. 

\subsection{Statistical Analysis}
\label{sec:statisticalAnalysis}

We conducted $50$ trials per experiment, setting the random seed for the initial state, the randomness of \adaptive{} and \bilevel{}, and the LLM's text generation.\footnote{The LLM APIs do not guarantee deterministic outputs, so our exact results may not be replicable. This does not affect our statistical claims.} We computed the average success rate and the average planning time \textit{for successful attempts} for each algorithm on each domain.

For success rates, we treated each run as a Bernoulli observation and used McNemar's test for paired binary data to determine if each pair of algorithms for one fixed LLM is statistically distinguishable. Because our sample sizes are small, we used the exact two-sided McNemar test on the discordant counts. To control the family-wise error rate across all pairwise comparisons, we adjusted $p$-values with the Holm step-down procedure and used a threshold of $p=0.05$. We summarize the results using a compact letter display (CLD): algorithms are declared distinct iff they do not share a letter.

For runtimes, we used a paired Wilcoxon signed-rank test. We studied only algorithms with $\geq30\%$ success rate on each domain, comparing pairs of algorithms on runs where both algorithms succeeded. This allows us to determine whether, over the problems where two algorithms succeed, one is faster than the other. We again summarize the results using a CLD with Holm corrections. For algorithms below $30\%$ success, we report their average success time but leave them out of the CLD, since they can rarely be distinguished from others. 

\section{Results and Analysis}

Fig.~\ref{fig:successRate} summarizes our success rate results. Non-LLM algorithms are statistically indistinguishable from the best in all cases, confirming the (well established) feasibility of TAMP in these domains. \direct{} methods consistently outperform \thinking{} variants. In 129 out of 132 pairwise comparisons, the \direct{} method is indistinguishable or better than its \thinking{} counterpart; the three exceptions are \adaptive{}-\integrated{} in Packing domains with Gemini 2.5 Flash. The superiority of \direct{} suggests that computation is better spent on the formal reasoning of the TAMP algorithm and not on the LLM's internal thinking, which is not guaranteed to be correct. In other words, our LLM-Modulo TAMP methods are most effective when the LLM quickly produces many candidate solutions (even if flawed), leaving the complex, geometrically-aware reasoning to the base TAMP system. Similarly, despite the intuitive appeal of leveraging the LLM's semantic knowledge to simultaneously process PDDL and geometric constraints, \integrated{} LLM approaches perform poorly: \adaptive{}-\integrated{} is distinguishably worse than the top two methods in all domains, and only \bilevel{}-\direct{}-\integrated{} is indistinguishable from the top performer in one domain (Blocked) with Gemini 2.5 Flash. This reinforces the claim that complex reasoning is better handled by the base TAMP system. Approaches that use the LLM to generate both \poses{} and \pddl{} plans (separately) are either indistinguishable or worse than the worse between \pddl{}-only and \poses{}-only LLM methods. Using the LLM to produce the domain \abstraction{} results in poor planning performance in all cases. 

Rovers is our most difficult domain, since it has the longest horizon and the geometric constraints impose the most limits on which PDDL plans are valid: each objective, rock, and soil can only be acted on by one specific rover, which is only apparent by reasoning about the obstacle that divides the domain in two. All LLM-based approaches perform poorly in Rovers domains. Even \integrated{} variants, which can reason simultaneously about PDDL and geometry, fail.
\pddl{}-only methods with Gemini LLMs work well on most Packing domains, where all valid PDDL plans are executable; exceptions are \thinking{} variants of Gemini 3 Flash with $k\ge4$. Most \pddl{}-only methods struggle in the Blocked domain, where no PDDL precondition expresses the need to move the red box before picking the nearest blue box. \poses{} methods perform surprisingly well across domains and LLMs. For GPT-5 mini, \poses{} approaches are the only ones that perform well. For Gemini 3 Flash, \adaptive{}-\direct{}-\poses{} is indistinguishable from the best performer in all non-Rovers domains, even in Packing with $k=5$, where the space of valid pose combinations is highly restricted.

Runtime results in Fig.~\ref{fig:time} statistically separate non-LLM and LLM methods in all domains. Among pairwise comparisons we execute (i.e., success pairs $\geq30\%$), \direct{} variants are faster  than \thinking{} variants in 32 out of 35 cases, and indistinguishable in the other three---exceptions are \bilevel{}-\integrated{} variants in Packing domains with Gemini 2.5 Flash. \direct{} variants compensate for additional errors with faster responses, which are corrected by the TAMP algorithm.

LLM-based planners can fail for giving up (i.e., stating that no plan or pose was possible), timing out, or exhausting the token limit. When used for \abstraction{} design, they can also produce a plan that is valid under the generated abstraction but invalid in the ground-truth domain. Giving up and generating invalid plans cannot be overcome with additional computation. Fig.~\ref{fig:failureReasons} shows the fraction of problems that failed for each reason. While resource constraints were the main culprit for most failures of LLM-based methods, LLM planners made multiple errors beyond such constraints. Note that these constraints are not artificial: robotics applications necessitate planning within time and computation budgets. We can further identify the algorithm-problem combinations for which time-outs did not dominate the failures. For Gemini 2.5 Flash, \bilevel{} methods failed primarily due to giving up in the Blocked domain, as did most \adaptive{} methods that sample continuous values in Rovers domains. For Gemini 3 Flash, only sampling-generating failures in Rovers domains were caused primarily by giving up. For GPT-5 mini, the majority of \direct{} methods failed by giving up in all domains. \abstraction{}-design approaches across LLMs and domains failed frequently for generating invalid plans. 

Fig.~\ref{fig:heatmaps} provides additional insight into how the various algorithms operate. The most intriguing finding is that introducing geometric information in the prompt may cause (non-thinking) LLMs to commit PDDL errors. Only \direct{} methods made a large number of PDDL errors, especially with Gemini 2.5 Flash (Fig.~\ref{fig:pddlFailuresHeatmap}). \direct{}-\integrated{} methods made many of these errors in the Packing domain, which requires simply a sequence of \texttt{move-pick-move-place} actions. Our logs demonstrate that these failures were primarily split between missing preconditions (e.g., attempting to pick before moving) and using invalid actions (e.g., applying an action to the incorrect optimistic object). Notably, the large differences in the number of PDDL errors between \integrated{} and \pddl{} or \pddl{}+\poses{} methods, whose PDDL prompts only differ in the inclusion of geometric information, reveal that the additional geometric information and constraints cause the (non-thinking) LLMs to divert attention away from PDDL constraints. PDDL failures in Rovers domains were primarily due to plans that did not achieve the goal (e.g., did not return the rovers to their initial position). Sampling failures primarily affected \adaptive{}-\direct{}-\integrated{} in Rovers domains, where Gemini 3 Flash produced many poses that violated geometric constraints (Fig.~\ref{fig:samplingFailuresHeatmap}). \bilevel{} methods in the Packing domain with $k=5$ boxes used considerably more samples than other methods in any problem. The increased sample count results from the fact that most placements of the boxes do not permit placing all remaining boxes (Fig.~\ref{fig:samplesHeatmap}). \direct{} LLMs used more tokens than their \thinking{} counterparts, and the most resource-intensive setting was \adaptive{}-\direct{}-\integrated{} with Gemini 3 Flash in Packing and Rovers domains (Fig.~\ref{fig:tokensHeatmap}). Approximating floating point operations (FLOPs) in terms of tokens and estimated parameters for LLM calls (an underestimate)~\cite{kaplan2020scaling} and in terms of energy expense for local computations (an overestimate), LLM-based methods used orders of magnitude more FLOPs than engineered baselines (Fig.~\ref{fig:flopsHeatmap}). Note that we did not account for local FLOPs for LLM-based planners, which would be dwarfed by the LLM inference FLOPs.

\abstraction{}-design methods failed for a breadth of factors, including: failing to satisfy the syntax requirements (e.g., using conditional effects despite the prompt explicitly stating they are not supported), ignoring the key requirement of ensuring collision-free motions, and providing an initial state and goal where the goal already holds. 

\section{Conclusions and Limitations}
We introduced 16 LLM-Modulo planners for TAMP that consume formal PDDLStream specifications and whose outputs are verified by a base TAMP method (either \bilevel{} or \adaptive{}). We conducted a comprehensive evaluation on seven existing TAMP domains, repeating each evaluation $50$ times for statistical significance. Our findings demonstrate that, while LLMs are capable of solving many novel TAMP problems, they cannot match the performance of the base TAMP methods when used as drop-in replacements for TAMP components. Imposing time and computation budgets (e.g., time-outs or token limits) exacerbates the gap between the non-LLM and LLM-based planners we studied. Qualitative analyses identified design choices and trade-offs.

Despite many trials, 
some performance differences are not statistically significant. While our LLM-based planners are representative of plausible solutions, there are many alternative choices that could improve LLM TAMP performance---e.g., providing example solutions, using a different LLM, or expressing problems in a format other than PDDLStream. 

\section*{Acknowledgment}

ChatGPT helped design the statistical analysis of Section~\ref{sec:statisticalAnalysis} and plotting code for Figs.~\ref{fig:LLMPlanners}--\ref{fig:heatmaps}, which were verified by the author. Gemini conducted proofreading and grammar checking. Evaluations were supported in part by SUNY System Administration through the SUNY AI Platform, and by the Institute for Advanced Computational Science at Stony Brook University through OpenAI API credits.

\bibliographystyle{IEEEtran}
\bibliography{LLMTAMP}
\end{document}